\documentclass{article}
\usepackage[T1]{fontenc}
\usepackage[utf8]{inputenc}
\usepackage{amsmath,amsfonts,amsthm,amssymb}
\usepackage{setspace}
\usepackage{fullpage}
\usepackage{fancyhdr}
\usepackage{xspace, xcolor}

\usepackage{verbatim}
\usepackage{lastpage}
\usepackage{extramarks}
\usepackage{natbib}
\usepackage{diagbox}

\usepackage{titletoc}

\usepackage{times}
\usepackage{setspace}
\usepackage{graphicx,float,wrapfig}
\usepackage{algorithm}
\usepackage{enumitem}
\usepackage[noend]{algpseudocode}
\usepackage{subcaption}
\allowdisplaybreaks

\newtheorem{theorem}{Theorem}[section]
\newtheorem{lemma}[theorem]{Lemma}

\newtheorem{proposition}[theorem]{Proposition}

\newtheorem{cond}{Condition}
\theoremstyle{definition}

\vfuzz \hfuzz
\usepackage{hyperref}
\hypersetup{
	colorlinks   = true, %
	urlcolor     = red, %
	linkcolor    = blue, %
	citecolor   = blue %
}

\usepackage{amsmath,amsfonts,bm}

\newcommand*{\defeq}{\triangleq}

\def\1{\bm{1}}

\makeatletter
\newcommand{\ve}{\@ifnextchar\bgroup{\velong}{{\bm{e}}}}
\newcommand{\velong}[1]{{\bm{#1}}}
\makeatother

\DeclareMathAlphabet{\mathsfit}{\encodingdefault}{\sfdefault}{m}{sl}
\SetMathAlphabet{\mathsfit}{bold}{\encodingdefault}{\sfdefault}{bx}{n}

\def\calA{{\mathcal{A}}}

\def\calC{{\mathcal{C}}}

\def\calE{{\mathcal{E}}}
\def\calF{{\mathcal{F}}}
\def\calG{{\mathcal{G}}}

\def\calN{{\mathcal{N}}}
\def\calO{{\mathcal{O}}}

\def\calR{{\mathcal{R}}}
\def\calS{{\mathcal{S}}}

\newcommand{\E}{\mathbb{E}}
\newcommand{\R}{\mathbb{R}}

\newcommand{\KL}{D_{\mathrm{KL}}}

\DeclareMathOperator*{\argmax}{argmax}
\DeclareMathOperator*{\argmin}{argmin}

\def\({\left(}
\def\){\right)}
\def\[{\left[}
\def\]{\right]}
\newcommand{\poly}{\mathrm{poly}}
\newcommand{\polylog}{\mathrm{polylog}}

\newcommand{\dotp}[2]{\left<#1, #2\right>}

\newcommand{\abs}[1]{{\left| {#1} \right|}}

\DeclareMathOperator{\supp}{supp}
\newcommand{\ind}[1]{\mathbb{I}\left[ #1 \right]}

\newcommand{\bigO}{\calO}

\newcommand{\dd}{\mathrm{d}}
\newcommand{\TV}{D_{\rm TV}}

\newcommand{\renyi}{R\'enyi\xspace}

\newcommand{\comp}{\calC}
\newcommand{\Deltamin}{\Delta_{\rm min}}
\newcommand{\Deltamax}{\Delta_{\rm max}}
\newcommand{\reg}{\mathrm{Reg}}

\newcommand{\acc}{\calE_{\rm acc}}
\newcommand{\init}{\calE_{\rm init}}
\newcommand{\event}{\calE}
\newcommand{\cset}{\Lambda}
\newcommand{\true}{\rm true}

\newcommand{\obsa}{o^{\rm sa}}
\newcommand{\obr}{o^{\rm r}}
\newcommand{\mumin}{\mu_{\rm min}}

\newcommand{\minit}{{m_{\rm init}}}
\newcommand{\ob}{o}
\newcommand{\Ob}{O}
\newcommand{\initf}{{\hat{f}}}
\newcommand{\truef}{{f^\star}}
\newcommand{\const}{c}
\newcommand{\Rmax}{R_{\rm max}}
\newcommand{\sigmamax}{\sigma_{\rm max}}
\newcommand{\sigmamin}{\sigma_{\rm min}}
\newcommand{\algname}{\textsc{T2C}\xspace}

\newcommand{\vol}{{\rm Vol}}

\renewcommand{\Pr}{\mathop{\rm Pr}\nolimits}

\title{Asymptotic Instance-Optimal Algorithms for Interactive Decision Making}

\author{Kefan Dong \\ 
	Stanford University \\
	\texttt{kefandong@stanford.edu}
	\and
	Tengyu Ma \\
	Stanford University \\
	\texttt{tengyuma@stanford.edu}
}

\begin{document}
	
	\maketitle
	
	\begin{abstract}
		Past research on interactive decision making problems (bandits, reinforcement learning, etc.) mostly focuses on the minimax regret that measures the algorithm's performance on the hardest instance. However, an ideal algorithm should adapt to the complexity of a particular problem instance and incur smaller regrets on easy instances than worst-case instances. In this paper, we design the first asymptotic instance-optimal algorithm for general interactive decision making problems with finite number of decisions under mild conditions. 
		On \textit{every} instance $f$, our algorithm outperforms \emph{all} consistent algorithms (those achieving non-trivial regrets on all instances), and has asymptotic regret $\mathcal{C}(f) \ln n$, where $\mathcal{C}(f)$ is an exact characterization of the complexity of $f$. 
		The key step of the algorithm involves hypothesis testing with active data collection. It computes the most economical decisions with which the algorithm collects observations to test whether an estimated instance is indeed correct; thus, the complexity $\mathcal{C}(f)$ is the minimum cost to test the instance $f$ against other instances.  
		Our results, instantiated on concrete problems, recover the classical gap-dependent bounds for multi-armed bandits \citep{lai1985asymptotically} and prior works on linear bandits \citep{lattimore2017end}, and improve upon the previous best instance-dependent upper bound \citep{xu2021fine} for reinforcement learning. 
	\end{abstract}
	
	\section{Introduction}\label{sec:intro}
Bandit and reinforcement learning (RL) algorithms demonstrated a wide range of successful real-life applications \citep{alphago16,alphago17,mnih2013playing,berner2019dota,vinyals2019grandmaster,mnih2015human,degrave2022magnetic}.
Past works have theoretically studied the regret or sample complexity of various interactive decision making problems, such as contextual bandits, reinforcement learning (RL), partially observable Markov decision process  (see \citet{azar2017minimax,jin2018q,dong2021provable,li2019nearly,agarwal2014taming,foster2020beyond,jin2020provably}, and references therein). 
Recently, \citet{foster2021statistical} present a unified algorithmic principle for achieving the minimax regret---the optimal regret for the worst-case problem instances.

However, minimax regret bounds do not necessarily always present a full picture of the statistical complexity of the problem. 
They characterize the complexity of the most difficult instances, but potentially many other instances are much easier. 
An ideal algorithm should adapt to the complexity of a particular instance and incur smaller regrets on easy instances than the worst-case instances. 
Thus, an ideal regret bound should be instance-dependent, that is, depending on some properties of each instance. 
Prior works designed algorithms with instance-dependent regret bounds that are stronger than minimax regret bounds, but they are often not directly comparable because they depend on different properties of the instances, such as the gap conditions and the variance of the value function~\citep{zanette2019tighter, xu2021fine, foster2020instance,tirinzoni2021fully}.  

A more ambitious and challenging goal is to design \emph{instance-optimal} algorithms that outperform, on \textit{every} instance, \textit{all} consistent algorithms (those achieving non-trivial regrets on all instances). Past works designed instance-optimal algorithms for multi-armed bandit \citep{lai1985asymptotically}, linear bandits \citep{kirschner2021asymptotically,hao2020adaptive}, Lipschitz bandits \citep{magureanu2014lipschitz}, and ergodic MDPs \citep{ok2018exploration}. 
However, instance-optimal regret bounds for tabular reinforcement learning remain an open question,  despite recent progress \citep{tirinzoni2021fully,tirinzoni2022near}.
The challenge partly stems from the fact that the \emph{existence} of such an instance-optimal algorithm is even a priori unclear for general interactive decision making problems. 
Conceivably, each algorithm can have its own specialty on a subset of instances, and no algorithm can dominate all others on all instances.

Somewhat surprisingly, we prove that there exists a simple algorithm (\algname, stated in Alg.~\ref{alg:main-finite}) that is asymptotic instance-optimal for general interactive decision making problems with finite number of decisions.%
We determine the \textit{exact} leading term of the optimal asymptotic regret for instance $f$ to be $\comp(f)\ln n$. Here, $n$ is the number of interactions and $\comp(f)$ is a complexity measure for the instance $f$ that intuitively captures the difficulty of distinguishing $f$ from other instances (that have different optimal decisions) using observations. Concretely, under mild conditions on the interactive decision problem, our algorithm achieves an asymptotic regret $\comp(f)\ln n$ (Theorem~\ref{thm:main}) for every instance $f$, while every consistent algorithm must have an asymptotic regret at least $\comp(f)\ln n$ (Theorem~\ref{thm:lower-bound}).

Our algorithm consists of three simple steps. 
First, it explores uniformly for $o(1)$-fraction of the steps and computes the MLE estimate of the instance with relatively low confidence. 
Then, it tests whether the estimate instance (or, precisely, its associated optimal decision) is indeed correct using the most economical set of queries/decisions. Concretely, it computes a set of decisions with minimal regret, such that, using a log-likelihood ratio test, it can either distinguish the estimated instance from all other instances (with different optimal decisions) with high confidence, or determine that our estimate was incorrect. Thirdly, with the high-confidence estimate, it commits to the optimal decision of the estimated instance in the rest of the steps. The algorithmic framework essentially reduces the problem to the key second step --- a problem of optimal hypothesis testing with active data collection. 

Our results recover the classical gap-dependent regret bounds for multi-armed bandits \citep{lai1985asymptotically} and prior works on linear bandits \citep{lattimore2017end,hao2020adaptive}. As an instantiation of the general algorithm, we present the first asymptotic instance-optimal algorithm for episodic tabular RL, improving upon prior instance-dependent algorithms~\citep{xu2021fine,simchowitz2019non,tirinzoni2021fully,tirinzoni2022near}.%

\subsection{Additional Related Works}\label{sec:related-work}
\paragraph{Instance-optimal and instance-dependent analysis.}
Prior works have designed instance-optimal algorithms for specific interactive decision making problems.
Variants of UCB algorithms are instance-optimal for bandits with various assumptions \citep{lattimore2020bandit,gupta2021multi,tirinzoni2020novel,degenne2020structure,magureanu2014lipschitz}, but are suboptimal for linear bandits \citep{lattimore2017end}. These algorithms rely on the optimism-in-face-of-uncertainty principle to deal with exploration-exploitation tradeoff, whereas our algorithm explicitly finds the best tradeoff.
There are also non-optimistic instance-optimal algorithms for linear bandits \citep{kirschner2021asymptotically,lattimore2017end,hao2020adaptive}, ergodic MDPs \citep{ok2018exploration,burnetas1997optimal}, and others \citep{graves1997asymptotically,rajeev1989asymptotically}.

Another line of research focuses on instance-optimal pure exploration algorithms in multi-armed bandits \citep{mason2020finding,marjani2022complexity,chen2017towards,chen2017nearly} and deterministic MDPs \citep{tirinzoni2022near}, where the goal is to use minimum number of samples to identify an $\epsilon$-optimal decision.

Many algorithms' regret bounds depend on some properties of instances such as the gap condition. \citet{foster2020instance} prove a gap-dependent regret bound for contextual bandits. For reinforcement learning, the regret bound may depend on the variance of the optimal value function~\citep{zanette2019tighter} or the gap of the $Q$-function~\citep{xu2021fine,simchowitz2019non,yang2021q}.
\citet{xu2021fine,foster2020instance} also prove that the gap-dependent bounds cannot be improve on some instances. To some extent, these instance-dependent lower bounds can be viewed as minimax bounds for a fine-grained instance family (e.g., all instances with the same $Q$-function gap), and therefore are different from ours.

\paragraph{Instance-optimal algorithm via log-likelihood ratio test.} The idea of using log-likelihood ratio test to design instance-optimal algorithms traces back to \citet{rajeev1989asymptotically,graves1997asymptotically}. \citet{rajeev1989asymptotically} design an asymptotic instance-optimal algorithm for general decision making problems with finite hypothesis class, finite state-action space, and known rewards. The algorithm in \citet{graves1997asymptotically} is instance-optimal for infinite horizon MDPs where the Markov chain induced by every policy is uniformly recurrent, meaning that the probability of visiting every state is uniformly lower bounded away from 0. In comparison, our analysis is applicable to infinite hypothesis settings without the uniform recurrent assumption (e.g., episodic tabular RL).
	\section{Setup and Notations}\label{sec:setup}

\paragraph{Additional notations.} We use $\poly(n)$ to denote a polynomial in $n$. For two non-negative sequences $a_n,b_n$, we write $a_n=\bigO(b_n)$ if $\limsup_{n\to\infty}a_n/b_n<\infty$ and $a_n=o(b_n)$ if $\limsup_{n\to\infty}a_n/b_n=0.$ %

\paragraph{Interactive decision making problems.} 
We focus on interactive decision making with structured observations \citep{foster2021statistical}, which includes bandits and reinforcement learning as special cases. An interactive decision making problem is defined by a family of decisions $\Pi$, a space of observations $\Ob$, a reward function $R:\Ob\to\R$, and a function $f$ (also called an instance) that maps a decision $\pi\in\Pi$ to a distribution over observations $f[\pi]$. We use $f[\pi](\cdot):\Ob\to\R_+$ to denote the density function of the distribution $f[\pi]$. We assume that the reward $R$ is a deterministic and known function.

An environment picks an ground-truth instance $\truef$ from a instance family $\calF$, and then an agent (which knows the instance family $\calF$ but not the ground-truth $\truef$) interacts with the environment for $n$ total rounds. In round $t\le n$,
\begin{enumerate}
	\item the learner selects a decision $\pi_t$ from the decision class $\Pi$, and
	\item the environment generates an observation $\ob_t$ following the ground-truth distribution $\truef[\pi_t]$ and reveals the observation. Then the agent receives a reward $R(\ob_t)$.
\end{enumerate}

As an example, multi-armed bandits, linear bandits, and reinforcement learning all belong to interactive decision making problems. For bandits, a decision $\pi$ corresponds to an action and an observation $\ob$ corresponds to a reward. For reinforcement learning, a decision $\pi$ is a deterministic policy that maps from states to actions, and an observation $\ob$ is a trajectory (including the reward at each step). In other words, a round of interactions in the interactive decision making problem corresponds to an episode of the reinforcement learning problem. We will formally discuss the reinforcement learning setup in Section~\ref{sec:infinite}.

Let $R_f(\pi)=\E_{\ob\sim f[\pi]}[R(\ob)]$ be the expected reward for decision $\pi$ under instance $f$, and $\pi^\star(f)\defeq \argmax_\pi R_f(\pi)$ the optimal decision of instance $f$. The expected regret measures how much worse the agent's decision is than the optimal decision:
\begin{align}
	\reg_{f,n}=\E_f\[\sum_{t=1}^{n}\Big(\max_{\pi\in\Pi}R_f(\pi)-R_f(\pi_t)\Big)\].
\end{align}
We consider the case where the decision family $\Pi$ is \emph{finite}, and every instance $f\in\calF$ has a unique optimal decision, denoted by $\pi^\star(f)$. We also assume that for every $f\in\calF$ and $\pi\in\Pi$, $0\le R(\ob)\le \Rmax$ almost surely when $\ob$ is drawn from the distribution $f[\pi]$, and $\sup_\ob f[\pi](\ob)< \infty$.

\paragraph{Consistent algorithms and asymptotic instance-optimality.} 
We first note that it's unreasonable to ask for an algorithm that can outperform or match any \emph{arbitrary} algorithm on \emph{every} instance. This is because, for any instance $f\in\calF$, a bespoke algorithm that always outputs $\pi^\star(f)$ can achieve zero regret on instance $f$ (though it has terrible regrets for other instances). Therefore, if an algorithm can outperform or match any algorithm on any instance, it must have zero regret on every instance, which is generally impossible. Instead, we are interested in finding an algorithm that is as good as any other reasonable algorithms that are not only customized to a single instance and completely fail on other instances.

We say an algorithm is consistent if its expected regret satisfies $\reg_{f,n}=o(n^p)$ for \textit{every} $p>0$ and $f\in\calF$ \citep{lai1985asymptotically}. Most of the reasonable algorithms are consistent, such as the UCB algorithm for multi-armed bandits \citep{lattimore2020bandit}, UCBVI, and UCB-Q algorithm for tabular reinforcement learning \citep{simchowitz2019non,yang2021q}, because all of them achieve asymptotically $\bigO(\ln n)$ regrets on any instance, where $\bigO$ hides constants that depend on the property of the particular instance.\footnote{Here the regret scales in $\log n$ because we are in the asymptotic setting where the instance is fixed and $n$ tends to infinity. If the instance can depend on $n$ (which is not the setting we are interested in), then the minimax regret typically scales in $O(\sqrt{n})$.} However, consistent algorithms exclude the algorithm mentioned in the previous paragraph which is purely customized to a particular instance.

We say an algorithm is asymptotically \emph{instance-optimal} if on \emph{every} instance $f\in\calF$, $\limsup_{n\to\infty}\reg_{f,n}/\ln n$ is the smallest among all \emph{consistent} algorithms \citep{lai1985asymptotically}.
We note that even though an instance-optimal algorithm only needs to perform as well as every consistent algorithm, a priori, it's still unclear if such an algorithm exists.%

Some prior works have also used a slightly weaker definition in place of consistent algorithms, e.g., $\alpha$-uniformly good algorithms in~\citet{tirinzoni2021fully}, which allows a sublinear regret bound $\bigO( n^\alpha)$ for \textit{some} constant $\alpha < 1$. The alternative definition, though apparently includes more algorithms, does not change the essence. Our algorithm is still instance-optimal \textit{up to a constant factor}---a simple modification of the lower bound part of the proof shows that its asymptotic regret is at most $(1-\alpha)^{-1}$ factor bigger than any $\alpha$-uniformly good algorithms on any instance. This paper thus primarily compares with consistent algorithms for simplicity.

	\section{Main Results}\label{sec:mainresults}

In this section, we present an intrinsic complexity measure $\comp(f)$ of an instance $f$ (Section~\ref{sec:lowerbound}), and an instance-optimal algorithm that achieves an asymptotic regret $\comp(f)\ln n$ (Section~\ref{sec:algorithm}).
\subsection{Complexity Measure and Regret Lower Bounds}\label{sec:lowerbound}
In this section, our goal is to understand the minimum regret of consistent algorithms, which is an intrinsic complexity measure of an instance. We define an instance-dependent complexity measure $\comp(f)$ and prove that any consistent algorithm must have asymptotic regret at least $\comp(f)\ln n.$

The key observation is that any consistent algorithm has to collect enough information from the observations to tell apart the ground-truth instance from other instances with different optimal decisions. Consider the situation when a sequence of $n$ decisions is insufficient to distinguish two instances, denoted by $f$ and $g$, with different optimal decisions. If an algorithm achieves a sublinear regret on $f$, the sequence must contain $\pi^\star(f)$ (the optimal decision for $f$) at least $n-o(n)$ times. As a result, if the true instance were $g$, the algorithm suffers a linear regret (due to the linear number of $\pi^\star(f)$), and therefore cannot be consistent.

An ideal algorithm should find out decisions that can \textit{most efficiently} identify the ground-truth instance (or precisely the family of instances with the same optimal decision as the ground-truth). 
However, decisions collect information but also incur regrets. So the algorithm should pick a list of decisions with the best tradeoff between these two effects---minimizing the decisions' regret and collecting sufficient information to identify the true instance. 
Concretely, suppose a sequence of decisions includes $w_\pi$ occurrences of decision $\pi$. The regret of these decisions is $\sum_{\pi\in\Pi}w_\pi\Delta(f,\pi)$, where  $\Delta(f,\pi)\defeq R_f(\pi^\star(f))-R_f(\pi)$ is the sub-optimality gap of decision $\pi$. 
We use KL divergence between the observations of $\pi$ under two instances $f$ and $g$ (denoted by $\KL(f[\pi]\|g[\pi])$) to measure $\pi$'s power to distinguish them. The following optimization problem defines the optimal mixture of the decisions (in terms of the regret) that has sufficient power to distinguish $f$ from all other instances with different optimal decision. 
\begin{align}
	\comp(f, n)  \defeq \min_{w\in \R^{|\Pi|}_+}\;&\sum_{\pi\in\Pi}w_\pi\Delta(f,\pi)\label{equ:comp-def-1}\\
	\text{s.t.}\quad &\sum_{\pi\in\Pi}w_\pi\KL(f[\pi]\|g[\pi])\ge 1,\;\forall g\in\calF,\pi^\star(g)\neq\pi^\star(f),\label{equ:comp-def-2}\\
	&\|w\|_\infty\le  n.\label{equ:comp-def-3}
\end{align}
The last constraint makes sure that $w_\pi<\infty$ even if the decision has no regret, and we only care about the case when $n$ approaches infinity. The asymptotic complexity of $f$ is
\begin{align}
	\comp(f)\defeq \lim_{n\to\infty}\comp(f,n).
\end{align}
In Eq.~\eqref{equ:comp-def-2}, we only require separation between instances $f,g$ when they have different optimal decisions. As there are only finite decisions, there are finite equivalent classes, so $f$ and $g$ are in principle separable with sufficient number of observations. Thus, the complexity measure $\comp(f,n)$ is well defined as stated in the following lemma.
\begin{lemma}\label{lem:comp}
For any $f\in\calF$, $\comp(f,n)$ is non-increasing in $n$, and there exists $n_0>0$ such that for all $n>n_0$, $\comp(f,n)<\infty.$ As a corollary, $\comp(f)<\infty$ and is well defined.
\end{lemma}
Proof of Lemma~\ref{lem:comp} is deferred to Appendix~\ref{app:pf-comp}. 
In Section~\ref{sec:lowerbound}, we formalize the intuition above and prove that $\comp(f)$ is a lower bound of the asymptotic regret, as stated in the following theorem.
\begin{theorem}\label{thm:lower-bound}
	For every instance $f\in\calF$, the expected regret of any consistent algorithm satisfies
	\begin{align}
		\limsup_{n\to\infty}\frac{\reg_{f,n}}{\ln n}\ge \comp(f).
	\end{align}
\end{theorem}
We prove Theorem~\ref{thm:lower-bound} in Appendix~\ref{app:pf-lower-bound}. The proof of Theorem~\ref{thm:lower-bound} is inspired by previous works \citep{lattimore2017end,tirinzoni2021fully}. When applied to RL problems, our lower bound is very similar to that in \citet{tirinzoni2021fully} and the only difference is that their optimization problems omit the constraint Eq.~\eqref{equ:comp-def-3}. (Informally, the lower bound of \citet{tirinzoni2021fully} is equal to $\comp(f,\infty)$ in our language.) For some carefully-constructed hypothesis classes, there exists an instance $f$ such that $\lim_{n\to\infty}\comp(f,n)\neq \comp(f,\infty)$, meaning that the lower bound in \citet{tirinzoni2021fully} is not tight (see Appendix~\ref{sec:comp-tirinzoni-lb} for the construction and corresponding proof). However, this constraint can be removed for linear bandits (see \citet[Lemma 17]{lattimore2017end}). For the standard tabular RL problems whose hypothesis class is the entire set of possible RL problems with fixed number of states and actions, we conjecture that our lower bound would be the same as that in \citet{tirinzoni2021fully}.
We also note that \citet{tirinzoni2021fully} do not have a matching upper bound.

The complexity measure $\comp(f)$ can be computed explicitly in concrete settings.
$\comp(f)$ reduces to the well-known inverse action gap bound $\bigO(\sum_{a\in\calA,\Delta(a)>0}1/\Delta(a))$ for multi-armed bandits (Proposition~\ref{prop:comp-MAB}), and recovers the result of \citet{lattimore2017end} (Proposition~\ref{prop:comp-LB}). For reinforcement learning, \citet{tirinzoni2021fully} prove that the instance-optimal bound can be smaller than the gap-dependent bound $\bigO(\sum_{s,a:\Delta(s,a)>0}1/\Delta(s,a))$ in \citet{xu2021fine} and \citet{simchowitz2019non}.

\subsection{Instance-Optimal Algorithms}\label{sec:algorithm}
We first present the regret bound for our algorithm, and then discuss the \algname algorithm. For simplicity, we consider a finite hypothesis (i.e., $|\calF|<\infty$) here, and extend to infinite hypothesis case in Section~\ref{sec:infinite}.

We start by stating a condition that excludes abnormal observation distributions $f[\pi]$.
Recall that for any $\zeta\in(0,1)$, the \renyi divergence of two distributions $p,q$ is
\begin{align}
	D_{\zeta}(p\|q)=\frac{1}{\zeta-1}\ln \int_{x}p(x)^\zeta q(x)^{1-\zeta}\dd x.
\end{align}
The \renyi divergence $D_{\zeta}(p\|q)$ is non-decreasing in $\zeta$, and $\lim_{\zeta\uparrow 1}D_{\zeta}(p\|q)=\KL(p\|q)$ \citep{van2014renyi}.
We require the limits converge uniformly for all instances $g\in\calF$ and decisions $\pi\in\Pi$ with a $\zeta$ bounded away from 1 (the choice of the constants in Condition~\ref{cond:uniform-convergence} is not critical to our result), as stated below.
\begin{cond}\label{cond:uniform-convergence}
	For any fixed $\alpha>0,\epsilon>0$, instance $f\in\calF$, there exists $\lambda_0(\alpha,\epsilon,f)>0$ such that for all $\lambda\le \lambda_0(\alpha,\epsilon,f)$, $g\in\calF$ and $\pi\in\Pi$,
	$
		D_{1-\lambda}(f[\pi]\|g[\pi])\ge \min\{\KL(f[\pi]\|g[\pi])-\epsilon, \alpha\}.
	$
	Moreover, we require $\lambda_0(\alpha,\epsilon,f)\ge \epsilon^{\const_1}\min\{1/\alpha,\const_2\}^{\const_3}\iota(f)$ for some universal constants $\const_1,\const_2,\const_3>0$, where $\iota(f)>0$ is a function that only depends on $f$. 
\end{cond}
Condition~\ref{cond:uniform-convergence} holds for a wide range of distributions, such as Gaussian, Bernoulli, multinomial, Laplace with bounded mean, Log-normal \citep{gil2013renyi}, and tabular RL where $f[\pi]$ is a distribution over a trajectory consists of state, action and reward tuples (see Theorem~\ref{thm:rl-condition} for the proof of tabular RL). A  stronger but more interpretable variant of Condition~\ref{cond:uniform-convergence} is that the log density ratio of $f[\pi]$ and $g[\pi]$ has finite fourth moments (see Proposition~\ref{prop:bounded-moments}), therefore Condition~\ref{cond:uniform-convergence} can also be potentially verified for other distributions.

The main theorem analyzing Alg.~\ref{alg:main-finite} is shown below. The asymptotic regret of Alg.~\ref{alg:main-finite} matches the constants in the lower bound (Theorem~\ref{thm:lower-bound}), indicating the asymptotic instance-optimality of our algorithm.
\begin{theorem}\label{thm:main-finite}
	Suppose $\calF$ is a finite hypothesis class and satisfies Condition~\ref{cond:uniform-convergence} 
	The regret of Alg.~\ref{alg:main-finite} satisfies
	\begin{align}
		\limsup_{n\to\infty}\frac{\reg_{\truef,n}}{\ln n}\le \comp(\truef).
	\end{align}
\end{theorem}

\begin{algorithm}[t]
	\caption{Test-to-Commit (T2C)}
	\label{alg:main-finite}
	\begin{algorithmic}[1]
		\algblock[Name]{Start}{End}
		\algtext*{End}
		
		\State Parameters: the number of rounds of interactions $n$.
		
		\vspace{3pt}
		\hspace{-32pt}\textbf{Step 1: Initialization.}
		\State Play each decision $\lceil\frac{\ln n}{\ln \ln n}\rceil$ times. We denote these decisions by $\{\hat\pi_i\}_{i=1}^{\minit}$, where $\minit=|\Pi|\lceil\frac{\ln n}{\ln \ln n}\rceil$, and the corresponding observations by $\{\hat\ob_i\}_{i=1}^{\minit}$.

		\State Compute the max likelihood estimation (MLE) with arbitrary tie-breaking\label{line:2}
		\begin{align}
			\initf=\argmax_{f\in\calF}\;\sum_{i=1}^{\minit}\ln f[\hat\pi_i](\hat\ob_i).
		\end{align}
	
		\hspace{-32pt}\textbf{Step 2: Identification.}
		\State Let $\hat{w}$ be the solution of the program defining $\comp(\initf,(\ln\ln n)^{1/4})$ and $\bar{w}_\pi=\(1+\frac{1}{(\ln\ln n)^{1/4}}\)\hat{w}_\pi+\frac{1}{(\ln\ln n)^{1/4}}$ for all $\pi\in\Pi$.\label{line:3}
		
		\State Play each decision $\pi$ for $\lceil \bar{w}_\pi\ln n\rceil$ times. Denote these decisions by $w=\{\pi_i\}_{i=1}^{m}$ where $m=\sum_{\pi}\lceil \bar{w}_\pi\ln (n)\rceil$, and the corresponding observations by $\{\ob_{i}\}_{i=1}^{m}.$\label{line:4}
		\State 
		Run the log-likelihood ratio test on instance $\initf$, the sequence of decision $\{\pi_i\}_{i=1}^{m}$ and its corresponding observations $\{\ob_{i}\}_{i=1}^{m}$ (that is, compute the event $\acc^\initf$ defined in Eq.~\eqref{equ:acc}).\label{line:5}
		
		\vspace{3pt}
		\hspace{-32pt}\textbf{Step 3: Exploitation.}
		\If{$\acc^\initf=\true$}
		\State Commit to $\pi^\star(\initf)$ (i.e., run $\pi^\star(\initf)$ for the remaining steps).
		\Else
		\State Run UCB for the remaining steps.
		\EndIf
	\end{algorithmic}
\end{algorithm}
Our algorithm is stated in Alg.~\ref{alg:main-finite} and consists of three steps: initialization, identification, and exploitation. In the initialization step, we explore uniformly for a short period and compute the MLE estimate $\initf$ of the true instance (Line~\ref{line:2} of Alg.~\ref{alg:main-finite}), where we only requires $\initf$ to be accurate with moderate probability (i.e., $1-1/\ln n$). The estimation is used to compute the lowest-regret list of decisions that can distinguish the optimal decision of $\initf$. Since we only collect $o(\ln n)$ samples, the regret of this step is negligible asymptotically.

In the identification step, we hold the belief that $\initf$ is the true instance and solve $\comp(\initf, (\ln\ln n)^{1/4})$ to get the best list of decisions to fortify this belief (see Line~\ref{line:3} of Alg.~\ref{alg:main-finite}). Then we collect more samples using this list (with minor decorations) to boost the confidence of our initial estimation to $1-1/n$ (or reject it when the estimation is not accurate) by running the following log-likelihood ratio test:
\begin{align}\label{equ:acc}
	\acc^{\initf}&=\mathbb{I}\bigg[\forall g\in \calF\text{ and }\pi^\star(g)\neq \pi^\star(\initf), \sum_{i=1}^{m}\ln\frac{\initf[\pi_i](\ob_i)}{g[\pi_i](\ob_i)}\ge \ln n\bigg].
\end{align}
Intuitively, if $\initf$ is not the ground-truth $\truef$, $\acc^{\initf}$ is unlikely to hold because the expected log-likelihood ratio is non-positive: $$\E_{\ob\sim \truef[\pi]}[\ln(\initf[\pi](\ob)/\truef[\pi](\ob))]=-\KL(\truef[\pi]\|\initf[\pi])\le 0.$$ So with high probability a wrong guess cannot be accepted. On the other hand, the first constraint (Eq.~\eqref{equ:comp-def-2}) in the definition of $\comp(\initf)$ guarantees that when $\initf=\truef$, the expected log-likelihood ratio is large for all $g\in\calF$ and $\pi^\star(g)\neq \pi^\star(\initf)$, so an accurate guess will be accepted.
In this step $m=\tilde\Theta(\ln n)$, so Step 2 dominates the regret of Alg.~\ref{alg:main-finite}. In other words, the efficiency of the log-likelihood test is critical to our analysis.

Finally in the exploitation step, the algorithm commits to the optimal decision of $\initf$ if it believes this estimation with confidence $1-1/n$, or run a standard UCB algorithm as if on a multi-armed bandits problem (with a well-known $\bigO(\ln n)$ regret) when the initial estimation is not accurate (which happens with probability at most $1/\ln n$). Therefore the expected regret here is $\bigO(\ln n)/\ln n+\bigO(n)/n=\bigO(1)$.

Our three-stage algorithm is inspired by \citet{lattimore2017end}, where their test in Step 2 is specially designed for linear bandits with Gaussian noise. In contrast, the log-likelihood ratio test is a general hypothesis testing method. Hence, our algorithm works for a large family of interactive decision making problems including tabular RL.
	\section{Proof Sketches of the Main Results}\label{sec:proof-sketches}
In this section, we discuss the proof sketch of our main results. Section~\ref{sec:pf-lower-bound} discusses the proof sketch of the lower bound, and Section~\ref{sec:pf-upper-bound} shows the main lemmas for each step in Alg.~\ref{alg:main-finite}. In Section~\ref{sec:hypothesis-test}, we discuss the log-likelihood ratio test in detail. 

\subsection{Proof Sketch of Theorem~\ref{thm:lower-bound}}\label{sec:pf-lower-bound}
Recall that $\comp(f)$ is the minimum regret of a list of decisions that can distinguish the instance $f$ with all other instances $g$ (with a different optimal decision). So to prove the lower bound, we show that the sequence of decisions played by any consistent algorithm must also distinguish $f$, and thus, $\comp(f)$ lower bounds the regret of any consistent algorithm.

For any consistent algorithm, number of interactions $n>0$, and two instances $f,g\in\calF$ with different optimal decisions, let $P_{f,n}$ and $P_{g,n}$ denote the probability space generated by running the algorithm on instances $f$ and $g$ respectively for $n$ rounds. Since $f,g$ have different optimal decisions, $P_{f,n}$ and $P_{g,n}$ should be very different. Indeed, let $N_{\pi}$ be the random variable denoting the number of times decision $\pi$ is executed, and we consider the event $A=\ind{N_{\pi^\star(f)}\ge \frac{n}{2}}$. By the basic property of TV distance, we have
$$\Pr_{f,n}(A)-\Pr_{g,n}(A)\le \TV(P_{f,n}\|P_{g,n}).$$
Since the algorithm has sub-polynomial regret, for every $p>0$ we must have $\Pr_{f,n}(A)\ge 1-\bigO(1/n^{1-p})$ and $\Pr_{g,n}(A)\le \bigO(1/n^{1-p})$ (otherwise the algorithm incurs $\bigO(n^p)$ regret on either $f$ or $g$). Consequently, 
\begin{align}\label{equ:lb-sketch-1}
	\forall p>0,\TV(P_{f,n}\|P_{g,n})\ge 1-\bigO(1/n^{1-p}).
\end{align}
This states that the observations from the two instances should be distinguishable, although the TV distance is complex and hard to deal with (especially when the algorithm chooses its decision sequence adaptively).
So we translate the requirement from TV distance to a more localized quantity. First, we invoke a extended version of Pinsker's inequality (see \citet{sampson1992nonparametric})
\begin{align}
	\TV(P_{f,n}\|P_{g,n})\le 1-\exp(-\KL(P_{f,n}\|P_{g,n}))/2.
\end{align}
Combining with Eq.~\eqref{equ:lb-sketch-1} we get $\KL(P_{f,n}\|P_{g,n})\ge (1+o(1))\ln n.$ Let the random variable $\pi_i$ be the decision of the algorithm in round $i$. The chain rule of KL divergence shows
\begin{align*}
	(1+o(1))\ln n\le \KL(P_{f,n}\|P_{g,n})=\E_{f,n}\bigg[\sum_{i=1}^{n}\KL(f[\pi_i]\|g[\pi_i])\bigg]=\sum_{\pi}\E_{f,n}[N_\pi]\KL(f[\pi]\|g[\pi]).
\end{align*}
Now consider the vector $w\in\R^{|\Pi|}_+$ where $w_\pi=\E_f[N_\pi]/((1+o(1))\ln n)$. Based on the derivation above, we can verify that $w$ is a valid solution to $\comp(f,n),$ and therefore
\begin{align}
	\reg_{f,n}=\sum_{\pi}\E_f[N_\pi]\Delta(f,\pi)=\sum_{\pi}w_\pi\Delta(f,\pi)(1+o(1))\ln n\ge \comp(f,n)(1+o(1))\ln n.
\end{align}
Then the final result is proved by the fact that $\comp(f)=\lim_{n\to\infty}\comp(f,n)$.

\subsection{Proof Sketch of Theorem~\ref{thm:main-finite}}\label{sec:pf-upper-bound}
In the following, we discuss main lemmas and their proof sketches for the three steps of Alg.~\ref{alg:main-finite}.

\paragraph{Step 1: Initialization.} In this step we show that the max likelihood estimation can find the exact instance (i.e., $\initf=\truef$) with probability at least $1-1/\ln n$ and negligible regret. Note that the failure probability is not small enough to directly commit to the optimal decision of $\initf$ (i.e., play $\pi^\star(\initf)$ forever), which would incur linear regret when $\initf\neq\truef$.
Formally speaking, the main lemma for Step 1 is stated as follows, whose proof is deterred to Appendix~\ref{app:pf-main-init-finite}.
\begin{lemma}\label{lem:main-init-finite}
	Under Condition~\ref{cond:uniform-convergence}, with probability at least $1-1/\ln n$ we get $\initf=\truef.$ In addition, the regret of Step 1 is upper bounded by $\bigO(\frac{\ln n}{\ln\ln n}).$
\end{lemma}
Lemma~\ref{lem:main-init-finite} is not surprising since the MLE only requires $\bigO(\log(1/\delta))$ samples to reach a failure probability $\delta$ in general \citep[Section 7]{geer2000empirical}. In Step 1, we require $1/\ln n$ failure probability but use $(|\Pi|\ln n)/(\ln\ln n)=\Omega(\ln\ln n)$ samples, so the result is expected for large enough $n$.

\paragraph{Step 2: Identification.} In the identification step, we boost the failure probability to $1/n$ using the log-likelihood test.
To this end, we compute the optimal list of decisions $w=\{\pi_1,\cdots,\pi_m\}$ that distinguishes $\initf$ by solving $\comp(\initf,(\ln\ln n)^{1/4})$ (Line~\ref{line:3} of Alg.~\ref{alg:main-finite}). The choice of $(\ln\ln n)^{1/4}$ is not critical, and could be replaced by any smaller quantity that approaches infinity as $n\to\infty$. Then we run the log-likelihood ratio test using the observations collected by executing the list of decision $w$, and achieve the following:
\begin{enumerate}[label=(\alph*)]
	\item when the true instance $\truef$ and the estimation $\initf$ have different optimal decisions,
	accept $\initf$ with probability at most $1/n$;
	\item when $\initf=\truef$, accept $\initf$ with probability at least $1-1/\ln n$.
\end{enumerate}
In Section~\ref{sec:hypothesis-test}, we will discuss the log-likelihood ratio test in detail. Note that the regret after we falsely reject the true instance is $\bigO(\ln n)$ (by the regret bound of UCB algorithm), so we only require a $1/\ln n$ failure probability for (b) because then it leads to a $\bigO(\ln n)/\ln n=\bigO(1)$ expected regret.
The main lemma for Step 2 is stated as follows, and its proof is deferred to Appendix~\ref{app:pf-main-ident-finite}
\begin{lemma}\label{lem:main-ident-finite}
	Under Condition~\ref{cond:uniform-convergence}, for any finite hypothesis $\calF$, for large enough $n$ the following holds:
	\begin{enumerate}[label=(\alph*)]
		\item conditioned on the event $\pi^\star(\initf)\neq \pi^\star(f^\star)$, $\acc^\initf$ is true with probability at most $1/n$;
		\item conditioned on the event $\initf=\truef$, $\acc^\initf$ is true with probability at least $1-1/\ln n$;
		\item conditioned on the event $\initf=\truef$, the expected regret of Step 2 is upper bounded by $$\(\comp(f^\star,(\ln\ln n)^{1/4})+o(1)\)\ln n.$$ Otherwise, the expected regret of Step 2 is upper bounded by $\bigO(\ln n\ln\ln n).$
	\end{enumerate}
\end{lemma}

\paragraph{Step 3: Exploitation.} Finally, in Step 3 we either commits to the optimal decision $\pi^\star(\initf)$ when the estimation $\initf$ is accepted, or run a standard UCB algorithm with $\bigO(\ln n)$ regret \citep{lattimore2020bandit}. Combining Lemma~\ref{lem:main-init-finite} and Lemma~\ref{lem:main-ident-finite} we can prove that
\begin{enumerate}
	\item the probability of Step 3 incurring a $\bigO(n)$ regret is at most $1/n$, and
	\item the probability of Step 3 incurring a $\bigO(\ln n)$ regret is at most $1/\ln n$.
\end{enumerate}
As a result, the expected regret in Step 3 is $\bigO(1)$ and negligible .
Finally Theorem~\ref{thm:main-finite} is proved by stitching the three steps together, and deferred to Appendix~\ref{app:pf-main-finite}.

\subsection{The Log-Likelihood Ratio Test}\label{sec:hypothesis-test}
The goal of the log-likelihood ratio test is to boost the confidence of our initial estimation $\initf$ to $1-1/n$, so that the algorithm can safely commit to its optimal decision in Step 3. In other words, the test should reject a wrong estimation but also accept a correct one.

Formally speaking, in the test we observe a list of observations $\{o_1,\cdots,o_m\}$ collected by the list of decision $\{\pi_1,\cdots,\pi_m\}$ on the \emph{true instance} $\truef$, and achieve the following two goals simultaneously,
\begin{enumerate}[label=(\alph*)]
	\item when $\initf$ and $\truef$ are sufficiently different (i.e., their KL divergence is large in the sense that $\sum_{i=1}^{m}\KL(\initf[\pi_i]\|\truef[\pi_i])\ge (1+o(1))\ln n$), accept $\initf$ with probability at most $1/ n$, and
	\item when $\initf=\truef$, accept $\initf$ with probability at least $1-1/\ln n$.
\end{enumerate}
We prove that the log-likelihood ratio test with proper parameters achieves (a) and (b) under Condition~\ref{cond:uniform-convergence} in the next lemma, whose proof is deferred to Appendix~\ref{app:pf-acc-main}.
\begin{lemma}\label{lem:acc-main}
	Given two sequences of distributions $P=\{P_i\}_{i=1}^{m}$ and $Q=\{Q_i\}_{i=1}^{m}$, and a sequence of independent random variables $o=\{\ob_i\}_{i=1}^{m}$. For any fixed $\lambda>0,c>0$, and
	$
		\beta=\frac{1}{m}\sum_{i=1}^{m}D_{1-\lambda}(P_i\|Q_i),
	$
	the test event $$\acc=\mathbb{I}\bigg[\sum_{i=1}^{m}\ln \frac{P_i(\ob_i)}{Q_i(\ob_i)}\ge c\bigg]$$ satisfies
	\begin{align}
		&\Pr_{o\sim Q}(\acc)\le \exp(-c),\label{equ:acc-1}\\
		&\Pr_{o\sim P}(\acc)\ge 1-\exp(-\lambda(m\beta-c)).\label{equ:acc-2}
	\end{align}
\end{lemma}
To use Lemma~\ref{lem:acc-main} in the proof of Lemma~\ref{lem:main-ident}, we set $c=\ln n$ and $\lambda=\poly(\ln\ln n)$. Note that the choice of $w$ in Line~\ref{line:3} of Alg.~\ref{alg:main-finite} and Condition~\ref{cond:uniform-convergence} implies $m\beta =\sum_{i=1}^{m}D_{1-\lambda}(P_i\|Q_i)\gtrapprox \sum_{i=1}^{m}\KL(\initf[\pi_i]\|\truef[\pi_i])=(1+o(1))\ln n$ when $\initf,\truef$ have different optimal policies. So the conclusion of this lemma matches the first two items in Lemma~\ref{lem:main-ident-finite}.

Lemma~\ref{lem:acc-main} is closely related to the Chernoff-Stein lemma (see \citet{chernoff1959sequential} and \citet[Theorem 4.11]{mao2021hypothesis}). The failure probability of Eq.~\eqref{equ:acc-2} in classical Chernoff-Stein lemma is a constant, while here it decreases with $m$.  The proof of Eq.~\eqref{equ:acc-1} is the same as in Chernoff-Stein lemma, and proof of Eq.~\eqref{equ:acc-2} only requires an one-sided concentration of the empirical log-likelihood ratio.
Indeed, the expectation of empirical log-likelihood ratio can be lower bounded by
\begin{align}
	\E_P\bigg[\sum_{i=1}^{m}\ln\frac{P_i(o_i)}{Q_i(o_i)}\bigg]=\sum_{i=1}^{m}\KL\(P_i\|Q_i\)\ge \sum_{i=1}^{m}D_{1-\lambda}\(P_i\|Q_i\)= m\beta.
\end{align}
So $\Pr_P(\acc)$ is the probability that the empirical log-likelihood ratio is (approximately) larger than its expectation. We also note that the concentration is non-trivial because we do not make assumptions on the boundedness on the tail of $Q_i$.
	\section{Extensions to Infinite Hypothesis Class}\label{sec:infinite}
Now we extend our analysis to infinite hypothesis settings, and instantiate our results on tabular RL.
Our results in the infinite hypothesis case need two additional conditions. The first condition requires an upper bound on covering number of the hypothesis, and is used to prove a infinite-hypothesis version of Lemma~\ref{lem:main-ident-finite}. Formally speaking, for any $f,g\in \calF$, define their distance as 
\begin{align}
	d(f,g)=\sup_{\pi\in\Pi,\ob}\abs{f[\pi](\ob)-g[\pi](\ob)}.
\end{align}
An $\epsilon$ (proper) covering of $\calF$ is a subset $\calC\subseteq \calF$, such that for any $g\in\calF$, there exists $g'\in\calC$ with $d(g,g')\le \epsilon.$ The covering number $\calN(\calF,\epsilon)$ is the size of the minimum $\epsilon$ covering of $\calF$.
\begin{cond}\label{cond:covering-number}
	There exists constant $\const$ that depends on $\calF$ such that $\ln\calN(\calF,\epsilon)\le \bigO(\const\ln(1/\epsilon))$ for every $\epsilon>0.$ In addition, the base measure of the probability space has a finite volume $\vol<\infty.$
\end{cond}

The second condition upper bounds the distance of two instances by a polynomial of their KL divergence, and is used to prove a stronger version of Lemma~\ref{lem:main-init-finite}.
\begin{cond}\label{cond:TV-vs-inf}
	There exists a constant $\const_{\rm min}>0$ (which may depend on $\truef$) such that for all $\pi\in\Pi$ and $\ob\in\supp(\truef[\pi])$
	$
	\truef[\pi](\ob)> \const_{\rm min}.
	$
	In addition, there exists a constant $\iota(\truef)>0$ that only depends on $\truef$ and $\const_5>0$, such that for all $f\in\calF,\pi\in\Pi$, when $\KL(\truef[\pi]\|f[\pi])\le 1$
	$$
		\|\truef[\pi]-f[\pi]\|_\infty\le \iota(\truef)\KL(\truef[\pi]\|f[\pi])^{\const_5}.
	$$
\end{cond}

A lot of interactive decision making problems satisfy Conditions~\ref{cond:covering-number},~\ref{cond:TV-vs-inf}, such as multi-armed bandits and linear bandits with Bernoulli reward or truncated Gaussian reward. 
For bandit problems, Conditions~\ref{cond:covering-number} and ~\ref{cond:TV-vs-inf} only depend on the noise of the reward function and can be verified easily.

For tabular RL, an observation $\ob$ denotes a trajectory $(s_1,a_1,r_1,\cdots,s_H,a_H,r_H)$ where the state-action pairs $(s_h,a_h)$ are discrete random variables but the rewards $r_h$ are continuous. A decision $\pi:\calS\to\calA$ is a mapping from the state space to action space, so $|\Pi|=|\calA|^{|\calS|}<\infty.$ In Appendix~\ref{app:rl}, we formally define the tabular RL problems where the reward $r_h$ given $(s_h,a_h)$ follows a truncated Gaussian distribution, and
prove that Conditions~\ref{cond:uniform-convergence}-\ref{cond:TV-vs-inf} are satisfied as stated in the following theorem.
\begin{theorem}\label{thm:rl-condition}
	Let $\calF$ be the family of tabular RL with truncated Gaussian reward and unique optimal policies. Then Conditions~\ref{cond:uniform-convergence},~\ref{cond:covering-number},~\ref{cond:TV-vs-inf} holds simultaneously.
\end{theorem}
For tabular RL with truncated Gaussian reward, the observation $\ob$ is a mixture of discrete random variables (i.e., the states and actions $s_h,a_h$) and continuous random variables (i.e., the rewards $r_h$). To prove Conditions~\ref{cond:uniform-convergence}-\ref{cond:TV-vs-inf}, we deal with the discrete and continuous part of the observation separately. We also have flexibility in the reward distribution, e.g., our proof technique can deal with other distributions such as truncated Laplace distribution.

We present these unified conditions for bounded random variables, but Conditions~\ref{cond:covering-number} and~\ref{cond:TV-vs-inf} do not hold for unbounded random variables because the base measure has an infinite volume (which contradicts to Condition~\ref{cond:covering-number}), and the density of the observation cannot be lower bounded (which contradicts to Condition~\ref{cond:TV-vs-inf}). However, we can still prove the same result using a slightly different approach (see Appendix~\ref{app:rl-Gaussian}).

With Conditions~\ref{cond:uniform-convergence}-\ref{cond:TV-vs-inf}, we can prove our main results for infinite hypothesis class. The asymptotic regret of Alg.~\ref{alg:main-finite} in this case still matches the lower bound exactly. The proof of Theorem~\ref{thm:main} is deferred to Appendix~\ref{app:pf-main}.
\begin{theorem}\label{thm:main}
	Suppose $\calF$ is an infinite hypothesis class that satisfies  Conditions~\ref{cond:uniform-convergence}-\ref{cond:TV-vs-inf}, the regret of Alg.~\ref{alg:main-finite} satisfies
	\begin{align}
		\limsup_{n\to\infty}\frac{\reg_{\truef,n}}{\ln n}\le \comp(\truef).
	\end{align}
\end{theorem}
As a corollary of Theorem~\ref{thm:rl-condition} and Theorem~\ref{thm:main}, our algorithm strictly improves upon previous best gap-dependent bounds $\bigO(\sum_{s,a:\Delta(s,a)>0}1/\Delta(s,a))$ in \citet{xu2021fine} and \citet{simchowitz2019non} because the gap-dependent bounds are not tight for many instances \citep{tirinzoni2021fully}.

In the following we discuss the challenges when extending our analysis to infinite hypothesis settings.

\paragraph{Covering Number.} 
With infinite hypothesis, we need to accept an accurate estimation even when there are infinitely many other choices. Recall that the accept event is \begin{align}
	\acc^{\initf}&=\ind{\forall g\in \calF\text{ and }\pi^\star(g)\neq \pi^\star(\initf), \sum_{i=1}^{m}\ln\frac{\initf[\pi_i](\ob_i)}{g[\pi_i](\ob_i)}\ge \ln n}.
\end{align}
So informally speaking, we need to show that with high probability, 
\begin{align}
	\forall g\in \calF\text{ and }\pi^\star(g)\neq \pi^\star(\initf),\quad \sum_{i=1}^{m}\ln \frac{\initf[\pi_i](\ob_i)}{g[\pi_i](\ob_i)}\gtrsim (D^w_{1-\lambda}(\initf\|g)-\epsilon)m\ge \ln n,
\end{align}
which is essentially an uniform concentration inequality as discussed in Section~\ref{sec:hypothesis-test}. So we resort to a covering number argument.
The standard covering argument is not directly suitable in this case --- even if $d(g,g')$ is small, it's still possible that $\ln(f[\pi](\ob)/g[\pi](\ob))$ is very different from $\ln(f[\pi](\ob)/g'[\pi](\ob))$ (especially when $g[\pi](\ob)$ is very close to 0). Instead, we consider the distribution with density $\hat{g}[\pi](\ob)\defeq(g'[\pi](\ob)+\epsilon)/Z$ where $Z$ is the normalization factor, and only prove a one-sided covering (that is, $\ln(f[\pi](\ob)/g[\pi](\ob))\ge \ln(f[\pi](\ob)/\hat{g}[\pi](\ob))-\bigO(\epsilon)$). We state and prove the uniform concentration in Appendix~\ref{app:uniform-concentration}.

\paragraph{Initialization.} With infinite hypothesis, we cannot hope to recover the true instance $\truef$ exactly --- some instances can be arbitrarily close to $\truef$ and thus indistinguishable. Instead, we prove that the estimation in Step 1 satisfies $\sup_{\pi\in\Pi}\KL(\truef[\pi]\|\initf[\pi])\le \poly(\frac{\ln\ln n}{\ln n})$. The main lemma of Step 1 in the infinite hypothesis class case is stated in Lemma~\ref{lem:main-init}.
	\section{Conclusion}\label{sec:conclusion}
In this paper, we design instance-optimal algorithms for general interactive decision making problems with finite decisions. As an instantiation, our algorithm is the first instance-optimal algorithm for tabular MDPs. For future works, we raise the following open questions.
\begin{enumerate}
	\item To implement Alg.~\ref{alg:main-finite}, we need to solve $\comp(f,(\ln \ln n)^{1/4}/2)$, which is a linear programming with $|\Pi|$ variables and infinitely many constraints. However, $|\Pi|$ is exponentially large for tabular MDPs. Can we compute this optimization problem efficiently for tabular MDPs?
	\item Although our algorithm is asymptotically optimal, the lower order terms may dominate the regret unless $n$ is very large. Can we design non-asymptotic instance optimal algorithms?
\end{enumerate}
	
	\subsection*{Acknowledgment}
	The authors would like to thank Yuanhao Wang, Jason D. Lee for helpful discussions, and the support from JD.com.
	
	\bibliographystyle{plainnat}
	\bibliography{all.bib}
	
	\newpage
	\appendix
	\section*{List of Appendices}
	\startcontents[sections]
	\printcontents[sections]{l}{1}{\setcounter{tocdepth}{2}}
	\newpage
	
	\section{Proofs for Finite Hypothesis Class}
In this section we show the missing proofs in Section~\ref{sec:proof-sketches}.
\subsection{Proof of Lemma~\ref{lem:main-init-finite}}\label{app:pf-main-init-finite}
In this section, we prove Lemma~\ref{lem:main-init-finite}. 
\begin{proof}[Proof of Lemma~\ref{lem:main-init-finite}]
	Recall that $\minit=|\Pi|\lceil\frac{\ln n}{\ln\ln n}\rceil$. Let $w=\{\hat\pi_i\}_{i=1}^{\minit}$ be the sequence of decisions in the initialization step of Alg.~\ref{alg:main-finite}, and $\{\hat{\ob}_i\}_{i=1}^{\minit}$ the corresponding observations. Define
	\begin{align}
		\rho=\min_{g\in\calF,g\neq \truef}\max_{\pi\in\Pi}D_{1/2}(\truef[\pi]\|g[\pi]).
	\end{align}
	For finite hypothesis we have $\rho>0.$ Then Lemma~\ref{lem:individual-concentration}, for any $\epsilon>0$ we get
	\begin{align}
		\Pr\(\frac{1}{\minit}\sum_{i=1}^{\minit}\ln \frac{\truef[\hat\pi_i](\hat{\ob}_i)}{g[\hat\pi_i](\hat{\ob}_i)}\ge \frac{1}{\minit}\sum_{i=1}^{\minit}D_{1/2}(\truef[\hat\pi_i]\|g[\hat\pi_i])-\epsilon\)\ge 1-\exp(-\minit\epsilon/2).
	\end{align}
	Let $\epsilon=\frac{\rho}{2|\Pi|}$. By definition of $\rho$ we get, for all $g\in\calF\setminus\{\truef\}$
	\begin{align}
		&\Pr\(\frac{1}{\minit}\sum_{i=1}^{\minit}\ln \frac{\truef[\hat\pi_i](\hat{\ob}_i)}{g[\hat\pi_i](\hat{\ob}_i)}\ge \frac{\rho}{2|\Pi|}\)\\
		\ge\;&\Pr\(\frac{1}{\minit}\sum_{i=1}^{\minit}\ln \frac{\truef[\hat\pi_i](\hat{\ob}_i)}{g[\hat\pi_i](\hat{\ob}_i)}\ge \frac{1}{\minit}\sum_{i=1}^{\minit}D_{1/2}(\truef[\hat\pi_i]\|g[\hat\pi_i])-\frac{\rho}{2|\Pi|}\)\\
		\ge\;& 1-\exp\(-\frac{\rho\ln n}{4\ln\ln n}\).
	\end{align}
	By union bound, with probability at least $1-|\calF|\exp\(-\frac{\rho\ln n}{4\ln\ln n}\)$ we have
	\begin{align}
		\forall g\in\calF\setminus\{\truef\},\quad \sum_{i=1}^{\minit}\ln \frac{\truef[\hat\pi_i](\hat{\ob}_i)}{g[\hat\pi_i](\hat{\ob}_i)}>0= 
		\sum_{i=1}^{\minit}\ln \frac{\truef[\hat\pi_i](\hat{\ob}_i)}{\truef[\hat\pi_i](\hat{\ob}_i)},
	\end{align}
	which implies that 
	\begin{align}\label{equ:pf-minitf-1}
		\forall g\in\calF\setminus\{\truef\},\quad \sum_{i=1}^{\minit}\ln g[\hat\pi_i](\hat{\ob}_i)< 
		\sum_{i=1}^{\minit}\ln \truef[\hat\pi_i](\hat{\ob}_i).
	\end{align}
	Recalling that $\initf=\argmax_{f\in\calF}\sum_{i=1}^{\minit}\ln f[\hat\pi_i](\hat{\ob}_i)$, Eq.~\eqref{equ:pf-minitf-1} implies $\initf=\truef.$
	
	By algebraic manipulation, for large enough $n$ we have
	\begin{align}
		\exp\(-\frac{\rho\ln n}{4\ln\ln n}\)\le \exp(-\ln|\calF|-\ln\ln n)=\frac{1}{|\calF|\ln n}.
	\end{align}
	As a result, for large enough $n$ we get $\Pr(\initf=\truef)\ge 1-1/\ln n.$ In addition, the regret of Step 1 is upper bounded by $\bigO(\minit)=\bigO(\frac{\ln n}{\ln\ln n}).$
\end{proof}

\subsection{Proof of Lemma~\ref{lem:main-ident-finite}} \label{app:pf-main-ident-finite}
In this section, we prove Lemma~\ref{lem:main-ident-finite}. To this end, we need the following lemma.
\begin{lemma}\label{lem:renyi-large}
	Consider an instance $f\in\calF$ and $n>0$. Let $\delta=(\ln\ln n)^{-1/4}$ and $\epsilon=(\ln\ln n)^{-1}.$ Define 
	\begin{align}
		\lambda=\lambda_0\(4(\ln\ln n)^{3/4},\frac{1}{\ln\ln n},f\)\label{equ:pf-mif-0}
	\end{align}
	as the value that Condition~\ref{cond:uniform-convergence} holds with corresponding parameters. 
	
	Consider any $\hat{w}\in\R^{|\Pi|}_+$ such that $\|\hat{w}\|_\infty\le (\ln\ln n)^{1/4}$. 
	Let $w=\{\pi_i\}_{i=1}^{m}$ be a list of decisions where a decision $\pi$ occurs $\lceil ((1+\delta)\hat{w}_\pi+\delta)\ln n\rceil$ times for every $\pi\in\Pi$, and $m=\sum_\pi \lceil ((1+\delta)\hat{w}_\pi+\delta)\ln n\rceil$. 
	
	Define the set 
	$\calF(\hat{w},f)=\{g\in\calF:\sum_{\pi\in\Pi}\hat{w}_\pi\KL(f[\pi]\|g[\pi])\ge 1\}.$
	For any constant $c>0$, there exits $n_0>0$ such that for all $n>n_0$,
	\begin{align}
		D^w_{1-\lambda}(f\|g)\ge \frac{\ln n}{m}+c\epsilon,\quad \forall g\in\calF(\hat{w},f).
	\end{align}
	In addition, we have $\lambda^{-1}=O(\poly(\ln\ln n))$ and $m\ge \frac{|\Pi|\ln n}{(\ln\ln n)^{1/4}}.$
\end{lemma}
\begin{proof}
	To prove this lemma, we invoke Condition~\ref{cond:uniform-convergence} with proper parameters.
	
	First we bound the value of $m$. By the assumption of this lemma, we have $\|\hat{w}\|_\infty\le (\ln\ln n)^{1/4}$. Consequently, for large enough $n$ we get
	\begin{align}
		m=\sum_\pi \lceil ((1+\delta)\hat{w}_\pi+\delta)\ln n\rceil\le 2|\Pi|\ln n(\ln\ln n)^{1/4}.
	\end{align}
	On the other hand, 
	\begin{align}
		m=\sum_\pi \lceil ((1+\delta)\hat{w}_\pi+\delta)\ln n\rceil\ge |\Pi|\delta\ln n=\frac{|\Pi|\ln n}{(\ln\ln n)^{1/4}}.
	\end{align}
	Define $\alpha=\frac{\ln n}{m}+c\epsilon.$ Then for large enough $n$ we have $\alpha+\epsilon\le \frac{2}{|\Pi|}(\ln\ln n)^{1/4}.$
	We can also lower bound $\gamma=\frac{1}{m}\min_\pi \lceil ((1+\delta)\hat{w}_\pi+\delta)\ln n\rceil$. By the upper bound of $m$ and the fact that $\lceil ((1+\delta)\hat{w}_\pi+\delta)\ln n\rceil\ge \delta \ln n=\frac{\ln n}{(\ln\ln n)^{1/4}}$, we get $\gamma\ge \frac{1}{2|\Pi|(\ln\ln n)^{1/2}}.$
	
	We invoke Condition~\ref{cond:uniform-convergence} with parameters $((\alpha+\epsilon)/\gamma, \epsilon,f)$. Note that $\lambda$ defined in Eq.~\eqref{equ:pf-mif-0} satisfies $\lambda\le \lambda_0((\alpha+\epsilon)/\gamma, \epsilon,f)$ because $(\alpha+\epsilon)/\gamma\le 4(\ln\ln n)^{3/4}$. Therefore we get
	\begin{align}\label{equ:pf-mif-1}
		D_{1-\lambda}(f[\pi]\|g[\pi])\ge \min\{\KL(f[\pi]\|g[\pi])-\epsilon,(\alpha+\epsilon)/\gamma\},\quad\forall g\in \calF,\pi\in\Pi.
	\end{align}
	In the following, we prove that $D^w_{1-\lambda}(f\|g)\ge \alpha$ for all $g\in\calF(\hat{w},f)$. First of all, for large enough $n$, for all $g\in\calF(\hat{w},f)$ we get
	\begin{align}
		&\KL^w(f\|g)=\frac{1}{m}\sum_{i=1}^{m}\KL(f[\pi_i]\|g[\pi_i])=\frac{1}{m}\sum_{\pi\in\Pi}\lceil ((1+\delta)\hat{w}_\pi+\delta)\ln n\rceil\KL(f[\pi]\|g[\pi])\\
		\ge&\; \frac{1}{m}\sum_{\pi\in\Pi}(1+\delta)(\ln n)\hat{w}_\pi\KL(f[\pi]\|g[\pi])\ge \frac{1}{m}(1+\delta)\ln n
		\ge\frac{1}{m}(\ln n+(c+1)m\epsilon),\label{equ:pf-mif-3}
	\end{align}
	where the last inequality comes from the fact that $\delta\ln n=\frac{\ln n}{(\ln\ln n)^{1/4}}\gtrsim 2(c+1)|\Pi|\frac{\ln n}{(\ln\ln n)^{3/4}}\ge (c+1)m\epsilon.$
	
	Now for a fixed $g\in\calF(\hat{w},f)$, consider the following two cases.
	
	\paragraph{Case 1: $\exists\bar{\pi}\in \Pi, \KL(f[\bar{\pi}]\|g[\bar{\pi}])\ge (\alpha+\epsilon)/\gamma.$} In this case we have
	\begin{align}
		&D^w_{1-\lambda}(f\|g)=\frac{1}{m}\sum_{i=1}^{m}D_{1-\lambda}(f[\pi_i]\|g[\pi_i])\\
		\ge\;& \frac{\lceil ((1+\delta)\hat{w}_{\bar{\pi}}+\delta)\ln n\rceil}{m}D_{1-\lambda}(f[\bar{\pi}]\|g[\bar{\pi}])\\
		\ge\;&\gamma D_{1-\lambda}(f[\bar{\pi}]\|g[\bar{\pi}])\ge \alpha,
	\end{align}
	where the last inequality comes from Eq.~\eqref{equ:pf-mif-1}.
	
	\paragraph{Case 2: $\forall \pi\in \Pi, \KL(f[\pi]\|g[\pi])\le (\alpha+\epsilon)/\gamma.$} In this case, Eq.~\eqref{equ:pf-mif-1} implies that 
	\begin{align}\label{equ:pf-mif-2}
		D_{1-\lambda}(f[\pi]\|g[\pi])\ge \KL(f[\pi]\|g[\pi])-\epsilon,\quad\forall \pi\in\Pi.
	\end{align}
	Consequently,
	\begin{align*}
		&D^w_{1-\lambda}(f\|g)=\frac{1}{m}\sum_{i=1}^{m}D_{1-\lambda}(f[\pi_i]\|g[\pi_i])
		\ge \frac{1}{m}\sum_{i=1}^{m}(\KL(f[\pi_i]\|g[\pi_i])-\epsilon)=\KL^w(f\|g)-\epsilon.
	\end{align*}
	By Eq.~\eqref{equ:pf-mif-3} we get $\KL^{w}(f\|g)\ge \alpha+\epsilon.$ Therefore $D^w_{1-\lambda}(f\|g)\ge \alpha.$
	
	Combining the two cases together we get the desired result. The lower bound of $\lambda^{-1}$ follows directly from Condition~\ref{cond:uniform-convergence}.
\end{proof}

Now we are ready to prove Lemma~\ref{lem:main-ident-finite}.
\begin{proof}[Proof of Lemma~\ref{lem:main-ident-finite}]
	We prove the four items in Lemma~\ref{lem:main-ident-finite} separately. We will invoke Lemma~\ref{lem:renyi-large} and Lemma~\ref{lem:acc-main} in the proof.
	Following Lemma~\ref{lem:renyi-large}, let $\delta=(\ln\ln n)^{1/4}, \epsilon=(\ln\ln n)^{-1}$ and 
	\begin{align}
		\lambda=\lambda_0\(4(\ln\ln n)^{3/4},\frac{1}{\ln\ln n},f\).
	\end{align}
	Recall that $\cset(f)=\{g\in\calF:\pi^\star(g)\neq\pi^\star(f)\}.$ 
	
	\paragraph{Proof of item (a).} In this case we have $\truef\in\cset(\initf).$
	By Lemma~\ref{lem:acc-main} we get
	\begin{align}
		&\Pr_{\truef}\(\acc^{\initf}\)=\Pr_{\truef}\(\forall g\in \cset(\initf),\sum_{i=1}^{m}\ln\frac{\initf[\pi_i](o_i)}{g[\pi_i](o_i)}\ge \ln n\)\\
		\le&\;\Pr_{\truef}\(\sum_{i=1}^{m}\ln\frac{\initf[\pi_i](o_i)}{\truef[\pi_i](o_i)}\ge \ln n\)
		\le\exp(-\ln n)=1/n.
	\end{align}
	
	\paragraph{Proof of item (b).}  Recall that in this case we have $\initf=\truef$. Since $\hat{w}$ is the solution to $\comp(\initf,(\ln\ln n)^{1/4})$ (Line~\ref{line:3} of Alg.~\ref{alg:main-finite}), we have $\sum_{\pi\in\Pi}\hat{w}_\pi\KL(\initf[\pi]\|g[\pi])\ge 1$ for all $g\in\cset(\initf)$. Recall that $w$ is the list of decisions computed by Line~\ref{line:4} of Alg.~\ref{alg:main-finite}.
	By Lemma~\ref{lem:renyi-large} we get
	\begin{align}
		D^w_{1-\lambda}(f\|g)\ge \frac{\ln n}{m}+\epsilon,\quad\forall g\in\cset(\initf).
	\end{align}
	Let $\beta=\frac{\ln n}{m}+\epsilon$. By Lemma~\ref{lem:acc-main}, for every $g\in\cset(\initf)$ we have
	\begin{align}
		\Pr_{\truef}\(\sum_{i=1}^{m}\ln\frac{\truef[\pi_i](o_i)}{g[\pi_i](o_i)}\ge \ln n\)\ge 1-\exp(-m\lambda\epsilon).
	\end{align}
	Lemma~\ref{lem:renyi-large} also yields $\lambda^{-1}\le \poly(\ln\ln n)$ and $m\ge \frac{|\Pi|\ln n}{(\ln\ln n)^{1/4}}$. Therefore $m\lambda\epsilon\ge \frac{\ln n}{\poly(\ln\ln n)}$. Consequently, for large enough $n$ we have
	\begin{align}
		\exp(-m\lambda\epsilon)\le \frac{\ln n}{|\calF|}.
	\end{align}
	Applying union bound, under the event $\ind{\initf=\truef}$ we get
	\begin{align}
		\Pr_{\truef}\(\acc^{\initf}\)=\Pr_{\truef}\(\forall g\in \cset(\initf),\sum_{i=1}^{m}\ln\frac{\initf[\pi_i](o_i)}{g[\pi_i](o_i)}\ge \ln n\)\ge 1-\ln n.
	\end{align}

	\paragraph{Proof of item (c).} Recall that $\Delta(f,\pi)$ is the sub-optimality gap of decision $\pi$ on instance $f$, and $\Deltamax(f)=\max_\pi \Delta(f,\pi)$ is the maximum decision gap of instance $f$.
	Since $\hat{w}$ is the solution to $\comp(\initf,(\ln\ln n)^{1/4})$, we have $\|\hat{w}\|_\infty\le (\ln\ln n)^{1/4}.$ As a result, the regret of Step 2 is upper bounded by 
	\begin{align}
		\sum_{\pi}\lceil ((1+\delta)\hat{w}_\pi+\delta)\ln n\rceil \Deltamax(\truef)\lesssim |\Pi|\ln n(\ln\ln n)^{1/4},
	\end{align}
	which proves the second part of (c). For the first part, when $\initf=\truef$ we have
	\begin{align}
		&\sum_{\pi}\lceil ((1+\delta)\hat{w}_\pi+\delta)\ln n\rceil \Delta(\truef,\pi)\\
		\le\;&\sum_{\pi} (1+\delta)\hat{w}_\pi(\ln n) \Delta(\truef,\pi)+ |\Pi|\Deltamax(\truef)(1+\delta\ln n)\\
		=\;&((1+\delta)\comp(\truef,(\ln\ln n)^{1/4})+o(1))\ln n.
	\end{align}
	By Lemma~\ref{lem:constant-solution}, $\comp(\truef,(\ln\ln n)^{1/4})=\bigO(1)$. As a result,
	\begin{align}
		&((1+\delta)\comp(\truef,(\ln\ln n)^{1/4})+o(1))\ln n\le (\comp(\truef,(\ln\ln n)^{1/4})+o(1))\ln n.
	\end{align}
\end{proof}

\subsection{Proof of Lemma~\ref{lem:acc-main}}\label{app:pf-acc-main}
In this section, we prove Lemma~\ref{lem:acc-main}. To this end, we need the following concentration inequality.
\begin{lemma}\label{lem:individual-concentration} Consider two sequences of distributions $P=\{P_i\}_{i=1}^{m}$ and $Q=\{Q_i\}_{i=1}^{m}$, and a sequence of random variables $\ob=\{\ob_i\}_{i=1}^{m}$ independently drawn from $P$. For any $\lambda\in (0,1)$ and $\epsilon>0$ we have
	\begin{align}
		\Pr_{\ob\sim P}\(\frac{1}{m}\sum_{i=1}^{m}\ln \frac{P_i(\ob_i)}{Q_i(\ob_i)}\ge \frac{1}{m}\sum_{i=1}^{m}D_{1-\lambda}(P_i\|Q_i)-\epsilon\)\ge 1-\exp(-m\lambda\epsilon).
	\end{align}
\end{lemma}
\begin{proof}[Proof of Lemma~\ref{lem:individual-concentration}]
	We prove Lemma~\ref{lem:individual-concentration} by moment method. Consider any $\lambda\in(0,1), \epsilon>0$ and  let $\beta=\frac{1}{m}\sum_{i=1}^{m}D_{1-\lambda}(P_i\|Q_i).$ Then we have
	\begin{align}
		&\Pr_{\ob\sim P}\(\sum_{i=1}^{m}\ln \frac{P_i(\ob_i)}{Q_i(\ob_i)}\le m(\beta-\epsilon)\)\\
		=\;&\Pr_{\ob\sim P}\(\sum_{i=1}^{m}\ln \frac{Q_i(\ob_i)}{P_i(\ob_i)}\ge -m(\beta-\epsilon)\)\\
		=\;&\Pr_{\ob\sim P}\(\exp\(\lambda\sum_{i=1}^{m}\ln \frac{Q_i(\ob_i)}{P_i(\ob_i)}\)\ge \exp(-\lambda m(\beta-\epsilon))\)\\
		\le\;& 
		\exp(\lambda m(\beta-\epsilon))\E_{\ob\sim P}\[\exp\(\lambda\sum_{i=1}^{m}\ln \frac{Q_i(\ob_i)}{P_i(\ob_i)}\)\]\label{equ:ic-1}\\
		\le\;& 
		\exp(\lambda m(\beta-\epsilon))\prod_{i=1}^{m}\E_{\ob_i\sim P_i}\[\(Q_i(\ob_i)^{\lambda}P_i(\ob_i)^{-\lambda}\)\] \\
		\le\;& 
		\exp(\lambda m(\beta-\epsilon))\prod_{i=1}^{m}\(\int_{\ob}\(Q_i(\ob)^{\lambda}P_i(\ob)^{1-\lambda}\)\dd o\) \\
		\le\;& 
		\exp(\lambda m(\beta-\epsilon))\prod_{i=1}^{m}\exp((\lambda-1)D_{\lambda}(Q_i\|P_i)).
	\end{align}
	where Eq.~\eqref{equ:ic-1} follows from Markov inequality.
	By \citet[Proposition 2]{van2014renyi} we get $D_{\lambda}(Q\|P)=\frac{\lambda}{1-\lambda}D_{1-\lambda}(P\|Q)$ for any distributions $P,Q$. As a result,
	\begin{align}
		&\exp(\lambda m(\beta-\epsilon))\prod_{i=1}^{m}\exp((\lambda-1)D_{\lambda}(Q_i\|P_i))\\
		=\;& 
		\exp(\lambda m(\beta-\epsilon))\prod_{i=1}^{m}\exp(-\lambda D_{1-\lambda}(P_i\| Q_i))\\
		=\;&\exp(\lambda m(\beta-\epsilon))\exp\(-\lambda m\beta\) \\
		=\;&\exp\(-m\lambda\epsilon\).
	\end{align}
\end{proof}
Now we are ready to prove Lemma~\ref{lem:acc-main}.
\begin{proof}[Proof of Lemma~\ref{lem:acc-main}]
	First we prove Eq.~\eqref{equ:acc-1}. By the moment method, for any $c>0$
	\begin{align}
		&\Pr_Q(\acc)=\Pr_Q\(\sum_{i=1}^{m}\ln \frac{P_i(\ob_i)}{Q_i(\ob_i)}\ge c\)\\
		=\;&\Pr_Q\(\exp\(\sum_{i=1}^{m}\ln \frac{P_i(\ob_i)}{Q_i(\ob_i)}\)\ge \exp(c)\)\\
		\le\;&\exp(-c)\E_Q\[\prod_{i=1}^{m}\frac{P_i(\ob_i)}{Q_i(\ob_i)}\]\tag{Markov inequality}\\
		\le\;&\exp(-c)\prod_{i=1}^{m}\E_{\ob_i\sim Q_i}\[\frac{P_i(\ob_i)}{Q_i(\ob_i)}\]\tag{independence of $\ob_i$'s}\\
		=\;&\exp(-c),
	\end{align}
	where the last line comes from the fact that both $P_i,Q_i$ are valid distributions.
	For Eq.~\eqref{equ:acc-2}, we can directly invoke Lemma~\ref{lem:individual-concentration} with $\epsilon=\beta-c/m$.
\end{proof}

\subsection{Proof of Theorem~\ref{thm:main-finite}}\label{app:pf-main-finite}
We prove Theorem~\ref{thm:main-finite} by stitching Lemma~\ref{lem:main-init-finite} and Lemma~\ref{lem:main-ident-finite} together. 
\begin{proof}[Proof of Theorem~\ref{thm:main-finite}]
	We upper bound the regret of Alg.~\ref{alg:main-finite} by discussing the regret under following events separately. Let $\reg_{\rm Step 1},\reg_{\rm Step 2},$ and $\reg_{\rm Step 3}$ be the regret incurred in Step 1, 2, and 3 respectively. Let $\init=\ind{\initf=\truef}$ be the event that the initial estimation is accurate.
	
	\paragraph{Regret of Step 1:} By Lemma~\ref{lem:main-init-finite} we have, $$\limsup_{n\to\infty}\frac{1}{\ln n}\E[\reg_{\rm Step 1}]\le 0.$$
	
	\paragraph{Regret of Step 2.} By item (c) of Lemma~\ref{lem:main-ident-finite} and Lemma~\ref{lem:main-init-finite},
	\begin{align}
		&\E\[\reg_{\rm Step 2}\]= \E\[\reg_{\rm Step 2}\mid \init\]\Pr(\init)+\E\[\reg_{\rm Step 2}\mid \init^c\]\Pr(\init^c)\\
		\le\;&(\comp(\truef,(\ln\ln n)^{1/4})+o(1))\ln(n)+\bigO(\ln n\ln\ln n)/\ln n.
	\end{align}
	As a result,
	\begin{align}
		&\limsup_{n\to\infty}\frac{1}{\ln n}\E\[\reg_{\rm Step 2}\]\le \limsup_{n\to\infty}\comp(\truef,(\ln\ln n)^{1/4})=\comp(\truef).
	\end{align}
	
	\paragraph{Regret of Step 3.} 
	First focus on the event $\init$. Under event $\acc^\initf\cap \init$, there is no regret in Step 3. On the other hand, by item~(b) of Lemma~\ref{lem:main-ident-finite} we have $\Pr(\init\cap(\acc^\initf)^c)\le 1/\ln n.$
	Since UCB gives logarithmic regret, we have
	\begin{align}
		\limsup_{n\to\infty}\frac{1}{\ln n}\E\[\ind{\init,(\acc^\initf)^c}\reg_{\rm Step 3}\] \le\limsup_{n\to\infty}\frac{\bigO(\ln n)}{\ln n}\Pr(\init\cap(\acc^\initf)^c)\le 0.
	\end{align}
	As a result,
	\begin{align}\label{equ:event2-4-finite}
		\limsup_{n\to\infty}\frac{1}{\ln n}\E\[\ind{\init}\reg_{\rm Step 3}\]\le 0.
	\end{align}
	
	Now we focus on the event $\init^c$. Let $\event_1=\{\pi^\star(\initf)=\pi^\star(\truef)\}.$ Under event $\acc^\initf\cap \event_1$ the algorithm has no regret in Step 3:
	\begin{align}
		&\E\[\ind{\init^c,\event_1,\acc^\initf}\reg_{\rm Step 3}\]=0.\label{equ:event2-1-finite}
	\end{align}
	On the other hand, consider the event $\acc^\initf\cap \event_1^c$. By item~(a) of  Lemma~\ref{lem:main-ident-finite} we have
	\begin{align}
		&\E\[\ind{\init^c,\event_1^c,\acc^\initf}\reg_{\rm Step 3}\]
		\le n\Deltamax\Pr\(\ind{\pi^\star(\truef)\neq \pi^\star(\initf),\acc^\initf}\)\le \frac{n\Deltamax}{n}\le \bigO(1).\label{equ:event2-2-finite}
	\end{align}
	Under the event $(\acc^\initf)^c$, Step 3 incurs logarithmic regret. As a result, 
	\begin{align}
		&\E\[\ind{\init^c,(\acc^\initf)^c}\reg_{\rm Step 3}\]
		\le \bigO(\ln n)\Pr\(\init^c\)\le \bigO(1).\label{equ:event2-3-finite}
	\end{align}
	Combining Eqs.~\eqref{equ:event2-1-finite},~\eqref{equ:event2-2-finite} and \eqref{equ:event2-3-finite} we get
	\begin{align}\label{equ:event2-5-finite}
		&\limsup_{n\to\infty}\frac{1}{\ln n}\E\[\ind{\init^c}\reg_{\rm Step 3}\]=0.
	\end{align}
	Therefore, combining Eqs.~\eqref{equ:event2-4-finite} and \eqref{equ:event2-5-finite},
	\begin{align}
		&\limsup_{n\to\infty}\frac{1}{\ln n}\E\[\reg_{\rm Step 3}\]=0.
	\end{align}
	
	Stitching the regrets from steps 1-3 together we prove Theorem~\ref{thm:main-finite}.
\end{proof}
	\section{Missing Proofs in Section~\ref{sec:lowerbound}}
In this section, we present the missing proofs in section~\ref{sec:lowerbound}.
\subsection{Proof of Lemma~\ref{lem:comp}}\label{app:pf-comp}
To prove Lemma~\ref{lem:comp}, we need the following lemma.
\begin{lemma}\label{lem:constant-solution}
	For any $f\in\calF$, let $n_0=5(\Deltamin(f)/\Rmax)^{-2}.$ Consider $w\in\R^{|\Pi|}_+$ such that $w_\pi=n_0$ for all $\pi\in\Pi$. Then $w$ is a valid solution to $\comp(f,n)$ for all $n>n_0.$ As a corollary, $\comp(f,n)\le \Deltamax(f)|\Pi|n_0$ for all $n>n_0.$
\end{lemma}
\begin{proof}[Proof of Lemma~\ref{lem:constant-solution}]
	Recall that $\Delta(f,\pi)=\max_{\pi'\in\Pi}R_f(\pi')-R_f(\pi)$ is the reward gap of decision $\pi$ under instance $f$, and $\Deltamin(f)=\min_{\pi:\Delta(f,\pi)>0}\Delta(f,\pi)$ is the minimum decision gap of the instance $f$.
	
	Let $\cset(f)=\{g\in\calF,\pi^\star(g)\neq\pi^\star(f)\}.$
	For any instance $g\in\cset(f)$, consider the following two decisions: $\pi_1=\pi^\star(f),\pi_2=\pi^\star(g).$ We claim that 
	\begin{align}\label{equ:comp-1}
		\max_{\pi\in\{\pi_1,\pi_2\}}\abs{R_{f}(\pi)-R_{g}(\pi)}\ge \frac{\Deltamin(f)}{3}.
	\end{align}
	This claim can be proved by contradiction. Suppose on the contrary that 
	$$\max_{\pi\in\{\pi_1,\pi_2\}}\abs{R_{f}(\pi)-R_{g}(\pi)}< \Deltamin(f)/3.$$
	Then we have
	\begin{align*}
		R_{g}(\pi_1)\ge R_{f}(\pi_1)-\frac{\Deltamin(f)}{3}\ge R_{f}(\pi_2)+\frac{2\Deltamin(f)}{3}\ge R_{g}(\pi_2)+\frac{\Deltamin(f)}{3}>R_{g}(\pi_2),
	\end{align*}
	which contradicts with the fact that $\pi_2=\pi^\star(g).$ 
	
	Recall that $R_f(\pi)=\E_{\ob\sim f[\pi]}[R(\ob)].$ As a result, $\abs{R_{f}(\pi)-R_{g}(\pi)}\le \Rmax\TV(f[\pi]\|g[\pi])$.
	Now, by Pinsker's inequality, for any $\pi\in\Pi$ we have
	\begin{align}
		\KL(f[\pi]\|g[\pi])\ge 2\TV(f[\pi]\|g[\pi])^2\ge 2(\abs{R_{f}(\pi)-R_{g}(\pi)}/\Rmax)^2.
	\end{align}
	Combining with Eq.~\eqref{equ:comp-1}, for any $g\in\cset(f)$ we get
	\begin{align}
		\sum_{\pi\in\Pi}n_0\KL(f[\pi]\|g[\pi])\ge 1.
	\end{align}
	Since this inequality holds for every $g\in\cset(f)$, we prove the desired result.
\end{proof}
Now we present the proof of Lemma~\ref{lem:comp}.
\begin{proof}[Proof of Lemma~\ref{lem:comp}]
	By Lemma~\ref{lem:constant-solution}, for $n>5(\Deltamin(f)/\Rmax)^{-2}$, we get $\comp(f,n)<\infty$.
\end{proof}

\subsection{Proof of Theorem~\ref{thm:lower-bound}}\label{app:pf-lower-bound}
Now we show the proof of Theorem~\ref{thm:lower-bound}.
\begin{proof}[Proof of Theorem~\ref{thm:lower-bound}]
	In the following we fix a consistent algorithm $\calA$. Consider any instance $g\in \calF$ such that $\pi^\star(g)\neq \pi^\star(f)$. 
	
	Recall that $\Delta(f,\pi)=\max_{\pi'\in\Pi}R_f(\pi')-R_f(\pi)$ is the reward gap of decision $\pi$ under instance $f$ and $\Deltamin(f)=\min_{\pi:\Delta(f,\pi)>0}\Delta(f,\pi)$ is the minimum decision gap of the instance $f$.
	Let $\epsilon=\min\{\Deltamin(g),\Deltamin(f)\}/2.$ Consider two distributions (over observations and decisions) $P_{f,n},P_{g,n}$ induced by running $n$ steps of algorithm $\calA$ on instance $f,g$ respectively. In addition, let $N_\pi$ be the random variable indicates the number of times decision $\pi$ is executed, and $\pi_i$ the random variable indicating the decision executed at step $i$.
	
	Let $\reg_{f,n}$ and $\reg_{g,n}$ be the regret of running algorithm $\calA$ on instances $f$ and $g$ respectively. By definition we have
	\begin{align}
		\frac{\reg_{f,n}+\reg_{g,n}}{\epsilon n}\ge \Pr\nolimits_{f}\(N_{\pi^\star(f)}\le \frac{n}{2}\)+\Pr\nolimits_{g}\(N_{\pi^\star(f)}> \frac{n}{2}\)\label{equ:lb-1}
	\end{align}
	By basic inequality of KL divergence \cite[Lemma 5]{lattimore2017end} we get,
	\begin{align}
		\Pr\nolimits_{f}\(N_{\pi^\star(f)}\le \frac{n}{2}\)+\Pr\nolimits_{g}\(N_{\pi^\star(f)}> \frac{n}{2}\)\ge \frac{1}{2}\exp(-\KL(P_{f,n}\|P_{g,n})).\label{equ:lb-2}
	\end{align}
	Combining Eq.~\eqref{equ:lb-1} and Eq.~\eqref{equ:lb-2} we get,
	\begin{align}
		\KL(P_{f,n}\|P_{g,n})\ge \ln\(\frac{\epsilon n}{2\(\reg_{f,n}+\reg_{g,n}\)}\).
	\end{align}
	Now applying the chain rule for KL divergence (see, e.g., \citep[Theorem 2.5.3]{cover1999elements}), we get
	\begin{align}
		\KL(P_{f,n}\|P_{g,n})=\E_{f}\[\sum_{i=1}^{n}\KL(f[\pi_i]\|g[\pi_i])\]=\sum_{\pi\in\Pi}\E_{f}[N_\pi]\KL(f[\pi]\|g[\pi]).
	\end{align}
	Therefore we have
	\begin{align}
		\sum_{\pi\in\Pi}\frac{\E_{f}[N_\pi]}{\ln\(\epsilon n\)-\ln\({2\(\reg_{f,n}+\reg_{g,n}\)}\)}\KL(f[\pi]\|g[\pi])\ge 1.
	\end{align}
	Consider the mixture of decisions $$w_\pi=\frac{\E_{f}[N_\pi]}{\ln\(\epsilon n\)-\ln\({2\(\reg_{f,n}+\reg_{g,n}\)}\)}.$$ Since $\reg_{f,n}+\reg_{g,n}\le n\Deltamax$ and $\E_{f}[N_\pi]\le n$, $w_\pi$ satisfies the constraint of $\comp\(f,n\)$ for large enough $n$. In addition, the expected regret of algorithm $\calA$ is 
	\begin{align}
		&\reg_{f,n}=\E_{f}\[\sum_{i=1}^{n}\Delta_{\pi_i}\]=\sum_\pi\E_{f}[N_\pi]\Delta_{\pi}\\
		\ge\;& \sum_\pi \Delta_{\pi}w_\pi\ln\(\frac{\epsilon n}{2\(\reg_{f,n}+\reg_{g,n}\)}\)\\
		\ge\;& \comp\(f,n\)\ln\(\frac{\epsilon n}{2\(\reg_{f,n}+\reg_{g,n}\)}\).
	\end{align}
	Since $\calA$ is consistent, for any $p>0$ we have $\reg_{f,n}+\reg_{g,n}=O(n^p).$ Consequently, for any $p>0$, 
	\begin{align}
		&\limsup_{n\to\infty}\frac{\reg_{f,n}}{\ln(n)}\\
		\ge\;& \limsup_{n\to\infty}\comp\(f,n\)\frac{\ln(n)-\ln\(2\(\reg_{f,n}+\reg_{g,n}\)/\epsilon\)}{\ln(n)}\\
		\ge\;& \comp(f)(1-p).
	\end{align}
	Since $p>0$ is an arbitrary constant, we get
	\begin{align}
		\limsup_{n\to\infty}\frac{\reg_{f,n}}{\ln(n)}\ge \comp(f).
	\end{align}
\end{proof}

\subsection{Instantiation of the Complexity Measure}
In this section, we instantiate our complexity measure $\comp(f)$ on concrete settings.

First we consider multi-armed bandits with unit Gaussian reward. The family of decisions is $\Pi=\{1,2,\cdots,A\}$. Let $(\mu_1,\mu_2,\cdots,\mu_A)$ be the mean rewards of each decision. An instance $f$ in this case is characterized by the mean rewards, and $f[i]=\calN(\mu_i,1).$
\begin{proposition}\label{prop:comp-MAB}
	For an multi-armed bandit instance $f$ with unique optimal decision and unit Gaussian noise, let $(\mu_1,\mu_2,\cdots,\mu_A)$ be the mean reward of each action. Then $$\comp(f)\le\sum_{i\in[1,A]\text{ and }\Delta_i> 0}\frac{2}{\Delta_i}$$ where $\Delta_i=\max_{i'}\mu_{i'}-\mu_i.$
\end{proposition}
\begin{proof}
	We prove this proposition by constructing a solution $w\in\R^{A}_+$ to the optimization problem $\comp(f,n)$ for large enough $n$.
	
	Without loss of generality, we assume $\mu_1>\mu_i$ for all $i\ge 2.$ Consider the action frequency 
	\begin{align}
		w_i=\begin{cases}
			\frac{2n}{n\Delta_i^2-3},&i\ge 2,\\
			n,&i=1.
		\end{cases}
	\end{align}
	Then when $n$ is large enough we have
	$\|w\|_\infty\le n.$ On the other hand, consider any $g\in\calF$ such that $\pi^\star(g)\neq \pi^\star(f).$ Suppose $\pi^\star(g)=i.$ In the following we show that 
	\begin{align}
		\sum_{i=1}^{n}w_i \KL(f[i]\|g[i])\ge 1.
	\end{align}
	Let $(\mu_1',\mu_2',\cdots,\mu_A')$ be the mean reward of instance $g$. For multi-armed bandits with unit Gaussian noise, we have
	\begin{align}
		\sum_{i=1}^{n}w_i \KL(f[i]\|g[i])=\frac{1}{2}\sum_{i=1}^{n}w_i(\mu_i-\mu_i')^2\ge \frac{1}{2}\(w_1(\mu_1-\mu_1')^2+w_i(\mu_i-\mu_i')^2\).
	\end{align}
	By the condition that $\pi^\star(g)=i$, we get $\mu_i'\ge \mu_1'.$ Combining with the fact that $\mu_1>\mu_i$ we get
	\begin{align}
		&\min_{\mu_1',\mu_i':\mu_1'\le \mu_i'}\(w_1(\mu_1-\mu_1')^2+w_i(\mu_i-\mu_i')^2\)\\
		=&\; \min_{\mu_1'\in[\mu_i,\mu_1]}\(w_1(\mu_1-\mu_1')^2+w_i(\mu_i-\mu_1')^2\)\\
		=&\; \frac{w_1w_i}{w_1+w_i}(\mu_1-\mu_i)^2.
	\end{align}
	By the definition of $w_i$, we have 
	\begin{align}
		&\frac{w_1w_i}{w_1+w_i}(\mu_1-\mu_i)^2=\frac{\frac{2n^2}{n\Delta_i^2-3}}{n+\frac{2n}{n\Delta_i^2-3}}\Delta_i^2
		=\frac{2\Delta_i^2n}{n\Delta_i^2-1}\ge 2.
	\end{align}
	As a result, 
	\begin{align}
		\comp(f,n)\le \sum_{i=2}^{A}\frac{2n}{n\Delta_i^2-3}\Delta_i.
	\end{align}
	It follows that
	\begin{align}
		\comp(f)=\lim_{n\to\infty}\comp(f,n)\le \lim_{n\to\infty}\sum_{i=2}^{A}\frac{2n}{n\Delta_i^2-3}\Delta_i=\sum_{i=2}^{A}\frac{2}{\Delta_i}.
	\end{align}
\end{proof}

In the following, we focus on a linear bandit instance $f$ where the action set is $\calA\subset \R^{d}.$ 
The mean of an action $x\in\calA$ is given by $\mu_x=\dotp{x}{\theta}$ for some $\theta\in\R^{d}.$ Let $x^\star=\argmax_x \mu_x$ be the optimal action, and  $\Delta_x\defeq \mu_{x^\star}-\mu_x$ be the sub-optimality gap of action $x$. Define $\calA^-=\calA\setminus \{x^\star\}.$
An linear bandit instance is characterized by the vector $\theta$, and $f[x]=\calN(\dotp{x}{\theta},1).$

We assume that $\calA$ is discrete, full rank, and $\|x\|_2\le 1$ for all $x\in\calA.$ Then we have the following proposition.
\begin{proposition}\label{prop:comp-LB}
	For an linear bandit instance $f$ with unique optimal decision and unit Gaussian noise, our complexity measure $\comp(f)$ recovers that in \citet{lattimore2017end}. That is,
	\begin{align}
		\comp(f)\le \inf_{w\in \R^A_+}\;&\sum_{x\in\calA^-}w_x\Delta_x,\label{equ:clb-1}\\
		\text{s.t. }\;&\|x\|_{H(w)^{-1}}^2\le \frac{\Delta_x^2}{2},\quad\forall x\in\calA^-,\label{equ:clb-4}
	\end{align}
	where $H(w)=\sum_{x}w_xxx^\top.$
\end{proposition}
\begin{proof}
	Let $\hat{w}$ be the solution to the RHS of Eq.~\eqref{equ:clb-1}. In the following we construct a solution to our optimization problem $\comp(f,n)$ from $\hat{w}$. Recall that 
	\begin{align}
		\comp(f, n)  \defeq \min_{w\in \R^{|\calA|}_+}\;&\sum_{x\in\calA}w_\pi\Delta(f,x)\\
		\text{s.t.}\quad &\sum_{x\in\calA}w_x\KL(f[x]\|g[x])\ge 1,\;\forall g\in\calF,\pi^\star(g)\neq\pi^\star(f),\label{equ:clb-2}\\
		&\|w\|_\infty\le n.
	\end{align}
	For any $w\in\R^{A}_+$ and any instance $g\in\calF$ associates with parameter $\theta'$, we have
	\begin{align}
		\sum_{x\in\calA}w_x\KL(f[x]\|g[x])=\sum_{x\in\calA}w_x\KL(\calN(\dotp{x}{\theta},1)\|\calN(\dotp{x}{\theta'},1))=\frac{\|\theta-\theta'\|_{H(w)}^2}{2}.
	\end{align}
	Consider any $g$ such that $\pi^\star(g)\neq \pi^\star(f)$. Suppose $\pi^\star(g)=x\neq x^\star$. It follows that $\dotp{x-x^\star}{\theta'}>0.$ Recall that $\Delta_x=\dotp{x^\star-x}{\theta}.$ 
	By algebraic manipulation we have,
	\begin{align}
		\min_{\theta':\dotp{x^\star-x}{\theta-\theta'}> \Delta_x}\frac{1}{2}\|\theta-\theta'\|_{H(w)}^2=\frac{1}{2}\frac{\Delta_x^2}{\|x^\star-x\|_{H(w)^{-1}}^2}\label{equ:clb-3}
	\end{align}
	Therefore to satisfy Eq.~\eqref{equ:clb-2}, it's enough to construct a solution $w$ such that \begin{align}
		\frac{1}{2}\frac{\Delta_x^2}{\|x^\star-x\|_{H(w)^{-1}}^2}\ge 1,\forall x\in\calA^-.\label{equ:clb-5}
	\end{align}
	
	Define $A=\sum_{x\in\calA^-}\hat{w}_xxx^\top+(x^\star)(x^\star)^\top.$ When the action set $\calA$ is full rank, $A$ is positive definite (see \citet[Appendix C]{lattimore2017end}). We use $\sigmamax(A),\sigmamin(A)$ to denote the maximum/minimum singular value of a matrix $A$ respectively. Then for any $n>0$, consider the following solution
	\begin{align}
		w_x=\begin{cases}
			\hat{w}_x\(1-\frac{8}{\Deltamin^2}\frac{\sigmamax(A^{-1})}{1+(n-1)\sigmamin(A^{-1})}\)^{-1},&\text{ when } x\neq x^\star,\\
			n,&\text{ when } x=x^\star.
		\end{cases}
	\end{align}
	For large enough $n$ we get $\|w\|_\infty\le n.$ In the following, we prove that $w$ satisfies Eq.~\eqref{equ:clb-3}. Let $$c_n=\(1-\frac{8}{\Deltamin^2}\frac{\sigmamax(A^{-1})}{1+(n-1)\sigmamin(A^{-1})}\)^{-1}$$ for shorthand. 
	Since $\Delta_{x^\star}=0$, we have $\hat{w}_{x^\star}=\infty.$ Therefore $\|x^\star\|_{H(\hat{w})^{-1}}=0.$
	Then for any $x\in\calA^-$, by Eq.~\eqref{equ:clb-4} we have
	\begin{align}
		&(x^\star-x)^\top H(c_n\hat{w})^{-1}(x^\star-x)=c_n^{-1}(x^\star-x)^\top H(\hat{w})^{-1}(x^\star-x)\\
		=\;&c_n^{-1}x^\top H(\hat{w})^{-1}x
		\le \frac{\Delta_x^2}{2}-\frac{4\sigmamax(A^{-1})}{1+(n-1)\sigmamin(A^{-1})}.
	\end{align}
	Invoking Lemma~\ref{lem:lb-cap} we get
	\begin{align*}
		(x^\star-x)^\top H(w)^{-1}(x^\star-x)\le (x^\star-x)^\top H(c_n\hat{w})^{-1}(x^\star-x)+\frac{4\sigmamax(A^{-1})}{1+(n-1)\sigmamin(A^{-1})}=\frac{\Delta_x^2}{2},
	\end{align*}
	which implies Eq.~\eqref{equ:clb-5}. As a result, $w$ is a valid solution ot $\comp(f,n).$ Consequently,
	\begin{align}
		\comp(f,n)\le \sum_{x}\Delta_xw_x\le c_n\sum_{x}\hat{w}_x\Delta_x.
	\end{align}
	By definition we have $\lim_{n\to\infty}c_n=1$. Consequently, $\comp(f)=\lim_{n\to\infty}\comp(f,n)\le \sum_{x}\hat{w}_x\Delta_x.$
\end{proof}

\subsection{A Sufficient Condition for Condition~\ref{cond:uniform-convergence}}
In this section, we discuss a sufficient but more interpretable condition to  Condition~\ref{cond:uniform-convergence}.
\begin{proposition}\label{prop:bounded-moments}
	Suppose there exists a constant $\const_M$ such that for every $f,g\in\calF$ and $\pi\in\Pi$, $$\E_{\ob\sim f[\pi]}\[\(\ln \frac{f[\pi](\ob)}{g[\pi](\ob)}\)^4\]\le \const_M^4.$$ Then Condition~\ref{cond:uniform-convergence} holds with $\lambda_0(\alpha,\epsilon,f)=\min\{2\epsilon/\const_M^2,1/2\}.$
\end{proposition}
\begin{proof}
	For any fixed $f,g\in\calF$, $\pi\in\Pi$ and $\lambda<1/2$, by Lemma~\ref{lem:KL-renyi-difference} we get
	\begin{align}
		&\KL(f[\pi]\|g[\pi])-D_{1-\lambda}(f[\pi]\|g[\pi])\le \frac{\lambda}{2}\E_{\ob\sim f[\pi]}\[\(\ln \frac{f[\pi](\ob)}{g[\pi](\ob)}\)^4\]^{1/2}.
	\end{align}
	Therefore we have
	\begin{align}
		\KL(f[\pi]\|g[\pi])-D_{1-\lambda}(f[\pi]\|g[\pi])\le \frac{\lambda}{2}\const_M^2.
	\end{align}
	Since $D_{1-\lambda}(f[\pi]\|g[\pi])$ is monotonically decreasing with $\lambda$, we prove the desired result.
\end{proof}

\subsection{Comparison with Existing Lower Bounds}\label{sec:comp-tirinzoni-lb}
Recall that when applied to tabular RL problems, our lower bound is very similar to that in \citet{tirinzoni2021fully}, and the only difference is that their optimization problems omit the second constraint (Eq.~\eqref{equ:comp-def-3}),
On a high level, the constraint $\|w_\pi\|_{\infty}\le n$ is necessary to prevent degenerate cases and guarantee mathematical rigor.

The value $\comp(f,n)$ can be different from the solution without this constraint for some artificial hypothesis class $\mathcal{F}$. For example, we can construct a multi-armed bandit hypothesis class $\mathcal{F}=\{\mu\in\mathbb{R}^{A}:\mu(1)\neq 0.5\}\cup \{[0.5,0.1,\cdots,0.1]\}$ (where $\mu\in\mathbb{R}^{A}$ represents the mean reward of each arm). Then for $f=[0.5,0.1,\cdots,0.1]$, $\comp(f,n)>0$ for every $n>0$ (because there exits other instances in $\mathcal{F}$ whose mean reward of action 1 is arbitrarily close of 0.5). As a result, $\comp(f)=\lim_{n\to\infty}\comp(f,n)>0$ by the definition of limits and in this case $\lim_{n\to\infty} \comp(f,n)\neq \comp(f,\infty)$. However, without the constraint $\|w_\pi\|_{\infty}\le n$, the solution will be $0$, achieved by letting $w_1=\infty$ and $w_{i}=0,\forall i\neq 1$.

For other hypothesis classes (such as the standard MAB and linear bandits discussed in Proposition~\ref{prop:comp-MAB} \& ~\ref{prop:comp-LB}, and tabular RL), however, this constraint does not change the value of $C(f,n).$
	\section{Extension to Infinite Hypothesis Class}\label{app:infinite-hypothesis}
In this section, we extend our results to the infinite hypothesis case.
\subsection{Proof of Theorem~\ref{thm:main}}\label{app:pf-main}
To prove Theorem~\ref{thm:main}, we require the following lemmas for steps 1 and 2 by analogy with the finite hypothesis case.
\begin{lemma}[Main lemma for Initialization]\label{lem:main-init}
	Let $\init$ be the event that, there exists a universal constant $\const_4>0$ such that 
	\begin{enumerate}[label=(\alph*)]
		\item $\max_\pi\KL(f^\star[\pi]\|\initf[\pi])\le \(\frac{\ln\ln n}{\ln n}\)^{\const_4}$,
		\item $\abs{R_{\initf}(\pi)-R_{\truef}(\pi)}\le \Rmax\(\frac{\ln\ln n}{\ln n}\)^{\const_4}$, for all $\pi\in\Pi$,
		\item $\pi^\star(\initf)=\pi^\star(\truef).$
	\end{enumerate}
	Under Conditions~\ref{cond:uniform-convergence} and~\ref{cond:covering-number}, there exists $n_0>0$ such that when $n>n_0$, $\Pr(\init)\ge 1-1/\ln n.$ In addition, the regret of Step 1 is upper bounded by $\bigO(\frac{\ln n}{\ln\ln n}).$
\end{lemma}
Proof of Lemma~\ref{lem:main-init} is deferred to Appendix~\ref{app:pf-main-init}.

\begin{lemma}[Main lemma for Identification]\label{lem:main-ident}
	Under Conditions~\ref{cond:uniform-convergence}, \ref{cond:covering-number}, and \ref{cond:TV-vs-inf}, there exists $n_0>0$ such that when $n>n_0$, the following holds.
	\begin{enumerate}[label=(\alph*)]
		\item Conditioned on the event $\pi^\star(\initf)\neq \pi^\star(f^\star)$, $\Pr(\acc)\le 1/n$.
		\item Conditioned on the event $\init$, $\Pr(\acc)\ge 1-1/\ln n$.
		\item The expected regret of Step 2 is always upper bounded by $\bigO(\ln n\ln\ln n).$
		\item Conditioned on the event $\init$, the expected regret of Step 2 is upper bounded by $$\(\comp(f^\star,(\ln\ln n)^{1/4}/2)+o(1)\)\ln n.$$
	\end{enumerate}
\end{lemma}
Proof of Lemma~\ref{lem:main-ident} is deferred to Appendix~\ref{app:pf-main-ident}.

Then, proof of Theorem~\ref{thm:main} is the same as that of Theorem~\ref{thm:main-finite} by plugging in Lemma~\ref{lem:main-init} and Lemma~\ref{lem:main-ident},

\subsection{Proof of Lemma~\ref{lem:main-init}}\label{app:pf-main-init}
\begin{proof}[Proof of Lemma~\ref{lem:main-init}] We prove the two items in this lemma separately. 
	
	\paragraph{Proof of item (a):}
	Let $\const_1>0$ be the constant from Condition~\ref{cond:uniform-convergence}. Set $\const_6=\frac{1}{2\const_1+3}$,  $\alpha=\(\frac{\ln\ln n}{\ln n}\)^{\const_6}$, and $\epsilon=\frac{\alpha}{5}$. Let $w=\(\pi_1,\cdots,\pi_{\minit}\)$ be the list of decisions run by Step 1, and $\ob_1,\cdots,\ob_\minit$ the corresponding observations. Recall that by definition,
	\begin{align}
		\initf=\argmax_{g\in\calF} \sum_{i=1}^{\minit}\ln g[\pi_i](\ob_i).
	\end{align}
	Combining with the fact that $\truef\in\calF$, we have
	\begin{align}\label{equ:pmi-1}
		\sum_{i=1}^{\minit}\ln \frac{\truef[\pi_i](\ob_i)}{\initf[\pi_i](\ob_i)}\le \sum_{i=1}^{\minit}\ln \frac{\truef[\pi_i](\ob_i)}{\truef[\pi_i](\ob_i)}\le 0.
	\end{align}
	Let $\calG(\alpha)=\{g\in\calF:\exists\pi:\KL(\truef[\pi]\|g[\pi])\ge \alpha\}.$
	We will prove that for all $g\in\calG(\alpha)$, we have $\sum_{i=1}^{\minit}\ln \frac{\truef[\pi_i](\ob_i)}{g[\pi_i](\ob_i)}>0$. Combining with Eq.~\eqref{equ:pmi-1} we get $\KL(\truef[\pi]\|\initf[\pi])\le \alpha,\forall \pi.$
	
	To this end, we apply Lemma~\ref{lem:uniform-concentration} with parameters $(\alpha/|\Pi|,\epsilon/|\Pi|,w)$. Following Lemma~\ref{lem:uniform-concentration}, define $$\gamma=\frac{1}{\minit}\min_\pi\sum_{i=1}^{\minit}\ind{\pi_i=\pi}.$$ Then we have $\gamma=1/|\Pi|.$
	Let $\lambda=\lambda_0(\alpha,\epsilon/|\Pi|,f)$ be the value that satisfies Condition~\ref{cond:uniform-convergence}, and 
	$$\epsilon_0=\frac{\exp(-\alpha)(\epsilon \lambda /|\Pi|)^{1/\lambda}}{3\vol}.$$
	Recall that the condition of Lemma~\ref{lem:uniform-concentration} states 
	\begin{align}\label{equ:pmi-2}
		\minit\ge \frac{1}{\lambda\epsilon}\(\ln\calN(\calF,\epsilon_0)+\ln(1/\delta))\).
	\end{align}
	The failure probability in this case is $\delta=1/\ln n$. By Condition~\ref{cond:uniform-convergence}, for large enough $n$ we get $1/\lambda\lesssim \epsilon^{-\const_1}.$ By Condition~\ref{cond:covering-number} we get $$\ln\calN(\calF,\epsilon_0)\lesssim \ln (1/\epsilon_0)\lesssim \bigO(1)+\frac{1}{\lambda}\ln (1/(\epsilon\lambda))+\alpha.$$ As a result, when $n$ is large enough
	\begin{align}
		&\frac{1}{\lambda\epsilon}\(\ln\calN(\calF,\epsilon_0)+\ln(1/\delta))\)\lesssim \frac{1}{\epsilon^{\const_1+1}}\(\bigO(1)+\ln\ln n + \frac{1}{\epsilon^{\const_1}}\ln\frac{1}{\epsilon^{\const_1}}\)\\
		\lesssim\;&\epsilon^{-(2c_1+2)}\lesssim \(\frac{\ln n}{\ln\ln n}\)^{1-\frac{1}{2\const_1+3}},
	\end{align}
	where the last inequality comes from the definition of $\epsilon$, i.e., $\epsilon=\bigO\(\(\frac{\ln\ln n}{\ln n}\)^{\frac{1}{2\const_1+3}}\)$.
	Recall that $\minit\ge |\Pi|\frac{\ln n}{\ln\ln n}$. When $n$ is large enough, the condition of Lemma~\ref{lem:uniform-concentration} (i.e., Eq.~\eqref{equ:pmi-2}) is satisfied. 
	
	Because every policy appears in $w$ exactly the same number of times, we have $\KL^w(\truef\|g)\ge \alpha/|\Pi|$ for all $g\in\calG(\alpha)$.
	Therefore, by Lemma~\ref{lem:uniform-concentration} with paremeters $(\alpha/|\Pi|,\epsilon/|\Pi|,w)$,
	\begin{align}
		\forall g\in\calG(\alpha),\sum_{i=1}^{\minit}\ln \frac{\truef[\pi_i](\ob_i)}{g[\pi_i](\ob_i)}\ge \(\frac{\alpha}{|\Pi|}-4\frac{\epsilon}{|\Pi|}\)m>0.
	\end{align}
	Combining with Eq.~\eqref{equ:pmi-1}, we get $\initf\not\in \calG(\alpha)$. As a result 
	\begin{align}\label{equ:pf-mi-1}
		\forall \pi\in\Pi,\quad \KL(\truef[\pi]\|\initf[\pi])\le \alpha=\(\frac{\ln\ln n}{\ln n}\)^{\const_6}.
	\end{align}
	
	\paragraph{Proof of item (b):} Now we focus on item (b). For any fixed $\pi\in\Pi$, by Pinsker's inequality and Eq.~\eqref{equ:pf-mi-1} we get
	\begin{align}
		\TV(\truef[\pi]\| \initf[\pi])\le \sqrt{\frac{1}{2}\KL(\truef[\pi]\|\initf[\pi])}\le \(\frac{\ln\ln n}{\ln n}\)^{\const_6/2}.
	\end{align}
	By assumption we have $0\le R(\ob)\le \Rmax$ almost surely for both $\truef[\pi]$ and $\initf[\pi]$. It follows that
	\begin{align}
		\abs{R_{\initf}(\pi)-R_{\truef}(\pi)}\le \Rmax\TV(\truef[\pi]\| \initf[\pi]).
	\end{align}
	Then we prove item (b) with $\const_4=\const_6/2$ and $\iota(\truef)=\Rmax.$
	
	\paragraph{Proof of item (c):} Since $\Deltamin(\truef)>0$, (c) follows from (b) directly when $n$ is large enough.
	
	\paragraph{Proof of regret:} The number of samples collected in Step 1 is upper bounded by $\minit=|\Pi|\lceil \frac{\ln n}{\ln\ln n}\rceil.$ As a result, the regret is upper bounded by 
	\begin{align}
		\bigO(\Deltamax\minit)=\bigO\(\frac{\ln n}{\ln\ln n}\).
	\end{align}
\end{proof}

\subsection{Proof of Lemma~\ref{lem:main-ident}}\label{app:pf-main-ident}
\begin{proof}[Proof of Lemma~\ref{lem:main-ident}] We prove the four items in Lemma~\ref{lem:main-ident} separately.
	
	\paragraph{Item (a):}
	First we prove item (a) of Lemma~\ref{lem:main-ident}. By Markov inequality, for any $c>0$, we have
	\begin{align}
		&\Pr_{f^\star}\(\sum_{i=1}^{m}\ln\frac{\initf[\pi_i](\ob_i)}{\truef[\pi_i](\ob_i)}\ge \ln n\)=\Pr_{f^\star}\(\exp\(\sum_{i=1}^{m}\ln\frac{\initf[\pi_i](\ob_i)}{\truef[\pi_i](\ob_i)}\)\ge \exp(\ln n)\)\\
		\le\;&\exp(-\ln n)\E_{f^\star}\[\exp\(\sum_{i=1}^{m}\ln\frac{\initf[\pi_i](\ob_i)}{\truef[\pi_i](\ob_i)}\)\]
		=\exp(-\ln n)\prod_{i=1}^{m}\E_{f^\star}\[\frac{\initf[\pi_i](\ob_i)}{\truef[\pi_i](\ob_i)}\]\\
		=\;&1/n.
	\end{align}
	The last equality follows from the fact that $\initf[\pi]$ and $f^\star[\pi]$ are both probability distributions given any decision $\pi\in \Pi.$
	
	Recall that in this case $\pi^\star(\initf)\neq \pi^\star(f^\star)$. Therefore,
	\begin{align}
		&\Pr(\acc^\initf)=\Pr\(\forall g\in\calF\text{ and }\pi^\star(g)\neq\pi^\star(\initf), \sum_{i=1}^{m}\ln\frac{\initf[\pi_i](\ob_i)}{g[\pi_i](\ob_i)}\ge \ln n\)\\
		\le\;&\Pr\(\sum_{i=1}^{m}\ln\frac{\initf[\pi_i](\ob_i)}{\truef[\pi_i](\ob_i)}\ge \ln n\)
		\le 1/n.
	\end{align}
	
	\paragraph{Item (b):}
	Let $\epsilon=1/\ln\ln n$ and $\alpha=\frac{\ln n}{m}$. We prove this statement by invoking Lemma~\ref{lem:uniform-concentration} with parameters $(\alpha+5\epsilon,\epsilon,w)$. Following Lemma~\ref{lem:uniform-concentration}, let $\gamma=\frac{1}{m}\min_{\pi} \sum_{i=1}^{m}\ind{\pi_i=\pi}$ and $\lambda=\lambda_0((\alpha+5\epsilon)/\gamma,\epsilon,\truef)$ be the value that satisfies Condition~\ref{cond:uniform-convergence}. Let $$\epsilon_0=\frac{\exp(-(\alpha+5\epsilon)/\gamma)(\epsilon\lambda)^{1/\lambda}}{3\vol}.$$
	
	First of all, we prove that the condition for Lemma~\ref{lem:uniform-concentration} holds. That is,
	\begin{align}\label{equ:mi-1}
		m\ge \frac{1}{\lambda\epsilon}\(\ln\calN\(\calF,\epsilon_0\)+\ln\ln n\).
	\end{align}
	Recall that in Alg.~\ref{alg:main-finite}, $\hat{w}$ is the solution of $\comp(\initf,(\ln\ln n)^{1/4})$, $\bar{w}_\pi=\(1+\frac{1}{(\ln\ln n)^{1/4}}\)\hat{w}_\pi+\frac{1}{(\ln\ln n)^{1/4}}$, and
	$m=\sum_{x}\lceil \bar{w}_\pi\ln (n)\rceil$. As a result, $m\ge \frac{|\Pi|\ln n}{(\ln\ln n)^{1/4}}.$ Now consider the RHS of Eq.~\eqref{equ:mi-1}. By the definition of $\bar{w}_\pi$ we get $m\le 2|\Pi|\ln n(\ln\ln n)^{1/4}$, so $\alpha\le \frac{1}{|\Pi|}(\ln\ln n)^{1/4}$ and $\gamma^{-1}\le 2|\Pi|(\ln\ln n)^{1/2}$. It follows from Condition~\ref{cond:uniform-convergence} that $\lambda\ge \poly(1/\ln\ln n).$ By the definition of $\epsilon_0$ and Condition~\ref{cond:covering-number} we get
	\begin{align}
		\ln\calN\(\calF,\epsilon_0\)\lesssim \ln(1/\epsilon_0)\lesssim \poly(\ln\ln n).
	\end{align}
	Consequently, when $n$ is sufficiently large, Eq.~\eqref{equ:mi-1} holds. By Lemma~\ref{lem:uniform-concentration} we get, with probability at least $1-1/\ln n$,
	\begin{align}\label{equ:mi-2}
		\sum_{i=1}^{m}\ln \frac{\truef[\pi_i](\ob_i)}{g[\pi_i](\ob_i)}\ge (\alpha+\epsilon)m,\quad \forall g\in \calF(w,f^\star,\alpha+5\epsilon),
	\end{align}
	where $\calF(w,f^\star,\alpha+5\epsilon)=\{g\in\calF:\KL^w(\truef\|g)\ge \alpha+5\epsilon\}$.
	In the following, we prove that Eq.~\eqref{equ:mi-2} implies $\acc^{\initf}.$ Recall that $\acc^{\initf}$ is the event defined as follows:
	\begin{align}\label{equ:acc-initf}
		\acc^\initf&=\ind{\forall g\in \cset(\initf), \sum_{i=1}^{m}\ln\frac{\initf[\pi_i](\ob_i)}{g[\pi_i](\ob_i)}\ge \ln n}.
	\end{align}
	Recall that $\cset(\initf)=\{g\in\calF,\pi^\star(g)\neq \pi^\star(\initf)\}.$ Next, we apply Lemma~\ref{lem:KL-closeness-f} to show that $\cset(\initf)\subseteq \calF(w,f^\star,\alpha+5\epsilon).$ To verify the condition of Lemma~\ref{lem:KL-closeness-f}, we have
	$\TV(\truef[\pi]\|\initf[\pi])\lesssim \(\frac{\ln\ln n}{\ln n}\)^{\const_4}$ for all $\pi\in\Pi$ by item (a) of Lemma~\ref{lem:main-init}. By the definition of $\hat{w}$ we get
	\begin{align}
		\forall g\in\cset(\initf), \quad \sum_{\pi\in\Pi}\hat{w}_\pi\KL(\initf[\pi]\|g[\pi])\ge 1.
	\end{align}
	Therefore when $n$ is large enough, 
	\begin{align}
		\forall g\in\cset(\initf), \KL^w(f^\star\|g)\ge \frac{\ln n}{m}+5\epsilon=\alpha+5\epsilon.
	\end{align}
	Then $\cset(\initf)\subseteq\calF(w,\truef,\alpha+5\epsilon)$. It follows from Eq.~\eqref{equ:mi-2} that 
	\begin{align}
		\sum_{i=1}^{m}\ln \frac{\truef[\pi_i](\ob_i)}{g[\pi_i](\ob_i)}\ge (\alpha+\epsilon)m,\quad\forall g\in\cset(\initf).
	\end{align}
	Finally, by Condition~\ref{cond:TV-vs-inf} we get $\truef[\pi](\ob)>\const_{\rm min}$ for any $\pi\in\Pi\text{ and }\ob\in \supp(\truef[\pi])$. As a result
	\begin{align}
		\abs{\ln \frac{\initf[\pi](\ob)}{\truef[\pi](\ob)}}\le \abs{\ln \(1+\frac{\initf[\pi](\ob)-\truef[\pi](\ob)}{\truef[\pi](\ob)}\)}\le \abs{\ln \(1+\frac{\initf[\pi](\ob)-\truef[\pi](\ob)}{\const_{\rm min}}\)}.
	\end{align}
	When $\|\initf-f^\star\|_\infty\le \const_{\rm min}/2$, applying the basic inequality $\abs{\ln(1+x)}\le 2x,\forall |x|\le 1/2$ we get
	\begin{align}
		\abs{\ln \(1+\frac{\initf[\pi](\ob)-\truef[\pi](\ob)}{\const_{\rm min}}\)}\le \frac{2}{\const_{\rm min}}\|\initf-f^\star\|_\infty\lesssim \frac{2}{\const_{\rm min}}\(\frac{\ln\ln n}{\ln n}\)^{\const_4\const_5},
	\end{align}
	where the last inequality comes from item (a) of Lemma~\ref{lem:main-init} and Condition~\ref{cond:TV-vs-inf}.
	Therefore, for large enough $n$ we get 
	$\abs{\ln \frac{\initf[\pi](\ob)}{\truef[\pi](\ob)}}\le \epsilon.$
	As a result, 
	\begin{align}
		\sum_{i=1}^{m}\ln \frac{\initf[\pi_i](\ob_i)}{g[\pi_i](\ob_i)}=\sum_{i=1}^{m}\ln \frac{\truef[\pi_i](\ob_i)}{g[\pi_i](\ob_i)}+\sum_{i=1}^{m}\ln \frac{\initf[\pi_i](\ob_i)}{\truef[\pi_i](\ob_i)}\ge \alpha m=\ln n.
	\end{align}
	Since $g\in\cset(\initf)$ is arbitrary, have
	\begin{align}
		&\Pr(\acc^{\initf})=\Pr\(\forall g\in \cset(\initf),\sum_{i=1}^{m}\ln \frac{\initf[\pi_i](\ob_i)}{g[\pi_i](\ob_i)}\ge \ln n\)\\
		\ge&\; \Pr\(\forall g\in \calF(w,f^\star,\alpha+5\epsilon),\sum_{i=1}^{m}\ln \frac{\truef[\pi_i](\ob_i)}{g[\pi_i](\ob_i)}\ge (\alpha+\epsilon)m\)\ge 1-1/\ln n.
	\end{align}
	
	\paragraph{Item (c):} By the definition of $\comp(\hat{f},(\ln\ln n)^{1/4})$, we have $\hat{w}_\pi\le (\ln\ln n)^{1/4}$ for every $\pi\in \Pi$. As a result, $m\le 2A\ln n(\ln\ln n)^{1/4}.$ Therefore, the expect regret of Step 2 is upper bounded by 
	\begin{align}
		\Deltamax2A\ln n(\ln\ln n)^{1/4}=O(\ln n\ln\ln n).
	\end{align}
	
	\paragraph{Item (d):} Recall that $\hat{w}$ is the solution of $\comp(\initf,(\ln\ln n)^{1/4}).$ As a result, the regret of Step 2 is upper bounded by 
	\begin{align}
		\(\sum_{\pi\in\Pi} \hat{w}_\pi\Delta(\truef,\pi)+o(1)\)\ln n
	\end{align}
	where $\Delta(\truef,\pi)$ is the sub-optimality gap of decision $\pi$ under instance $\truef$. In the following, we prove that
	\begin{align}\label{equ:mi-4}
		\sum_{\pi\in\Pi} \hat{w}_\pi\Delta(\truef,\pi)\le \comp(\truef,(\ln\ln n)^{1/4}/2).
	\end{align}
	Let $\hat{w}^\star$ be the solution to $\comp(\truef,(\ln\ln n)^{1/4}/2)$. Define $\delta=\frac{1}{(\ln\ln n)^{1/4}}.$ Let 
	$\bar{w}^\star\defeq \{\(1+\delta\)\hat{w}^\star_\pi+\delta\}_{\pi}$ and $m^\star=\sum_{\pi\in\Pi}\lceil \bar{w}_\pi^\star\ln n\rceil$. We will show that $\bar{w}^\star$ is also (approximately) a solution of $\comp(\initf,(\ln\ln n)^{1/4}/2)$
	
	By the definition of $\comp(\truef,(\ln\ln n)^{1/4}/2)$, for any $g\in\cset(\truef),$ we have
	\begin{align}
		\sum_{\pi\in\Pi}(\hat{w}^\star_\pi)\KL(\truef[\pi]\|g[\pi])\ge 1.
	\end{align}
	Let $w^\star$ be the list of decisions that $\pi$ appears $\lceil \bar{w}^\star\ln n\rceil$ times for every $\pi\in\Pi$. 
	Define $\alpha^\star=\frac{\ln n}{m^\star}$ and $\epsilon=\frac{1}{\ln\ln n}.$ 
	Next we apply Lemma~\ref{lem:KL-closeness-f} with parameters $(\epsilon,w^\star).$ To verify its condition, item (a) of Lemma~\ref{lem:main-init} gives
	\begin{align}
		\TV(\truef[\pi]\|\initf[\pi])\le \KL(\truef[\pi]\|\initf[\pi])^{1/2}\lesssim \(\frac{\ln\ln n}{\ln n}\)^{\const_4/2},\quad\forall \pi\in\Pi,
	\end{align}
	which satisfies the condition of Lemma~\ref{lem:KL-closeness-f}. Consequently, we get
	$
		\KL^{w^\star}(\initf[\pi]\|g[\pi])\ge \frac{\ln n}{m^\star}
	$ for every $g\in\cset(\truef)$.
	Therefore,
	\begin{align}
		\forall g\in\cset(\truef),\quad \sum_{\pi\in\Pi}\frac{\lceil \bar{w}^\star_\pi\ln n\rceil}{\ln n}\KL(\initf[\pi]\|g[\pi])\ge 1.
	\end{align}
	By item (c) of Lemma~\ref{lem:main-init}, $\cset(\initf)=\cset(\truef)$.
	When $n$ is large enough, we have $\frac{\lceil \bar{w}^\star_\pi\ln n\rceil}{\ln n}\le (\ln\ln n)^{1/4}.$ Therefore, $\left\{\frac{\lceil \bar{w}^\star_\pi\ln n\rceil}{\ln n}\right\}_{\pi\in\Pi}$ satisfies all the constraints of $\comp(\initf,(\ln\ln n)^{1/4})$.
	Recall that $\hat{w}$ is the solution to $\comp(\initf,(\ln\ln n)^{1/4})$. By the optimality of $\hat{w}$ we have
	\begin{align}
		&\sum_{\pi}\hat{w}_\pi\Delta(\initf,\pi)\le \sum_{\pi}\frac{\lceil \bar{w}^\star_\pi\ln n\rceil}{\ln n}\Delta(\initf,\pi)\le \sum_{\pi} \bar{w}^\star_\pi\Delta(\initf,\pi)+o(1).
	\end{align}
	By item (b) of Lemma~\ref{lem:main-init}, $\abs{\Delta(\initf,\pi)-\Delta(\truef,\pi)}\le \iota(\truef)\(\frac{\ln\ln n}{\ln n}\)^{\const_4}.$ As a result,
	\begin{align}
		&\sum_{\pi} \bar{w}^\star_\pi\Delta(\initf,\pi)\le \sum_{\pi} \bar{w}^\star_\pi\Delta(\truef,\pi)+o(1)\\
		\le\;&\sum_{\pi} \hat{w}^\star_\pi\Delta(\truef,\pi)+o(1)=\comp(\truef,(\ln\ln n)^{1/4}/2)+o(1).
	\end{align}
	In addition,
	\begin{align}
		\sum_{\pi\in\Pi} \hat{w}_\pi\Delta(\truef,\pi)\le \sum_{\pi}\hat{w}_\pi\Delta(\initf,\pi)+o(1).
	\end{align}
	Stitching the inequalities above we have
	\begin{align}
		\sum_{\pi}\hat{w}_\pi\Delta(\truef,\pi)\le \comp(\truef,(\ln\ln n)^{1/4}/2)+o(1).
	\end{align}
	As a result, the regret in Step 2 is bounded by $$\(\comp(\truef,(\ln\ln n)^{1/4}/2)+o(1)\)\ln n.$$
\end{proof}
	\section{Proof of Conditions for Tabular Reinforcement Learning}\label{app:rl}
In this section, we first prove that Conditions~\ref{cond:uniform-convergence}-~\ref{cond:TV-vs-inf} holds for tabular RL with truncated Gaussian reward. Consequently, Theorem~\ref{thm:main} gives the first asymptotic instance-optimal regret bound in this setting. Later in Appendix~\ref{app:rl-Gaussian}, we extend our analysis to Gaussian reward.

First of all, we define some additional notations. 
A finite horizon Markov Decision Process (MDP) is denoted by a tuple $(\calS,\calA,p,r,H)$. $\calS$ and $\calA$ are the state and action spaces respectively. $H>0$ denotes the horizon. The tansition function $p$ maps a state-action pair $(s,a)\in\calS\times\calA$ to a distribution over $\calS$, and  the reward function $r$ maps state-action pair $(s,a)$ to a truncated Gaussian distribution. In particular, for a fixed $(s,a)$, the mean reward is $\mu(s,a)\in[-1,1]$, and the density of the reward $r(s,a)$ is 
\begin{align}
	r[s,a](x)=\ind{x\in[-2,2]}\frac{1}{Z}\exp\(-\frac{(x-\mu(s,a))^2}{2}\),
\end{align}
where $Z=\int_{x\in[-2,2]}\exp\(-(x-\mu(s,a))^2/2\)$ is the normalization factor. We use a fixed truncation for every state-action pair regardless of $\mu(s,a)$ because otherwise the KL divergence of two instances will easily be infinity.
Without loss of generality, we assume $\calS$ is partitioned into $\calS_1,\cdots,\calS_H$ where for any state $s\in\calS_h$ and action $a\in\calA$, $\supp(p(s,a))\subseteq \calS_{h+1}$. We also assume that the initial state $s_1$ is fixed.

A decision (a.k.a. policy) $\pi:\calS\to\calA$ is a deterministic mapping from a state to an action, and $\Pi=\calA^\calS$ is the set of possible decisions with $|\Pi|= |\calA|^{|\calS|}<\infty.$ An observation $\ob$ is a trajectory $(s_1,a_1,r_1,\cdots,s_H,a_H,r_H)$. The reward of an observation is $R(\ob)=\sum_{h=1}^{H}r_h.$ We emphasize that $s_h,a_h$ are discrete random variables and $r_h\in\R$ is a continuous random variable. In the following, we use $\obsa=(s_1,a_1,\cdots,s_H,a_H)$ to denote the set of state-action pairs in $\ob$ and $\obr=(r_1,\cdots,r_H)$ the set of rewards. As a basic property of MDP, the rewards are independent conditioned on $\obsa$, and $r_h\sim r(s_h,a_h)$.

An instance $f$ is a represented by the transition function $p$ and reward function $r$ (which is uniquely determined by $\mu:\calS\times\calA\to[-1,1]$). The family of instances $\calF$ is defined as the set of all possible instances.

For an instance $f$ and a decision $\pi$, the density of an observation $\ob=(s_1,a_1,r_1,\cdots,s_H,a_H,r_H)$ is
\begin{align}
	f[\pi](\ob)=\prod_{h=1}^{H}p[s_h,a_h](s_{h+1})r[s_h,a_h](r_h).
\end{align} 
For simplicity, we also use $f_p[s,a](s')$ to denote the distribution $p[s,a](s')$, and $f_r[s,a](r)$ to denote $r[s,a](r)$.

By definition, for all $s,a,r,s'$ we have $p[s,a](s')r[s,a](r)<1.$ We also define
\begin{align}
	 f[\pi](\obsa)=\prod_{h=1}^{H}p[s_h,a_h](s_{h+1})
\end{align}
to be the marginal density of $\obsa,$ and $$f[\pi](\obr\mid \obsa)=\prod_{h=1}^{H}r[s_h,a_h](r_h)$$ the conditional distribution of $\obr.$

For an instance $f\in\calF$, we define $\mumin(f)=\min_{\obsa,\pi:f[\pi](\obsa)>0}f[\pi](\obsa).$ Since $\obsa,\pi$ are finite, we have $\mumin(f)>0$.

\subsection{Proof of Theorem~\ref{thm:rl-condition}}\label{app:pf-rl-condition}
In this section, we prove Theorem~\ref{thm:rl-condition}.
\begin{proof} In the following, we prove the three conditions separately.
	\paragraph{Proof of Condition~\ref{cond:uniform-convergence}.} We prove this condition by invoking Lemma~\ref{lem:uniform-convergence-rl}. For instances $f,g\in\calF$ and state-action pair $s,a\in\calS\times\calA$, let $f_r[s,a],g_r[s,a]$ be their reward distributions respectively. Since $f_r[s,a],g_r[s,a]$ are truncated Gaussian distributions, we get $\exp(-3)\le g_r[s,a](x)\le 1$ for every $x\in\supp(f_r[s,a])$ and $g\in\calF$. Therefore $\sup_x\abs{\ln \frac{f_r[s,a](x)}{g_r[s,a](x)}}\le 3$ for all $s\in\calS,a\in\calA.$ By Lemma~\ref{lem:uniform-convergence-rl} with $c_M=3$, we get Condition~\ref{cond:uniform-convergence}.
	
	\paragraph{Proof of Condition~\ref{cond:covering-number}.} The first part of this condition is proved by Lemma~\ref{lem:covering-rl}. On the other hand, we have
	\begin{align}
		\int 1\dd \ob=(2|\calS||\calA|)^H<\infty.
	\end{align}

	\paragraph{Proof of Condition~\ref{cond:TV-vs-inf}.} To prove the first part of this condition, recall that \begin{align}
		\truef[\pi](\ob)=\prod_{h=1}^{H}p[s_h,a_h](s_{h+1})r[s_h,a_h](r_h),
	\end{align}
	where $p$ is the transition function and $r$ is the reward distribution.
	Let $$\const_{\rm min}=\(\min_{s,a\in\calS\times\calA,s'\in\supp p[s,a](\cdot)}p[s,a](s')\)^H\exp(-4H),$$
	As a result, 
	$\truef[\pi](\ob)> \const_{\rm min}$ for all $\ob\in\supp(\truef[\pi])$.
	We prove the second part of this condition in Lemma~\ref{lem:TV-vs-inf-rl}.
\end{proof}

\subsection{Proof of Condition~\ref{cond:uniform-convergence}}\label{app:pf-cond-uc}
In this section, we present a lemma that establishes Condition~\ref{cond:uniform-convergence} for tabular RL.
\begin{lemma}\label{lem:uniform-convergence-rl}
	Consider any fixed RL instance $f$ with discrete state, action and general reward distribution. Suppose there exists a constant $c_M>0$ such that for any $g\in\calF,s\in\calS,a\in\calA$, the reward distributions of instance $f$ and $g$ at state $s$ and action $a$ (denoted by $f_r[s,a],g_r[s,a]$ respectively) satisfy
	\begin{align}
		\E_{x\sim f_r[s,a]}\[\(\ln \frac{f_r[s,a](x)}{g_r[s,a](x)}\)^4\]\le c_M^4.
	\end{align}
	Then for every $\alpha>0,\epsilon\in(0,1)$, Condition~\ref{cond:uniform-convergence} holds with
	\begin{align}
		\lambda_0(\alpha,\epsilon,f)=\frac{\epsilon}{32H^2}\min\left\{\frac{\mumin(f)}{4\alpha},\frac{\mumin(f)}{10},\frac{1}{\const_M}\right\}^{2}.
	\end{align}
\end{lemma}
\begin{proof}[Proof of Lemma~\ref{lem:uniform-convergence-rl}]
	Let 
	\begin{align}
		\kappa=\frac{\mumin(f)}{e^2}\exp\(-\frac{2\alpha}{\mumin(f)}\).
	\end{align}
	Recall that for reinforcement learning, an observation $\ob=(s_1,a_1,r_1,\cdots,s_H,a_H,r_H)$ represents a trajectory, and $\obsa=(s_1,a_1,s_2,a_2,\cdots,s_H,a_H)$ denotes the state-action pairs in the trajectory. Consider any fixed decision $\pi$, for any $\lambda<\lambda_0$, we prove the following two cases separately.
	
	\paragraph{Case 1: $\min_{\obsa:f[\pi](\obsa)>0}g[\pi](\obsa)<\kappa.$} In this case, we prove that $D_{1-\lambda}(f[\pi]\|g[\pi])\ge \alpha.$
	
	Let $\hat{\obsa}=\argmin_{\obsa:f[\pi](\obsa)>0}g[\pi](\obsa).$ By the condition of this case we have $g[\pi](\hat{\obsa})<\kappa.$ 
	Applying Lemma~\ref{lem:renyi-condition} we get
	\begin{align}\label{equ:uniform-convergence-1}
		D_{1-\lambda}(f[\pi](\ob)\|g[\pi](\ob))\ge D_{1-\lambda}(f[\pi](\obsa)\|g[\pi](\obsa)).
	\end{align}
	In the following we prove the RHS of Eq.~\eqref{equ:uniform-convergence-1} is larger than $\alpha$. We start with H\"older's inequality and the basic inequality that $(1-x)^{t}\le 1-t x$ for any $t\in (0,1), x\in (0,1)$:
	\begin{align}
		&\sum_\obsa f[\pi](\obsa)^{1-\lambda}g[\pi](\obsa)^{\lambda}\\
		=\; &\sum_{\obsa\neq \hat{\obsa}} f[\pi](\obsa)^{1-\lambda}g[\pi](\obsa)^{\lambda}+f[\pi](\hat\obsa)^{1-\lambda}g[\pi](\hat\obsa)^{\lambda}\\
		\le\; &\(\sum_{\obsa\neq \hat{\obsa}} f[\pi](\obsa)\)^{1-\lambda}\(\sum_{\obsa\neq \hat{\obsa}}g[\pi](\obsa)\)^{\lambda}+f[\pi](\hat\obsa)^{1-\lambda}g[\pi](\hat\obsa)^{\lambda}\\
		\le\;&\(1-f[\pi](\hat\obsa)\)^{1-\lambda}\(1-g[\pi](\hat\obsa)\)^{\lambda}+f[\pi](\hat\obsa)^{1-\lambda}g[\pi](\hat\obsa)^{\lambda}\\
		\le\;&\(1-f[\pi](\hat\obsa)\)^{1-\lambda}+f[\pi](\hat\obsa)^{1-\lambda}g[\pi](\hat\obsa)^{\lambda}\\
		\le\;&1-f[\pi](\hat\obsa)(1-\lambda)+f[\pi](\hat\obsa)^{1-\lambda}g[\pi](\hat\obsa)^{\lambda}.
	\end{align}
	Recall the basic inequality $\ln (1+x)\le x$ for all $x>-1$. Therefore,
	\begin{align}
		&\frac{1}{\lambda}\ln \(\sum_\obsa f[\pi](\obsa)^{1-\lambda}g[\pi](\obsa)^{\lambda}\)\\
		\le\;&\frac{1}{\lambda}\(-f[\pi](\hat\obsa)(1-\lambda)+f[\pi](\hat\obsa)^{1-\lambda}g[\pi](\hat\obsa)^{\lambda}\)\\
		\le\;&\frac{1}{\lambda}\(f[\pi](\hat\obsa)\(\(\frac{g(\hat\obsa)}{f(\hat\obsa)}\)^{\lambda}-1\)+\lambda f[\pi](\hat\obsa)\)\\
		\le\;&\frac{1}{\lambda}\(f[\pi](\hat\obsa)\(\(\frac{\kappa}{f(\hat\obsa)}\)^{\lambda}-1\)+\lambda f[\pi](\hat\obsa)\).
	\end{align}
	Recall the basic inequality that $\exp(x)\le 1+x/2$ for all $-1\le x\le 0$. Since we have $$\(\frac{\kappa}{f[\pi](\hat\obsa)}\)^{\lambda}=\exp\(\lambda\ln\(\frac{\kappa}{f[\pi](\hat\obsa)}\)\),$$ when $\lambda\le \(\ln\(f[\pi](\hat\obsa)/\kappa\)\)^{-1}$ we get
	\begin{align}
		&\frac{1}{\lambda}\(f[\pi](\hat\obsa)\(\(\frac{\kappa}{f[\pi](\hat\obsa)}\)^{\lambda}-1\)+\lambda f[\pi](\hat\obsa)\)\\
		\le\;&\frac{1}{2}f[\pi](\hat\obsa)\ln(\kappa/f[\pi](\hat\obsa))+f[\pi](\hat\obsa)=\frac{1}{2}f[\pi](\hat\obsa)\ln(e^2\kappa/f[\pi](\hat\obsa)).
	\end{align}
	By the definition of $\kappa$ we get
	\begin{align}
		\frac{1}{2}f[\pi](\hat\obsa)\ln(e^2\kappa/f[\pi](\hat\obsa))\le -\alpha,
	\end{align}
	which leads to $D_{1-\lambda}(f[\pi](\obsa)\|g[\pi](\obsa))\ge \alpha.$
	
	\paragraph{Case 2: $\min_{\obsa:f[\pi](\obsa)>0}g[\pi](\obsa)\ge \kappa.$} By Lemma~\ref{lem:KL-renyi-difference}, for any $\lambda\in\(0,1/2\)$ we get
	\begin{align}
		&\KL(f[\pi]\|g[\pi])-D_{1-\lambda}(f[\pi]\|g[\pi])\le \frac{\lambda}{2}\E_{\ob\sim f[\pi]}\[\(\ln \frac{f[\pi](\ob)}{g[\pi](\ob)}\)^4\]^{1/2}.
	\end{align}
	Let $f_r,g_r:\calS\times\calA\to \Delta(\R)$ be the reward distributions of instance $f$ and $g$ respectively, and $f_r[s,a](\cdot),g_r[s,a](\cdot)$ the densities of the reward given state-action pair $(s,a)$. Recall that for a trajectory $o=(s_1,a_1,r_1,\cdots,s_H,a_H,r_H)$ we have
	\begin{align}
		f[\pi](\ob)=f[\pi](\obsa)\prod_{h=1}^{H}f_r[s_h,a_h](r_h).
	\end{align}
	By H\"older's inequality we get
	\begin{align}
		&\E_{\ob\sim f[\pi]}\[\(\ln \frac{f[\pi](\ob)}{g[\pi](\ob)}\)^4\]\\
		=\;&\E_{\ob\sim f[\pi]}\[\(\ln\frac{f[\pi](\obsa)}{g[\pi](\obsa)}+\sum_{h=1}^{H}\ln \frac{f_r[s_h,a_h](r_h)}{g_r[s_h,a_h](r_h)}\)^4\]\\
		=\;&\E_{\ob\sim f[\pi]}\[(H+1)^3\(\(\ln\frac{f[\pi](\obsa)}{g[\pi](\obsa)}\)^4+\sum_{h=1}^{H}\(\ln \frac{f_r[s_h,a_h](r_h)}{g_r[s_h,a_h](r_h)}\)^4\)\]\\
		\le\;&(H+1)^3\ln(1/\kappa)^4+(H+1)^4\sup_{s\in\calS,a\in\calA}\E_{x\sim f_r[s,a]}\[\(\ln \frac{f_r[s,a](x)}{g_r[s,a](x)}\)^4\]\\
		\le\;&(H+1)^3\ln(1/\kappa)^4+(H+1)^4\const_M^4,
	\end{align}
	where the last inequality comes from item (c) of Condition~\ref{cond:rl-reward}. Therefore, when $\lambda\le\epsilon (H+1)^{-2}\min\{\ln(1/\kappa)^{-2},\const_M^{-2}\}$ we get
	\begin{align}
		&\KL(f[\pi]\|g[\pi])-D_{1-\lambda}(f[\pi]\|g[\pi])\le \frac{\lambda}{2}\E_{\ob\sim f[\pi]}\[\(\ln \frac{f[\pi](\ob)}{g[\pi](\ob)}\)^4\]^{1/2}\\
		\le\;&\frac{\epsilon}{2}(H+1)^{-2}\min\{\ln(1/\kappa)^{-2},\const_M^{-2}\}\((H+1)^3\ln(1/\kappa)^4+(H+1)^4\const_M^4\)^{1/2}\\
		\le\;&\frac{\epsilon}{2}\min\{\ln(1/\kappa)^{-2},\const_M^{-2}\}\(\ln(1/\kappa)^4+\const_M^4\)^{1/2}\\
		\le\;&\frac{\epsilon}{2}\min\{\ln(1/\kappa)^{-2},\const_M^{-2}\}\(\ln(1/\kappa)^2+\const_M^2\)\\
		\le\;&\epsilon.
	\end{align} 
	Recall that 
	\begin{align}
		\kappa=\frac{\mumin(f)}{e^2}\exp\(-\frac{2\alpha}{\mumin(f)}\).
	\end{align}
	By algebraic manipulation we get
	\begin{align}
		\frac{\epsilon}{32H^2}\min\left\{\frac{\mumin(f)}{4\alpha},\frac{\mumin(f)}{10},\frac{1}{\const_M}\right\}^{2}\le \epsilon (H+1)^{-2}\min\{\ln(1/\kappa)^{-2},\const_M^{-2}\},
	\end{align}
	which proves the desired result.
\end{proof}

\subsection{Proof of Condition~\ref{cond:covering-number}}\label{app:pf-cond-covering}
Recall that for two instances $f,g\in\calF$, their distance is $d(f,g)=\sup_{\pi,\ob}\abs{f[\pi](\ob)-g[\pi](\ob)}.$
\begin{lemma}\label{lem:covering-rl}
	Suppose $\calF$ represents tabular RL with truncated Gaussian reward with state space $\calS$ and action space $\calA$, then we have $\ln\calN(\calF,\epsilon)\le \bigO(|\calS||\calA|\ln (1/\epsilon))$ for every $\epsilon>0$.
\end{lemma}
\begin{proof}
	For instances $f\in\calF$ and state-action pair $s,a\in\calS\times\calA$, let $f_r[s,a]$ be its reward distribution and $f_p[s,a]$ its transition. 
	Recall that for tabular RL we have
	\begin{align}
		f[\pi](\ob)=\prod_{h=1}^{H}f_p[s_h,a_h](s_{h+1})f_r[s_h,a_h](r_h),
	\end{align}
	where the observation is $\ob=(s_1,a_1,r_1,\cdots,s_H,a_H,r_H)$. Consequently,
	\begin{align*}
		&\abs{f[\pi](\ob)-g[\pi](\ob)}\\
		\le &\sum_{h=1}^{H}\Bigg(\abs{f_p[s_h,a_h](s_{h+1})f_r[s_h,a_h](r_h)-g_p[s_h,a_h](s_{h+1})g_r[s_h,a_h](r_h)}\\
		&\qquad\prod_{h'=1}^{h-1}f_p[s_{h'},a_{h'}](s_{h'+1})f_r[s_{h'},a_{h'}](r_{h'})\prod_{h'=h+1}^{H}g_p[s_{h'},a_{h'}](s_{h'+1})g_r[s_{h'},a_{h'}](r_{h'})\Bigg)\\
		\le&\sum_{h=1}^{H}\abs{f_p[s_h,a_h](s_{h+1})f_r[s_h,a_h](r_h)-g_p[s_h,a_h](s_{h+1})g_r[s_h,a_h](r_h)}\\
		\le&\sum_{h=1}^{H}\(\abs{f_p[s_h,a_h](s_{h+1})-g_p[s_h,a_h](s_{h+1})}+\abs{f_r[s_h,a_h](r_h)-g_r[s_h,a_h](r_h)}\)
	\end{align*}
	Therefore, we construct a covering $\calC$ such that for every $f\in\calF$, there exists $g\in\calC$ with 
	\begin{align}
		\abs{f_p[s,a](s')-g_p[s,a](s')}\le \frac{\epsilon}{2H},\quad\forall s,a,s',\\
		\abs{f_r[s,a](r)-g_r[s,a](r)}\le \frac{\epsilon}{2H},\quad\forall s,a,r.
	\end{align}
	Since $f_p[s,a]$ is a discrete distribution, the covering number for the transition function is upper bounded by $(4H/\epsilon)^{|\calS|^2|\calA|}.$ On the other hand, $f_r[s,a](r)$ is a truncated Gaussian distribution, so covering number for the reward is also upper bounded by $(4H/\epsilon)^{|\calS||\calA|}.$ As a result,
	\begin{align}
		\ln\calN(\calF,\epsilon)\le \bigO(|\calS||\calA|\ln (1/\epsilon)).
	\end{align}
\end{proof}

\subsection{Proof of Condition~\ref{cond:TV-vs-inf}}\label{app:pf-cond-tv}
\begin{lemma}\label{lem:TV-vs-inf-rl}
	Consider tabular reinforcement learning with truncated Gaussian reward. For a fixed instance $\truef$, for all $f\in\calF,\pi\in\Pi$ we have
	\begin{align}
		\|\truef[\pi]-f[\pi]\|_\infty\le \frac{37H}{\mumin(\truef)^{1/6}}\KL(\truef[\pi]\|f[\pi])^{1/6}.
	\end{align}
\end{lemma}
\begin{proof}
	Recall that for tabular RL, an observation is $\ob=(s_1,a_1,r_1,\cdots,s_H,a_H,r_H)$. We use $\obsa=(s_1,a_1,s_2,a_2,\cdots,s_H,a_H)$ to denote the collection of states and actions in the trajectory $\ob$, and $\obr=(r_1,\cdots,r_H)$ the collection of of rewards.For instances $f\in\calF$ and state-action pair $s,a\in\calS\times\calA$, let $f_r[s,a]$ be its reward distribution and $f_p[s,a]$ its transition. 
	
	Let $\epsilon_0=\frac{H}{\mumin(\truef)^{1/6}}.$ Consider the random variables $\obsa$ and $\obr.$ By the chain rules of KL divergence we have
	\begin{align}\label{equ:pct-1}
		\KL(\truef[\pi]\|f[\pi])=\KL(\truef[\pi](\obsa)\|f[\pi](\obsa))+\E_{\obsa\sim \truef}\[\KL(\truef[\pi](\obr\mid \obsa)\|f[\pi](\obr\mid \obsa))\].
	\end{align}
	Since $\obsa$ is a discrete random variable, we have
	\begin{align}
		\abs{\truef[\pi](\obsa)-f[\pi](\obsa)}\le \TV((\truef[\pi]\|f[\pi])\le \KL(\truef[\pi]\|f[\pi])^{1/2}.
	\end{align}
	Therefore, for any $\obsa\not\in \supp(\truef[\pi])$,
	\begin{align}
		\abs{\truef[\pi](\obsa,\obr)-f[\pi](\obsa,\obr)}\le f[\pi](\obsa,\obr)\le f[\pi](\obsa)f[\pi](\obr\mid \obsa)\le \KL(\truef[\pi]\|f[\pi])^{1/2},
	\end{align}
	where the last inequality comes from the fact that $f[\pi](\obr\mid \obsa)=\prod_{(s_h,a_h,r_h)\in\ob}f_r[s_h,a_h](r_h)\le 1.$
	
	Now, for any $\obsa\in \supp(\truef[\pi])$, by Eq.~\eqref{equ:pct-1} we get
	\begin{align}
		\KL(\truef[\pi](\obr\mid \obsa)\|f[\pi](\obr\mid \obsa))\le \frac{1}{\mumin(\truef)}\KL(\truef[\pi]\|f[\pi]).
	\end{align}
	By the chain rule of KL divergence,
	\begin{align}
		\KL(\truef[\pi](\obr\mid \obsa)\|f[\pi](\obr\mid \obsa))=\sum_{h=1}^{H}\KL(\truef_r[s_h,a_h]\|f_r[s_h,a_h]).
	\end{align}
	As a result, for every $(s,a)\in\obsa$ we get
	\begin{align}
		\KL(\truef_r[s,a]\|f_r[s,a])\le \frac{1}{\mumin(\truef)}\KL(\truef[\pi]\|f[\pi]).
	\end{align}
	By Lemma~\ref{lem:closeness-truncated-Gaussian}, $\abs{\truef_r[s,a](r)-f_r[s,a](r)}\le 36 \(\frac{1}{\mumin(\truef)}\KL(\truef[\pi]\|f[\pi])\)^{1/6}.$ Therefore,
	\begin{align}
		&\abs{\truef[\pi](\obr\mid \obsa)-f[\pi](\obr\mid \obsa)}\\
		=\;&\abs{\prod_{h=1}^{H}\truef_r[s_h,a_h](r_h)-\prod_{h=1}^{H}f_r[s_h,a_h](r_h)}\\
		\le\;&\sum_{h=1}^{H}\abs{\truef_r[s_h,a_h](r_h)-f_r[s_h,a_h](r_h)}\\
		\le\;& 36H \(\frac{1}{\mumin(\truef)}\KL(\truef[\pi]\|f[\pi])\)^{1/6}.
	\end{align}
	It follows that,
	\begin{align}
		&\abs{\truef[\pi](\obsa,\obr)-f[\pi](\obsa,\obr)}\\
		\le\;& \abs{\truef[\pi](\obsa)-f[\pi](\obsa)}+\abs{\truef[\pi](\obr\mid \obsa)-f[\pi](\obr\mid \obsa)}\\
		\le\;& \KL(\truef[\pi]\|f[\pi])^{1/2}+36H \(\frac{1}{\mumin(\truef)}\KL(\truef[\pi]\|f[\pi])\)^{1/6}\\
		\le\;&37H \(\frac{1}{\mumin(\truef)}\KL(\truef[\pi]\|f[\pi])\)^{1/6}
	\end{align}
\end{proof}

\section{RL with General Reward Distribution}\label{app:rl-Gaussian}
In this section, we extend our analysis to RL with general reward distribution, where the support of the reward may have infinite volume (e.g., the real line $\R$ when the reward distribution is Gaussian). We assume that for any state-action pair $(s,a)$, the reward distribution $r[s,a]$ comes from a distribution family $\calR$ (e.g., $\calR=\{\calN(\mu,1):\mu\in[-1,1]\}$ when the reward is Gaussian).
For any $g\in\calR$, let $g(\cdot)$ be its density function and $\mu(g)\defeq \E_{x\sim g}[x]$ its mean and we assume $\sup_{g\in\calR}\mu(g)\le \Rmax.$
We emphasize that for general reward distributions, we do not require Conditions~\ref{cond:covering-number} and~\ref{cond:TV-vs-inf} because they do not hold for Gaussian distribution. Instead, we require the following.
\begin{cond}\label{cond:rl-reward}
	Let $\calR$ be the reward distribution family. Then
	\begin{enumerate}[label=(\alph*)]
		\item for all $f\in\calR$, there exists a constant $\const_7\in(0,1],\const_8\in(0,1],\const_M>0,$ such that for every $\delta>0$, 
		\begin{align}
			\Pr_{x\sim f}\(\forall g,g'\in\calR,\; \abs{\ln \frac{g(x)}{g'(x)}}>\TV(g\|g')^{\const_7}\polylog(1/\delta)\)\le \delta;
		\end{align}
		\item for all $g,g'\in\calR$, $\abs{\mu(g)-\mu(g')}\lesssim \TV(g\|g')^{\const_8}$;
		\item for all $g,g'\in\calR$, $$\E_{x\sim g}\[\(\ln\frac{g(x)}{g'(x)}\)^4\]\le c_M^4;$$
		\item for any $\epsilon>0$, there exists a covering $\calC(\calR,\epsilon)\subseteq\calR$ such that 
		\begin{align}
			\forall g\in\calR,\;\exists g'\in\calC(\calR,\epsilon),\;\TV(g\|g')\le \epsilon.
		\end{align}
		And $\log|\calC(\calR,\epsilon)|=\bigO(\log(1/\epsilon)).$
	\end{enumerate}
\end{cond}
Condition~\ref{cond:uniform-convergence} still holds in this case because of Lemma~\ref{lem:uniform-convergence-rl} and item (c) of Condition~\ref{cond:rl-reward}.

For tabular reinforcement learning problems, we only require Condition~\ref{cond:rl-reward} for the reward distribution of any fixed state-action pair, which is one-dimensional. Condition~\ref{cond:rl-reward} holds for Gaussian distribution (see Proposition~\ref{prop:rl-reward-Gaussian}), Laplace distribution, etc.

Our main result for tabular RL is stated as follows.
\begin{theorem}\label{thm:rl-main}
	Suppose $\calF$ is the hypothesis class representing tabular reinforcement learning with a reward distribution that satisfies Conditions~\ref{cond:rl-reward}, then the regret of Alg.~\ref{alg:main-finite} satisfies
	\begin{align}
		\limsup_{n\to\infty}\frac{\reg_{\truef,n}}{\ln n}\le \comp(\truef).
	\end{align}
\end{theorem}
In section~\ref{sec:rl-general-main-lemmas} we present the main lemmas for Steps 1-2. Then, Theorem~\ref{thm:rl-main} is proved using exactly the same approach as Theorem~\ref{thm:main} (by replacing the main lemmas for Steps 1-2).

\subsection{Lemmas for RL with General Reward Distribution}\label{sec:rl-general-main-lemmas}
In this section, we state the main lemmas for RL with general reward distribution (namely, Lemma~\ref{lem:main-init-RL} for Step 1 and Lemma~\ref{lem:main-ident-RL} for Step 2).
\begin{lemma}\label{lem:main-init-RL}
	Let $\init$ be the event that, there exists a universal constant $\const_4>0$ and a function $\iota(\truef)$ that only depends on $f^\star$ such that 
	\begin{enumerate}[label=(\alph*)]
		\item for all $\pi\in\Pi$, $\KL(\truef[\pi]\|\initf[\pi])\le \(\frac{\ln\ln n}{\ln n}\)^{\const_4}$,
		\item $\abs{R_{\initf}(\pi)-R_{\truef}(\pi)}\lesssim 2H\Rmax\(\frac{\ln\ln n}{\ln n}\)^{\const_4}$, for all $\pi\in\Pi$,
		\item $\pi^\star(\initf)=\pi^\star(\truef).$
	\end{enumerate}
	For tabular reinforcement learning with general reward distribution that satisfies Condition~\ref{cond:rl-reward}, there exists $n_0>0$ such that when $n>n_0$, $\Pr(\init)\ge 1-1/\ln n.$ In addition, the regret of Step 1 is upper bounded by $\bigO(\frac{\ln n}{\ln\ln n}).$
\end{lemma}
\begin{proof} In the following, we prove the three items separately. Recall that Condition~\ref{cond:uniform-convergence} holds by Lemma~\ref{lem:uniform-convergence-rl} and item (c) of Condition~\ref{cond:rl-reward}.
	
	\paragraph{Items (a) and (c):} 
	Proofs of (a) and (c) are the same as that of Lemma~\ref{lem:main-init} if we replace Lemma~\ref{lem:uniform-concentration} by Lemma~\ref{lem:uniform-concentration-Gaussian}.
	
	\paragraph{Item (b):} Recall that an observation is a sequence of state-action-reward tuples: $\ob=(s_1,a_1,r_1,\cdots,s_H,a_H,r_H)$. Therefore, for all $\pi\in\Pi$ we have
	\begin{align}
		&\abs{R_{\initf}(\pi)-R_{\truef}(\pi)}=\abs{\E_{\ob\sim \initf[\pi]}\[\sum_{h=1}^{H}r_h\]-\E_{\ob\sim \truef[\pi]}\[\sum_{h=1}^{H}r_h\]}\\
		\le\;&\abs{\E_{\obsa\sim \initf[\pi]}\[\sum_{h=1}^{H}\hat{\mu}(s_h,a_h)\]-\E_{\obsa\sim \truef[\pi]}\[\sum_{h=1}^{H}\mu^\star(s_h,a_h)\]},
	\end{align}
	where $\hat{\mu}$ and $\mu^\star$ are the mean reward of instance $\initf$,$\truef$ respectively. Since $\hat{\mu}\in[0,1],$ we have
	\begin{align}
		&\abs{\E_{\obsa\sim \initf[\pi]}\[\sum_{h=1}^{H}\hat{\mu}(s_h,a_h)\]-\E_{\obsa\sim \truef[\pi]}\[\sum_{h=1}^{H}\mu^\star(s_h,a_h)\]}\\
		\le\;&\abs{\E_{\obsa\sim \initf[\pi]}\[\sum_{h=1}^{H}\hat{\mu}(s_h,a_h)\]-\E_{\obsa\sim \truef[\pi]}\[\sum_{h=1}^{H}\hat\mu(s_h,a_h)\]}\\
		&\quad+\abs{\E_{\obsa\sim \truef[\pi]}\[\sum_{h=1}^{H}\hat{\mu}(s_h,a_h)\]-\E_{\obsa\sim \truef[\pi]}\[\sum_{h=1}^{H}\mu^\star(s_h,a_h)\]}\\
		\le\;&H\TV(\truef[\pi]\|\initf[\pi])+\E_{\obsa\sim \truef[\pi]}\[\sum_{h=1}^{H}|\hat\mu(s_h,a_h)-\mu^\star(s_h,a_h)|\].\label{equ:rGml-1}
	\end{align}
	In the following we upper bound the second term of Eq.~\eqref{equ:rGml-1}. Let $\truef_r,\initf_r:\calS\times\calA\to \Delta(\R)$ be the reward distributions of instance $\truef$ and $\initf$ respectively, and $\truef_r[s,a](\cdot),\initf_r[s,a](\cdot)$ the densities of the reward given state-action pair $(s,a)$. 
	By item (b) of Condition~\ref{cond:rl-reward} and Cauchy-Schwarz we get
	\begin{align}
		&\E_{\obsa\sim \truef[\pi]}\[\sum_{h=1}^{H}|\hat\mu(s_h,a_h)-\mu^\star(s_h,a_h)|\]\\
		\lesssim\;&\E_{\obsa\sim \truef[\pi]}\[\sum_{h=1}^{H}\KL(\truef_r[s_h,a_h]\|\initf_r[s_h,a_h])^{\const_8}\]\\
		\le\;&\E_{\obsa\sim \truef[\pi]}\[H^{1-\const_8}\(\sum_{h=1}^{H}\KL(\truef_r[s_h,a_h]\|\initf_r[s_h,a_h])\)^{\const_8}\]\\
		\le\;&H^{1-\const_8}\E_{\obsa\sim \truef[\pi]}\[\sum_{h=1}^{H}\KL(\truef_r[s_h,a_h]\|\initf_r[s_h,a_h])\]^{\const_8}.
	\end{align}
	Recall that $\truef_p,\initf_p$ denotes the transition function of instances $\truef,\initf$ respectively. By the chain rule of KL divergence, 
	\begin{align}
		&\KL(\truef[\pi]\|\initf[\pi])\\
		=\;&\E_{\obsa\sim \truef[\pi]}\[\sum_{h=1}^{H}(\KL\(\truef_p[s_h,a_h]\|\initf_p[s_h,a_h]\)+\KL(\truef_r[s_h,a_h]\|\initf_r[s_h,a_h]))\]\\
		\ge \;&\E_{\obsa\sim \truef[\pi]}\[\sum_{h=1}^{H}\KL(\truef_r[s_h,a_h]\|\initf_r[s_h,a_h])\].
	\end{align}
	By item (a) of this lemma, $\KL(\truef[\pi]\|\initf[\pi])\to 0$ as $n\to\infty$. Therefore for large enough $n$ we get
	\begin{align}
		&\abs{R_{\initf}(\pi)-R_{\truef}(\pi)}\lesssim H\TV(\truef[\pi]\|\initf[\pi])+H^{1-\const_8}\KL(\truef[\pi]\|\initf[\pi])^{\const_8}
		\le 2H\KL(\truef[\pi]\|\initf[\pi])^{\const_8}.
	\end{align}
	Combining with item (a) of this lemma, we prove the desired result.
\end{proof}

\begin{lemma}\label{lem:main-ident-RL}
	For tabular reinforcement learning with general reward distribution that satisfies Condition~\ref{cond:rl-reward}, there exists $n_0>0$ such that when $n>n_0$, the following holds.
	\begin{enumerate}[label=(\alph*)]
		\item Conditioned on the event $\pi^\star(\initf)\neq \pi^\star(f^\star)$, $\Pr(\acc)\le 1/n$.
		\item Conditioned on the event $\init$, $\Pr(\acc)\ge 1-1/\ln n$.
		\item The expected regret of Step 2 is always upper bounded by $\bigO(\ln n\ln\ln n).$
		\item Conditioned on the event $\init$, the expected regret of Step 2 is upper bounded by $$\(\comp(f^\star,(\ln\ln n)^{1/4}/2)+o(1)\)\ln n.$$
	\end{enumerate}
\end{lemma}
\begin{proof}	We prove the four items above separately.
	
	\paragraph{Items (a), (c), and (d):} Proofs of (a), (c), and (d) are the same as that of Lemma~\ref{lem:main-ident}.
	
	\paragraph{Item (b):} Let $\epsilon=1/\ln\ln n, \alpha=\frac{\ln n}{m}$ and $\delta=1/(2\ln n)$. We prove this statement by invoking Lemma~\ref{lem:uniform-concentration-Gaussian} with parameters $(\alpha+5\epsilon,\epsilon,w)$. Following Lemma~\ref{lem:uniform-concentration-Gaussian}, let $\gamma=\frac{1}{m}\min_{\pi} \sum_{i=1}^{m}\ind{\pi_i=\pi}$ and $\lambda=\lambda_0((\alpha+5\epsilon)/\gamma,\epsilon,\truef)$ be the value that satisfies Condition~\ref{cond:uniform-convergence}. Let $$\epsilon_0=\exp(-(\alpha+5\epsilon)/\gamma)(\epsilon\lambda)^{1/\lambda}.$$
	
	First of all, we prove that the condition for Lemma~\ref{lem:uniform-concentration-Gaussian} holds. That is,
	\begin{align}\label{equ:miG-1}
		m\gtrsim \frac{\poly(|\calS||\calA|H)}{\lambda\epsilon}\((\ln (m/\epsilon_0))+\ln(1/\delta)\).
	\end{align}
	Recall that $m=\sum_{x}\lceil \bar{w}_\pi\ln n\rceil$. As a result, $m\ge \frac{|\Pi|\ln n}{(\ln\ln n)^{1/4}}.$ Now consider the RHS of Eq.~\eqref{equ:miG-1}. By the definition of $\bar{w}_\pi$ we get $\alpha\le \frac{2}{|\Pi|}(\ln\ln n)^{1/4}$ and $\gamma^{-1}\le 2|\Pi|(\ln\ln n)^{1/2}$. It follows from Condition~\ref{cond:uniform-convergence} that $\lambda\ge \poly(1/\ln\ln n).$ By the definition of $\epsilon_0$ and Condition~\ref{cond:covering-number} we get
	\begin{align}
		\ln(m/\epsilon_0)\lesssim \poly(\ln\ln n).
	\end{align}
	Consequently, the RHS of Eq.~\eqref{equ:miG-1} is at most $\poly(\ln\ln n)$, and Eq.~\eqref{equ:miG-1} holds when $n$ is sufficiently large. By Lemma~\ref{lem:uniform-concentration-Gaussian} we get, with probability at least $1-1/(2\ln n)$,
	\begin{align}\label{equ:miG-2}
		\sum_{i=1}^{m}\ln \frac{\truef[\pi_i](\ob_i)}{g[\pi_i](\ob_i)}\ge (\alpha+\epsilon)m,\quad \forall g\in \calF(w,f^\star,\alpha+5\epsilon),
	\end{align}
	where $\calF(w,f^\star,\alpha+5\epsilon)=\{g\in\calF:\KL^w(\truef\|g)\ge \alpha+5\epsilon\}$.
	
	In the following, we prove that Eq.~\eqref{equ:miG-2} implies $\acc^{\initf}.$ Recall that $\acc^{\initf}$ is the event defined as follows:
	\begin{align}\label{equ:acc-initf-G}
		\acc^\initf&=\ind{\forall g\in \cset(\initf), \sum_{i=1}^{m}\ln\frac{\initf[\pi_i](\ob_i)}{g[\pi_i](\ob_i)}\ge \ln n}.
	\end{align}
	Next, we apply Lemma~\ref{lem:KL-closeness-f} to any $g\in\cset(\initf).$  To verify its condition, we have
	$\TV(\truef[\pi]\|\initf[\pi])\le \KL(\truef[\pi]\|\initf[\pi])^{1/2}\lesssim \(\frac{\ln\ln n}{\ln n}\)^{\const_4/2}$ for all $\pi\in\Pi$ by item (a) of Lemma~\ref{lem:main-init-RL}. Therefore when $n$ is large enough, 
	\begin{align}
		\KL^w(f^\star\|g)\ge \frac{\ln n}{m}+5\epsilon=\alpha+5\epsilon.
	\end{align}
	Then $g\in\calF(w,\truef,\alpha+5\epsilon)$. It follows from Eq.~\eqref{equ:miG-2} that 
	\begin{align}\label{equ:miG-4}
		\sum_{i=1}^{m}\ln \frac{\truef[\pi_i](\ob_i)}{g[\pi_i](\ob_i)}\ge (\alpha+\epsilon)m,\quad\forall g\in\cset(\initf).
	\end{align}
	
	By Lemma~\ref{lem:log-likelihood-ratio-general}, with probability at least $1-1/(2\ln n)$ we have
	\begin{align}\label{equ:miG-3}
		\sum_{i=1}^{m}\ln\frac{\initf[\pi_i](\ob_i)}{\truef[\pi_i](\ob_i)}\ge -m\(\frac{\ln\ln n}{\ln n}\)^{c_4c_6}\poly(\ln\ln n)\iota(\truef).
	\end{align}
	For large enough $n$, $\(\frac{\ln\ln n}{\ln n}\)^{c_4c_6}\poly(\ln\ln n)\iota(\truef)\le \frac{1}{\ln\ln n}=\epsilon.$
	As a result, combining Eq.~\eqref{equ:miG-4} and Eq.~\eqref{equ:miG-3}, with probability $1-1/\ln n$ we have
	\begin{align}
		\forall g\in\cset(\initf),\quad \sum_{i=1}^{m}\ln \frac{\initf[\pi_i](\ob_i)}{g[\pi_i](\ob_i)}\ge \alpha m=\ln n.
	\end{align}
\end{proof}

The following lemma is used in the proof of Lemma~\ref{lem:main-ident-RL}.
\begin{lemma}\label{lem:log-likelihood-ratio-general}
	Let $\truef,\initf$ be any fixed tabular RL instances with reward distribution that satisfies Condition~\ref{cond:rl-reward}. For any sequence of policies $\{\pi_i\}_{i=1}^{m}$, let $\ob_i\sim \truef[\pi_i](\cdot),\forall i\in[m]$ be a sequence of random observations drawn from $\truef$. 
	Then there exists constant $\const_6>0$ and $\iota(\truef)$ that only depends on $\truef$ such that, for any $\delta>0$, with probability at least $1-\delta$,
	\begin{align}\label{equ:llrG-result}
		\frac{1}{m}\sum_{i=1}^{m}\ln\frac{\initf[\pi_i](\ob_i)}{\truef[\pi_i](\ob_i)}> -\(\max_\pi \KL(\truef[\pi]\|\initf[\pi])\)^{\const_6}\iota(\truef)\polylog(mH/\delta)
	\end{align}
	assuming $\max_\pi \KL(\truef[\pi]\|\initf[\pi])^{1/2}\le \mumin(\truef)/2.$
\end{lemma}
\begin{proof}
	Recall that for reinforcement learning, an observation $\ob=(s_1,a_1,r_1,\cdots,s_H,a_H,r_H)$ represents a trajectory, and $\obsa=(s_1,a_1,s_2,a_2,\cdots,s_H,a_H)$ denotes the state-action pairs in the trajectory. 
	By algebraic manipulation, we get
	\begin{align}
		\sum_{i=1}^{m}\ln \frac{\initf[\pi_i](\ob_i)}{\truef[\pi_i](\ob_i)}=\sum_{i=1}^{m}\ln \frac{\initf[\pi_i](\obsa_i)}{\truef[\pi_i](\obsa_i)}+\sum_{i=1}^{m}\ln \frac{\initf[\pi_i](\obr_i\mid \obsa_i)}{\truef[\pi_i](\obr_i\mid \obsa_i)}.
	\end{align}
	Recall that for the instance $\truef$, we have
	\begin{align}
		\mumin(\truef)=\min_\pi\min_{\obsa\in\supp(\truef[\pi](\cdot))}\truef[\pi](\obsa).
	\end{align}
	Since both $\obsa$ and $\pi$ are finite for tabular RL, we get $\mumin(\truef)>0.$ On the one hand, for any $\obsa\in\supp(\truef[\pi])$, by Pinsker's inequality we get
	\begin{align}
		\abs{\truef[\pi](\obsa)-\initf[\pi](\obsa)}\le \TV(\truef[\pi]\|\initf[\pi])\le \KL(\truef[\pi]\|\initf[\pi])^{1/2}.
	\end{align}
	As a result
	\begin{align*}
		\abs{\ln \frac{\initf[\pi](\obsa)}{\truef[\pi](\obsa)}}\le \abs{\ln \(1+\frac{\initf[\pi](\obsa)-\truef[\pi](\obsa)}{\truef[\pi](\obsa)}\)}\le \abs{\ln \(1+\frac{\initf[\pi](\obsa)-\truef[\pi](\obsa)}{\mumin(\truef)}\)}.
	\end{align*}
	When $\max_\pi \KL(\truef[\pi]\|\initf[\pi])^{1/2}\le \mumin(\truef)/2$, applying the basic inequality $\abs{\ln(1+x)}\le 2x,\forall |x|\le 1/2$ we get for all $\pi\in\Pi$ and $\obsa\in\supp(\truef[\pi](\cdot))$,
	\begin{align}
		&\abs{\ln \(1+\frac{\initf[\pi](\obsa)-\truef[\pi](\obsa)}{\mumin(\truef)}\)}\le \frac{2}{\mumin(\truef)}\abs{\truef[\pi](\obsa)-\initf[\pi](\obsa)}\\
		\le\;&\frac{2}{\mumin(\truef)}\max_\pi\KL(\truef[\pi]\|\initf[\pi])^{1/2}.
	\end{align}
	
	On the other hand, let $\truef_r[s,a]$ and $\initf_r[s,a]$ be the reward distribution of instance $\truef$ and $\initf$ given state-action pair $(s,a)$ respectively. Then
	\begin{align}
		&\sum_{i=1}^{m}\ln \frac{\initf[\pi_i](\obr_i\mid \obsa_i)}{\truef[\pi_i](\obr_i\mid \obsa_i)}
		=\sum_{i=1}^{m}\sum_{h=1}^{H}\ln \frac{\initf_{r}[s_{i,h},a_{i,h}](r_{i,h})}{\truef_r[s_{i,h},a_{i,h}](r_{i,h})}.
	\end{align}
	For any $\pi\in\Pi$, by the chain rule of KL divergence we get
	\begin{align}
		&\KL(\truef[\pi]\|\initf[\pi])\\
		=\;&\E_{\obsa\sim \truef[\pi]}\[\sum_{h=1}^{H}(\KL\(\truef_p[s_h,a_h]\|\initf_p[s_h,a_h]\)+\KL(\truef_r[s_h,a_h]\|\initf_r[s_h,a_h]))\]\\
		\ge\;&\mumin(\truef)\KL(\truef_r[s,a]\|\initf_r[s,a])\ind{(s,a)\in\obsa\text{ for some }\obsa\in\supp(\truef[\pi])}.
	\end{align}
	Because $\ob_i\sim \truef[\pi_i]$, for any $i\in[m],h\in[H]$ we have $(s_{i,h},a_{i,h})\in\obsa_i$ and $\obsa_i\in\supp(\truef[\pi_i])$. As a result, for all $i\in[m],h\in[H]$, item (a) of Condition~\ref{cond:rl-reward} implies that with probability at least $1-\delta/(mH)$
	\begin{align}
		&\abs{\frac{\initf_{r}[s_{i,h},a_{i,h}](r_{i,h})}{\truef_r[s_{i,h},a_{i,h}]}}\le \TV(\truef_r[s_{i,h},a_{i,h}](r_{i,h})\|\initf_{r}[s_{i,h},a_{i,h}])^{\const_7}\polylog(mH/\delta)\\
		\le\;&\KL(\truef_r[s_{i,h},a_{i,h}]\|\initf_{r}[s_{i,h},a_{i,h}])^{\const_7/2}\polylog(mH/\delta)\\
		\le\;&\frac{1}{\mumin(\truef)^{\const_7/2}}\KL(\truef[\pi]\|\initf[\pi])^{\const_7/2}\polylog(mH/\delta).
	\end{align}
	Let $\epsilon=\frac{1}{\mumin(\truef)^{\const_7/2}}\(\max_\pi \KL(\truef[\pi]\|\initf[\pi])\)^{\const_7/2}$. Apply union bound and we get
	\begin{align}
		\Pr_{\ob\sim \truef[\pi]}\(\frac{1}{m}\sum_{i=1}^{m}\sum_{h=1}^{H}\abs{\ln \frac{\initf_{r}[s_{i,h},a_{i,h}](r_{i,h})}{\truef_r[s_{i,h},a_{i,h}](r_{i,h})}}\ge \epsilon H\polylog(mH/\delta)\)\le 1-\delta.
	\end{align}
	It follows that with probability at least $1-\delta$,
	\begin{align*}
		&\frac{1}{m}\sum_{i=1}^{m}\ln \frac{\initf[\pi_i](\ob_i)}{\truef[\pi_i](\ob_i)}\\
		\ge\;& -\frac{2}{\mumin(\truef)}\max_\pi\KL(\truef[\pi]\|\initf[\pi])^{1/2}-H\polylog(mH/\delta)\frac{1}{\mumin(\truef)^{\const_7/2}}\(\max_\pi \KL(\truef[\pi]\|\initf[\pi])\)^{\const_7/2}\\
		\ge\;& -\frac{2H}{\mumin(\truef)}\polylog(mH/\delta)\(\max_\pi \KL(\truef[\pi]\|\initf[\pi])\)^{\const_7/2}.
	\end{align*}
	Therefore, Eq.~\eqref{equ:llrG-result} is satisfied by setting 
	\begin{align}
		\iota(\truef)=\frac{2H}{\mumin(\truef)},\quad c_6=\const_7/2.
	\end{align}
\end{proof}

\subsection{Proof of Condition~\ref{cond:rl-reward} for Gaussian Distribution}
In this section, we prove that the reward distribution family $\calR=\{\calN(\mu,1),\mu\in[0,1]\}$ satisfies Condition~\ref{cond:rl-reward}.
\begin{proposition}\label{prop:rl-reward-Gaussian}
	Condition~\ref{cond:rl-reward} holds for the reward distribution family $\calR=\{\calN(\mu,1),\mu\in[0,1]\}$
\end{proposition}
\begin{proof}
	In the following we prove the four items of Condition~\ref{cond:rl-reward} respectively.
	\paragraph{Item (a).} For any fixed $f\in\calR$, let $\mu$ be the mean of $f$. In other words, $f=\calN(\mu,1).$ Then we have
	\begin{align}
		\Pr_{x\sim f}\(|x|>\mu+2\sqrt{\log(2/\delta)}\)\le \delta.
	\end{align}
	By definition, for any $g=\calN(\mu_g,1),g'=\calN(\mu_g',1)\in\calR$, we have $$\ln\frac{g(x)}{g'(x)}=\frac{1}{2}\((x-\mu_g)^2-(x-\mu_g')^2\)=(\mu_g'-\mu_g)(x-(\mu_g'+\mu_g)/2).$$ Therefore when $\mu_g,\mu_g'\in[0,1]$ we get
	\begin{align}
		\abs{\ln\frac{g(x)}{g'(x)}}\le |\mu_g-\mu_g'||x+1|.
	\end{align}
	As a result, with probability at least $1-\delta$ we have
	\begin{align}
		\forall g,g'\in\calR,\quad \abs{\ln\frac{g(x)}{g'(x)}}\le |\mu_g-\mu_g'|\(1+2\sqrt{\log(2/\delta)}\).
	\end{align}
	In addition, when $\mu_g,\mu_g'\in[0,1]$ we have $\TV(g\|g')\ge \frac{1}{10}|\mu_g-\mu_g'|.$ As a result, item (a) holds with $\const_7=1$.
	
	\paragraph{Item (b).} Recall that for Gaussian distribution with unit variance, when $\mu_g,\mu_g'\in[0,1]$ we have $\TV(g\|g')\ge \frac{1}{10}|\mu_g-\mu_g'|.$ Therefore item (b) holds with $\const_8=1$.
	
	\paragraph{Item (c).} Recall that for any $g=\calN(\mu_g,1),g'=\calN(\mu_g',1)\in\calR$, $\ln\frac{g(x)}{g'(x)}=\frac{1}{2}\((x-\mu_g)^2-(x-\mu_g')^2\)$. Therefore when $\mu_g,\mu_g'\in[0,1]$ we get
	\begin{align}
		&\E_{x\sim g}\[\(\ln \frac{g(x)}{g'(x)}\)^4\]=\frac{1}{2}\E_{x\sim g}\[\((x-\mu_g)^2-(x-\mu_g')^2\)^4\]\\
		\le\;& \frac{(\mu_g'-\mu_g)^4}{2}\E_{x\sim g}\[(x-(\mu_g'+\mu_g)/2)^4\]\le 3(\mu_g'-\mu_g)^4.
	\end{align}
	Therefore, item (c) holds with $c_M=2.$
	
	\paragraph{Item (d).} For $g=\calN(\mu_g,1),g'=\calN(\mu_g',1)\in\calR$, we have
	\begin{align}
		\TV(g\|g')\le \sqrt{\KL(g\|g')/2}=\frac{|\mu_g-\mu_g'|}{2}.
	\end{align}
	Therefore, we can set $\calC(\calR,\epsilon)=\{\calN(k\epsilon,1):k\in\{-\lfloor 1/\epsilon\rfloor,-\lfloor 1/\epsilon\rfloor+1,\cdots,\lfloor 1/\epsilon\rfloor\}\}$. Then $\log|\calC(\calR,\epsilon)|\le \log \lceil 2/\epsilon\rceil=\bigO(\log(1/\epsilon)).$
\end{proof}
	\section{Uniform Concentration}\label{app:uniform-concentration}
In this section, we present the uniform concentration results.
\subsection{Uniform Concentration with $\ell_\infty$ Covering}
In this section we prove uniform concentration results with $\ell_\infty$ covering. For two instances $g,g'$, define the $\ell_\infty$ distance as
\begin{align}
	\|g-g'\|_\infty\defeq\sup_{\pi,\ob}\abs{g[\pi](\ob)-g'[\pi](\ob)}.
\end{align} 
Let $\calN(\calF,\epsilon)$ be the proper covering number of $\calF$ w.r.t. the distance $\ell_\infty.$ Then we have the following lemma.

\begin{lemma}\label{lem:uniform-concentration}
	Consider any fixed $\alpha>0,0<\epsilon<\alpha/2$, list of decisions $w=(\pi_1,\cdots,\pi_m)$, $f\in\calF$. Let $\gamma=\frac{1}{m}\min_{\pi\in\Pi} \sum_{i=1}^{m}\ind{\pi_i=\pi}$ and let $\lambda=\lambda_0(\alpha/\gamma,\epsilon,f)$ be the value that satisfies Condition~\ref{cond:uniform-convergence}. Let $\calF(w,f,\alpha)=\{g\in\calF:\KL^w(f\|g)\ge \alpha\}$ and define 
	$$\epsilon_0=\frac{\exp(-\alpha/\gamma)(\epsilon \lambda )^{1/\lambda}}{3\vol}.$$
	Then for any $\delta>0$, when $m\ge \frac{1}{\lambda\epsilon}\(\ln\calN\(\calF,\epsilon_0\)+\ln(1/\delta)\)$ we have
	\begin{align}\label{equ:uc-conclusion}
		\Pr_{\ob_i\sim f[\pi_i],\forall i}\(\forall g\in \calF(w,f,\alpha),\sum_{i=1}^{m}\ln \frac{f[\pi_i](\ob_i)}{g[\pi_i](\ob_i)}\ge (\alpha-4\epsilon)m\)\ge 1-\delta.
	\end{align}
\end{lemma}
\begin{proof}
	For any $g\in \calF$, we define a induced distribution $\hat{g}$ for all $\pi\in\Pi$:
	\begin{align}
		\hat{g}[\pi](\ob)=\frac{g[\pi](\ob)+\epsilon_0}{1+\epsilon_0 \vol}.
	\end{align}
	Because $\hat{g}[\pi](\ob)\ge 0$ and $$\|\hat{g}[\pi](\ob)\|_1=\int \frac{g[\pi](\ob)+\epsilon_0}{1+\epsilon_0 \vol}\dd x=\frac{1}{1+\epsilon_0 \vol}\int g[\pi](\ob)+\epsilon_0\dd x=1,$$ the induced distribution $\hat{g}[\pi]$ is a valid distribution.
	
	\paragraph{Properties of the induced covering.}
	Let $Z=\epsilon_0\vol$. Consider the minimum $\epsilon_0$ covering set $\calC(w,f,\alpha)$ of $\calF(w,f,\alpha).$ For any $g\in \calF(w,f,\alpha)$, let $g'\in \calC(w,f,\alpha)$ be its cover and $\hat{g}$ the induced distribution of $g'$. Now, we prove that $\hat{g}$ satisfies the following two properties:
	\begin{enumerate}[label=(\alph*)]
		\item For any $\pi\in\Pi$ and $\ob\in\supp f[\pi]$, $\ln\frac{f[\pi](\ob)}{g[\pi](\ob)}\ge \ln\frac{f[\pi](\ob)}{\hat{g}[\pi](\ob)}-\epsilon,$ and
		\item $D^w_{1-\lambda}(f\|\hat{g})\ge \alpha-2\epsilon$ for $g\in\calF(w,f,\alpha).$
	\end{enumerate}
	
	First we prove item (a). Since $g'$ is the $\epsilon_0$ cover of $g$, we get $g'[\pi](\ob)+\epsilon_0\ge g[\pi](\ob)$ for any $\pi\in\Pi$ and $x\in\supp f[\pi]$. As a result,
	\begin{align}
		&\ln\frac{f[\pi][\ob]}{g[\pi](\ob)}\ge \ln\frac{f[\pi][\ob]}{g'[\pi](\ob)+\epsilon_0}
		=\ln\frac{f[\pi][\ob]}{\hat{g}[\pi](\ob)}-\ln(1+Z).
	\end{align}
	By the basic inequality $\ln(1+x)\le x,\forall x\in(-1,\infty)$ we get
	\begin{align}
		\ln(1+Z)\le Z=\epsilon_0\vol\le \epsilon.
	\end{align}
	As a result, item (a) follows directly. Now we prove item (b). By algebraic manipulation,
	\begin{align}
		&\TV(g[\pi]\|\hat{g}[\pi])\le \int \abs{\hat{g}[\pi](\ob)-g[\pi](\ob)}\dd \ob=\int \abs{\frac{g'[\pi](\ob)+\epsilon_0}{1+\epsilon_0 \vol}-g[\pi](\ob)}\dd \ob\\
		\le&\;\int \abs{\frac{g'[\pi](\ob)+\epsilon_0}{1+\epsilon_0 \vol}-g'[\pi](\ob)-\epsilon_0}\dd \ob+\int \abs{g'[\pi](\ob)+\epsilon_0-g[\pi](\ob)}\dd \ob\\
		\le&\;\frac{1}{1+\epsilon_0 \vol}\int \abs{\epsilon_0 \vol \(g'[\pi](\ob)+\epsilon_0\)}\dd \ob+2\epsilon_0\vol\\
		\le&\;\epsilon_0 \vol+2\epsilon_0\vol=3\epsilon_0\vol=\exp(-\alpha/\gamma)(\epsilon \lambda )^{1/\lambda}.
	\end{align}
	 Applying Lemma~\ref{lem:KL-closeness-g}, we get item (b).
	
	\paragraph{Uniform concentration via covering.}
	We define the induced covering set $\hat\calC(w,f,\alpha)=\{\hat {g}:g'\in \calC(w,f,\alpha)\}.$ Then $|\hat\calC(w,f,\alpha)|\le |\calC(w,f,\alpha)|\le \calN(\calF,\epsilon_0).$ Applying Lemma~\ref{lem:individual-concentration} and union bound we get, with probability at least $1-\delta$,
	\begin{align}\label{equ:uc-2}
		\sum_{i=1}^{m}\ln \frac{f[\pi_i](\ob_i)}{\hat g[\pi_i](\ob_i)}\ge m(D^w_{1-\lambda}(f\|\hat g)-\epsilon),\quad \forall \hat g\in \hat\calC(w,f,\alpha).
	\end{align}
	By item (a) of the covering property,
	\begin{align}\label{equ:uc-3}
		\sum_{i=1}^{m}\ln \frac{f[\pi_i](\ob_i)}{g[\pi_i](\ob_i)}\ge \sum_{i=1}^{m}\ln \frac{f[\pi_i](\ob_i)}{\hat{g}[\pi_i](\ob_i)}-m\epsilon.
	\end{align}
	By item (b) of the covering property,
	\begin{align}\label{equ:uc-4}
		D^w_{1-\lambda}(f\|\hat{g})\ge \alpha-2\epsilon
	\end{align} 
	Combining Eqs.~\eqref{equ:uc-2}, \eqref{equ:uc-3}, \eqref{equ:uc-4}, with probability at least $1-\delta$,
	\begin{align}
		\sum_{i=1}^{m}\ln \frac{f[\pi_i](\ob_i)}{g[\pi_i](\ob_i)}\ge m(\alpha-4\epsilon).
	\end{align}
\end{proof}

\subsection{Uniform Concentration for RL with General Reward Distribution}
In the following, we present an uniform concentration lemma for tabular RL with general reward. 
\begin{lemma}\label{lem:uniform-concentration-Gaussian}
	Let $\calF$ be the instance family representing tabular RL with general reward distribution that satisfies Condition~\ref{cond:rl-reward}. For any fixed $\alpha>0,0<\epsilon<\alpha/2$, list of decisions $w=(\pi_1,\cdots,\pi_m)$, $f\in\calF$, let $\gamma=\frac{1}{m}\min_{\pi:\pi\in w} \sum_{i=1}^{m}\ind{\pi_i=\pi}$ and $\lambda=\lambda_0(\alpha/\gamma,\epsilon,f)$ be the value that satisfies Condition~\ref{cond:uniform-convergence}. Let $\calF(w,f,\alpha)=\{g\in\calF:\KL^w(f\|g)\ge \alpha\}$ and define 
	$$\epsilon_0=\exp(-\alpha/\gamma)(\epsilon \lambda )^{1/\lambda}.$$
	Then for any $\delta>0$, when $m\gtrsim \frac{\poly(|\calS||\calA|H)}{\lambda\epsilon}\((\ln (m/\epsilon_0))+\ln(1/\delta)\)$ we have
	\begin{align}\label{equ:uc-conclusion-Gaussian}
		\Pr_{\ob_i\sim f[\pi_i],\forall i}\(\forall g\in \calF(w,f,\alpha),\sum_{i=1}^{m}\ln \frac{f[\pi_i](\ob_i)}{g[\pi_i](\ob_i)}\ge (\alpha-4\epsilon)m\)\ge 1-\delta.
	\end{align}
\end{lemma}
\begin{proof}
	Recall that a tabular RL instance $g\in\calF$, we use $g_r[s,a]$ to denote its reward distribution given state-action pair $(s,a)$, and $g_p[s,a]$ its transition. We prove this lemma using the covering argument.
	
	\paragraph{The covering set.}
	Define $$\epsilon_1=\min\left\{\frac{\epsilon}{2H|\calS|},\frac{\epsilon}{2H\polylog(2mH/\delta)},\frac{\exp(-\alpha/\gamma)(\epsilon\lambda)^{1/\lambda}}{4|\calS|^2|\calA|}\right\}.$$
	Let $\calC\subseteq\calF$ be the minimum covering of $\calF$ such that for any $g\in\calF$ (parameterized by $p,\mu$), there exists $g'\in\calF$ (parameterized by $p',\mu'$) such that
	\begin{align}
		\sup_{s,a,s'}\abs{g_p[s,a](s')-g'_{p}[s,a](s')}\le \epsilon_1,\quad \sup_{s,a}\TV\(g_r[s,a]\|g'_r[s,a]\)\le \epsilon_1^{1/\const_7}.
	\end{align}
	By item (d) of Condition~\ref{cond:rl-reward} and a standard covering argument for discrete distributions, we have $\ln|\calC|\lesssim |\calS||\calA|\ln(1/\epsilon_1).$ For any $g'\in\calC$, we consider the induced instance $\hat{g}$ defined by
	\begin{align}
		\hat{g}_{p}[s,a](s')=\frac{g'_p[s,a](s')+\epsilon_1}{1+\epsilon_1|\calS|},\quad\hat{g}_r[s,a](\cdot)=g'_r[s,a](\cdot).
	\end{align}
	In the following, we prove that 
	\begin{enumerate}[label=(\alph*)]
		\item with probability at least $1-\delta/2$, 
		\begin{align}\label{equ:uc-covering-Gaussian}
			\Pr_{\ob_i\sim f[\pi_i],\forall i}\(\forall g\in \calF,\sum_{i=1}^{m}\ln \frac{\hat{g}[\pi_i](\ob_i)}{g[\pi_i](\ob_i)}\ge -m\epsilon\)\ge 1-\delta/2.
		\end{align}
		\item $D^w_{1-\lambda}(f\|\hat{g})\ge \alpha-2\epsilon$ for $g\in\calF(w,f,\alpha).$
	\end{enumerate}

	To prove (a), recall that $\ob_i=\{(s_{i,h},a_{i,h},r_{i,h})\}_{h=1}^{H}$ represents a trajectory. Consequently,
	\begin{align}
		&\sum_{i=1}^{m}\ln \frac{\hat{g}[\pi_i](\ob_i)}{g[\pi_i](\ob_i)}\\
		=&\;\sum_{i=1}^{m}\sum_{h=1}^{H}\ln \frac{\hat{g}_p[s_{i,h},a_{i,h}](s_{i,h+1})}{g_p[s_{i,h},a_{i,h}](s_{i,h+1})}+\sum_{i=1}^{m}\sum_{h=1}^{H}\ln \frac{\hat{g}_r[s_{i,h},a_{i,h}](r_{i,h})}{g_r[s_{i,h},a_{i,h}](r_{i,h})}\\
		\ge&\;-\sum_{i=1}^{m}\sum_{h=1}^{H}\ln(1+|\calS|\epsilon_1)+\sum_{i=1}^{m}\sum_{h=1}^{H}\ln \frac{\hat{g}_r[s_{i,h},a_{i,h}](r_{i,h})}{g_r[s_{i,h},a_{i,h}](r_{i,h})}.
	\end{align}
	By item (a) of Condition~\ref{cond:rl-reward} and union bound we get with probability at least $1-\delta/2$
	\begin{align}
		\sum_{i=1}^{m}\sum_{h=1}^{H}\ln \frac{\hat{g}_r[s_{i,h},a_{i,h}](r_{i,h})}{g_r[s_{i,h},a_{i,h}](r_{i,h})}\ge -mH\epsilon_1\polylog(mH/\delta)\ge -m\epsilon/2.
	\end{align}
	Therefore,
	\begin{align}
		&\sum_{i=1}^{m}\ln \frac{\hat{g}[\pi_i](\ob_i)}{g[\pi_i](\ob_i)}
		\ge-\sum_{i=1}^{m}\sum_{h=1}^{H}|\calS|\epsilon_1-m\epsilon/2
		\ge-m\epsilon,
	\end{align}
	which proves (a). For (b), by algebraic manipulation we have
	\begin{align}
		&\TV(g[\pi]\|\hat{g}[\pi])\le \sum_{s,a}\(\TV(p[s,a]\|\hat{g}_p[s,a])+\TV(r[s,a]\|\hat{g}_r[s,a])\)\\
		\le&\; \sum_{s,a,s'}\abs{\frac{p'[s,a](s')+\epsilon_1}{1+\epsilon_1|\calS|}-p[s,a](s')}+|\calS||\calA|\epsilon_1\\
		\le&\;\sum_{s,a,s'}\abs{\frac{p'[s,a](s')+\epsilon_1}{1+\epsilon_1|\calS|}-p'[s,a](s')-\epsilon_1}+\sum_{s,a,s'}\abs{p'[s,a](s')+\epsilon_1-p[s,a](s')}+|\calS||\calA|\epsilon_1\\
		\le&\;4|\calS|^2|\calA|\epsilon_1\le \exp(-\alpha/\gamma)(\epsilon\lambda)^{1/\lambda}.
	\end{align}
	Applying Lemma~\ref{lem:KL-closeness-g}, we get item (b).

	\paragraph{Uniform concentration via covering.}
	Now we apply Lemma~\ref{lem:individual-concentration} and union bound. When $$m\gtrsim \frac{|\calS||\calA|\ln(1/\epsilon_1)}{\lambda\epsilon}\gtrsim \frac{\ln (|\calC|/\delta)}{\lambda\epsilon},$$ with probability at least $1-\delta/2$ we get
	\begin{align}\label{equ:uc-2G}
		\sum_{i=1}^{m}\ln \frac{f[\pi_i](\ob_i)}{\hat g[\pi_i](\ob_i)}\ge m(D^w_{1-\lambda}(f\|\hat g)-\epsilon),\quad \forall \hat g\in \calC.
	\end{align}
	By item (a) of the covering property, with probability at least $1-\delta/2$,
	\begin{align}\label{equ:uc-3G}
		\sum_{i=1}^{m}\ln \frac{f[\pi_i](\ob_i)}{g[\pi_i](\ob_i)}\ge \sum_{i=1}^{m}\ln \frac{f[\pi_i](\ob_i)}{\hat{g}[\pi_i](\ob_i)}-m\epsilon.
	\end{align}
	By item (b) of the covering property,
	\begin{align}\label{equ:uc-4G}
		D^w_{1-\lambda}(f\|\hat{g})\ge \alpha-2\epsilon
	\end{align} 
	Combining Eqs.~\eqref{equ:uc-2G}, \eqref{equ:uc-3G}, \eqref{equ:uc-4G}, with probability at least $1-\delta$,
	\begin{align}
		\sum_{i=1}^{m}\ln \frac{f[\pi_i](\ob_i)}{g[\pi_i](\ob_i)}\ge m(\alpha-4\epsilon).
	\end{align}
\end{proof}

\section{Helper Lemmas}
The following lemma shows that the \renyi divergence of a marginal distribution is smaller than that of the original distribution.
\begin{lemma}[Theorem 1 of \citet{van2014renyi}]\label{lem:renyi-condition}
	Consider any distributions $f,g$ over variables $x,y$. Let $f(x),g(x)$ denote their marginal distribution over $x$ respectively. For any $\alpha\in (0,1)$, we have
	\begin{align}
		D_\alpha(f(x)\|g(x))\le D_\alpha(f(x,y)\|g(x,y)).
	\end{align}
\end{lemma}
\begin{proof}
	When $\alpha\in(0,1)$, we have $\frac{1}{\alpha-1}< 0.$ As a result, we only need to prove that 
	\begin{align}
		\int f(x,y)^{\alpha}g(x,y)^{1-\alpha}\dd xy \le \int f(x)^{\alpha}g(x)^{1-\alpha}\dd x.
	\end{align}
	We prove this by H\"older's inequality. In particular,
	\begin{align}
		&\int f(x,y)^{\alpha}g(x,y)^{1-\alpha}\dd xy\\
		=&\int f(x)^{\alpha}g(x)^{1-\alpha}\int f(y\mid x)^{\alpha}g(y\mid x)^{1-\alpha}\dd y\dd x\\
		\le &\int f(x)^{\alpha}g(x)^{1-\alpha}\(\int f(y\mid x)\dd y\)^{\alpha}\(\int g(y\mid x)\dd y\)^{1-\alpha}\dd x\\
		\le &\int f(x)^{\alpha}g(x)^{1-\alpha}\dd x.
	\end{align}
\end{proof}

Intuitively, the following two lemmas upper bound the difference of \renyi divergence by the TV distance.
\begin{lemma}\label{lem:renyi-covering-g}
	For and fixed $\lambda\in (0,1),\alpha>0,\epsilon>0$ and distribution $f$, consider two distributions $g,\hat{g}$ such that $\TV(g\|\hat{g})\le \exp(-\alpha)(\lambda\epsilon)^{1/\lambda}.$ Then we have
	\begin{align}
		D_{1-\lambda}(f\|\hat{g})\ge \min\{\alpha,D_{1-\lambda}(f\|g)\}-\epsilon.
	\end{align}
\end{lemma}
\begin{proof}
	Let $\kappa=\min\{\alpha,D_{1-\lambda}(f\|g)\}.$  We start by proving 
	\begin{align}
		\abs{\int f(x)^{1-\lambda}g(x)^{\lambda}\dd x-\int f(x)^{1-\lambda}\hat{g}(x)^{\lambda}\dd x}\le \(\exp(\lambda\epsilon)-1\)\exp(-\lambda \kappa).
	\end{align}
	By H\'older's inequality we get
	\begin{align}
		&\abs{\int f(x)^{1-\lambda}g(x)^{\lambda}\dd x-\int f(x)^{1-\lambda}\hat{g}(x)^{\lambda}\dd x}\\
		\le\;&\(\int\abs{g(x)-\hat{g}(x)}\dd x\)^{\lambda} \(\int f(x)\dd x\)^{1-\lambda}\\
		\le\;&\exp(-\lambda \alpha)\lambda\epsilon\\
		\le\;& \exp(-\lambda \alpha)\(\exp(\lambda\epsilon)-1\)\label{equ:rc-0}\\
		\le\;& \exp(-\lambda \kappa)\(\exp(\lambda\epsilon)-1\),\label{equ:rc-1}
	\end{align}
	where the Eq.~\eqref{equ:rc-0} follows from the basic inequality $1+x\le \exp(x)$ for $x>0.$
	
	By the definition of \renyi divergence,
	\begin{align}
		\int f(x)^{1-\lambda}g(x)^{\lambda}\dd x= \exp(-\lambda  D_{1-\lambda}(f\|g))\le \exp(-\lambda \kappa).\label{equ:rc-2}
	\end{align}
	Combining Eq.~\eqref{equ:rc-1} and Eq.~\eqref{equ:rc-2} we get,
	\begin{align}
		\int f(x)^{1-\lambda}\hat{g}(x)^{\lambda}\dd x\le \exp(-\lambda(\kappa-\epsilon)).
	\end{align}
	It follows that
	\begin{align}
		D_{1-\lambda}(f\|\hat{g})=-\frac{1}{\lambda}\ln\int f(x)^{1-\lambda}\hat{g}(x)^{\lambda}\dd x\ge \kappa-\epsilon.
	\end{align}
\end{proof}

\begin{lemma}\label{lem:renyi-covering-f}
	For and fixed $\lambda\in (0,1/2),\alpha>0,\epsilon>0$ and distribution $g$, consider two distributions $f,\hat{f}$ such that $\TV(f\|\hat{f})\le \exp\(-\frac{\lambda}{1-\lambda}\alpha\)(\lambda\epsilon)^{1/(1-\lambda)}.$ Then we have
	\begin{align}
		D_{1-\lambda}(\hat{f}\|g)\ge \min\{\alpha,D_{1-\lambda}(f\|g)\}-\epsilon.
	\end{align}
\end{lemma}
\begin{proof}
	We use a similar proof as Lemma~\ref{lem:renyi-covering-g}. Let $\kappa=\min\{\alpha,D_{1-\lambda}(f\|g)\}.$ Then we have
	\begin{align}
		&\abs{\int f(x)^{1-\lambda}g(x)^{\lambda}\dd x-\int \hat{f}(x)^{1-\lambda}g(x)^{\lambda}\dd x}\\
		\le\;&\(\int\abs{f(x)-\hat{f}(x)}\dd x\)^{1-\lambda} \(\int g(x)\dd x\)^{\lambda}\\
		\le\;&\exp\(-\lambda \alpha\)\lambda\epsilon\\
		\le\;& \exp(-\lambda \kappa)\(\exp(\lambda\epsilon)-1\),\label{equ:rcf-1}
	\end{align}

	By the definition of \renyi divergence,
	\begin{align}
		\int f(x)^{1-\lambda}g(x)^{\lambda}\dd x\le \exp(-\lambda  D_{1-\lambda}(f\|g))\le \exp(-\lambda \kappa).\label{equ:rcf-2}
	\end{align}
	Combining Eq.~\eqref{equ:rcf-1} and Eq.~\eqref{equ:rcf-2} we get,
	\begin{align}
		\int \hat{f}(x)^{1-\lambda}g(x)^{\lambda}\dd x\le \exp(-\lambda(\kappa-\epsilon)).
	\end{align}
	It follows that
	\begin{align}
		D_{1-\lambda}(\hat{f}\|g)=-\frac{1}{\lambda}\ln\int \hat{f}(x)^{1-\lambda}g(x)^{\lambda}\dd x\ge \kappa-\epsilon.
	\end{align}
\end{proof}

The following lemma is used to prove that when $\initf$ is close to $\truef$ (measured by TV distance), we can use $\initf$ to approximately solve $\comp(\truef)$ (see Lemma~\ref{lem:main-ident}).
\begin{lemma}\label{lem:KL-closeness-f}
	For an instance $f\in\calF$ and $n>0$. Let $\delta=(\ln\ln n)^{-1/4}$ and $\epsilon=(\ln\ln n)^{-1}.$
	For any $\hat{w}\in\R^{|\Pi|}_+$ such that $\|\hat{w}\|_\infty\le (\ln\ln n)^{1/4}$, 
	define $w=\{\pi_i\}_{i=1}^{m}$ be a list of decisions where a decision $\pi$ occurs $\lceil ((1+\delta)\hat{w}_\pi+\delta)\ln n\rceil$ times for every $\pi\in\Pi$, and $m=\sum_\pi \lceil ((1+\delta)\hat{w}_\pi+\delta)\ln n\rceil$. 
	Consider two instances $f,\initf\in\calF$ such that there exists constant $\const_6>0$ with $$\TV(f[\pi]\|\hat{f}[\pi])\lesssim \(\frac{\ln\ln n}{\ln n}\)^{\const_6},\forall \pi\in\Pi.$$
	
	Define the set 
	$\calF(\hat{w},f)=\{g\in\calF:\sum_{\pi\in\Pi}\hat{w}_\pi\KL(f[\pi]\|g[\pi])\ge 1\}.$ Then for any constant $c>0$, there exits $n_0>0$ such that for all $n>n_0$,
	\begin{align}
		\KL^w(\hat{f}\|g)\ge \frac{\ln n}{m}+c\epsilon,\quad \forall g\in\calF(\hat{w},f).
	\end{align}
\end{lemma}
\begin{proof}[Proof of Lemma~\ref{lem:KL-closeness-f}]
	First we invoke Condition~\ref{cond:uniform-convergence} with proper parameters. Define 
	\begin{align}
		\lambda=\lambda_0\(4(\ln\ln n)^{3/4},\frac{1}{\ln\ln n},f\)
	\end{align}
	By the definition of $m$ and the fact that $\|\hat{w}\|_\infty\le (\ln\ln n)^{1/4}$ we get
	\begin{align}
		\frac{|\Pi|\ln n}{(\ln\ln n)^{1/4}}\le m\le 2|\Pi|\ln n(\ln\ln n)^{1/4}.
	\end{align}
	Let $\alpha\defeq \frac{\ln n}{m}+(c+2)\epsilon\lesssim \frac{2}{|\Pi|} (\ln\ln n)^{1/4}.$ Consider $\gamma=\min_{\pi}\frac{1}{m}\lceil ((1+\delta)\hat{w}_\pi+\delta)\ln n\rceil$. By the upper bound of $m$ we get $\gamma\ge \frac{1}{2|\Pi|(\ln\ln n)^{1/2}}.$
	
	We invoke Condition~\ref{cond:uniform-convergence} with parameters $(\alpha/\gamma,\epsilon,f)$. For large enough $n$ we have $\alpha/\gamma\le 4(\ln\ln n)^{3/4}$, which implies that $\lambda<\lambda_0(\alpha/\gamma,\epsilon,f)$. Then for any $\pi\in\Pi$ we get
	\begin{align}\label{equ:kcf-2}
		D_{1-\lambda}(f[\pi]\|g[\pi])\ge \min\{\alpha/\gamma,\KL(f[\pi]\|g[\pi])-\epsilon\}.
	\end{align}
	
	We claim that $\KL^w(f\|g)\ge \frac{\ln n}{m}+(c+2)\epsilon,\forall g\in\calF(\hat{w}, f)$ for large enough $n$. Indeed, by the definition of $w$ we get
	\begin{align}
		m\KL^w(f\|g)\ge \sum_\pi ((1+\delta)\hat{w}_\pi+\delta)(\ln n)\KL(f[\pi]\|g[\pi])\ge (1+\delta)\ln n\ge \ln n +(c+2)m\epsilon.
	\end{align}
	Let $$\epsilon_1=\exp(-\alpha/\gamma)(\lambda\epsilon)^2=\Omega\(\exp(-(\ln\ln n)^{3/4})\poly(\ln\ln n)\).$$
	By Lemma~\ref{lem:renyi-covering-f} with parameters $(\lambda,\alpha/\gamma,\epsilon)$ and the assumption that $\TV(f[\pi]\|\hat{f}[\pi])\lesssim \(\frac{\ln\ln n}{\ln n}\)^{\const_6}=o(\epsilon_1)$ for all $\pi\in\Pi$, we get
	\begin{align}\label{equ:kcf-1}
		D_{1-\lambda}(\hat{f}[\pi]\|g[\pi])\ge \min\{\alpha/\gamma,D_{1-\lambda}(f[\pi]\|g[\pi])\}-\epsilon,\forall \pi\in\Pi.
	\end{align} 
	Now we consider the following two cases.
	
	\paragraph{Case 1:} There exists $\pi\in w$ such that $\KL(f[\pi]\|g[\pi])\ge \alpha/\gamma$. In this case Eq.~\eqref{equ:kcf-2} implies that $D_{1-\lambda}(f[\pi]\|g[\pi])\ge \alpha/\gamma-\epsilon.$ Combining with Eq.~\eqref{equ:kcf-1} we have
	\begin{align}
		D_{1-\lambda}(\initf[\pi]\|g[\pi])\ge \alpha/\gamma-2\epsilon.
	\end{align} As a result,
	\begin{align}
		D^w_{1-\lambda}(\hat{f}\|g)\ge \frac{1}{m}\sum_{i=1}^{m}\ind{\pi_i=\pi}(\alpha/\gamma-2\epsilon)\ge \alpha-2\epsilon=\frac{\ln n}{m}+c\epsilon.
	\end{align}
	
	\paragraph{Case 2:} For all $\pi\in w$, $\KL(f[\pi]\|g[\pi])\le \alpha/\gamma$. In this case we also have $D_{1-\lambda}(f[\pi]\|g[\pi])\le \alpha/\gamma,\;\forall \pi\in w$. Therefore Eq.~\eqref{equ:kcf-1} and Eq.~\eqref{equ:kcf-2} implies that
	\begin{align}
		D_{1-\lambda}(\initf[\pi]\|g[\pi])&\ge D_{1-\lambda}(f[\pi]\|g[\pi])-\epsilon,\\
		D_{1-\lambda}(f[\pi]\|g[\pi])&\ge \KL(f[\pi]\|g[\pi])-\epsilon.
	\end{align} 
	As a result,
	\begin{align}
		&D^w_{1-\lambda}(\hat{f}\|g)\ge \frac{1}{m}\sum_{i=1}^{m}\(D_{1-\lambda}(f[\pi_i]\|g[\pi_i])-\epsilon\)\ge\frac{1}{m}\sum_{i=1}^{m}\(\KL(f[\pi_i]\|g[\pi_i])-2\epsilon\)\\
		=\;&\KL^w(f\|g)-2\epsilon \ge \frac{\ln n}{m}+c\epsilon.
	\end{align}
	
	Combining the two cases together, we get $D^w_{1-\lambda}(\hat{f}\|g)\ge \frac{\ln n}{m}+c\epsilon.$
\end{proof}

The following lemma is used to prove a nice property of the covering (see Lemma~\ref{lem:uniform-concentration}).
\begin{lemma}\label{lem:KL-closeness-g}
	Consider any $\epsilon>0,\alpha>0$, sequence of decisions $w=\{\pi_i\}_{i=1}^{m}$. Let $\gamma=\frac{1}{m}\min_{\pi:\pi\in w} \sum_{i=1}^{m}\ind{\pi_i=\pi}$ and $\lambda=\lambda_0(\alpha/\gamma,\epsilon,f)$ be the value that satisfies Condition~\ref{cond:uniform-convergence}.
	For two distributions $f,\initf\in\calF$ such that $$\TV(g[\pi]\|\hat{g}[\pi])\le \exp(-\alpha/\gamma)(\lambda\epsilon)^{1/\lambda},\forall \pi\in\Pi.$$ we have
	\begin{align}
		D^w_{1-\lambda}(f\|\hat{g})\ge \min\{\KL^{w}(f\|g),\alpha\}-2\epsilon.
	\end{align}
\end{lemma}
\begin{proof}[Proof of Lemma~\ref{lem:KL-closeness-g}]
Let $\epsilon_1=\exp(-\alpha/\gamma)(\lambda\epsilon)^{1/\lambda}$ and $\kappa=\min\{\KL^{w}(f\|g),\alpha\}$. 

By Lemma~\ref{lem:renyi-covering-g} and the fact that $\TV(g[\pi]\|\hat{g}[\pi])\le \epsilon_1,\forall \pi\in\Pi$, for any $\pi\in\Pi$ we have
\begin{align}\label{equ:kcg-1}
	D_{1-\lambda}(f[\pi]\|\hat{g}[\pi])\ge \min\{\alpha/\gamma,D_{1-\lambda}(f[\pi]\|g[\pi])\}-\epsilon.
\end{align}
Applying Condition~\ref{cond:uniform-convergence}, for any $\pi\in\Pi$ we have
\begin{align}\label{equ:kcg-2}
	D_{1-\lambda}(f[\pi]\|g[\pi])\ge \min\{\alpha/\gamma,\KL(f[\pi]\|g[\pi])-\epsilon\}.
\end{align}
Now we consider the following two cases.

\paragraph{Case 1:} There exists $\pi\in w$ such that $\KL(f[\pi]\|g[\pi])\ge \kappa/\gamma$. By the definition of $\kappa$ we have $\kappa\le \alpha$. In this case Eq.~\eqref{equ:kcg-2} implies that $D_{1-\lambda}(f[\pi]\|g[\pi])\ge \kappa/\gamma-\epsilon.$ Combining with Eq.~\eqref{equ:kcg-1} we have
\begin{align}
	D_{1-\lambda}(f[\pi]\|\hat{g}[\pi])\ge \kappa/\gamma-2\epsilon.
\end{align} As a result,
\begin{align}
	D^w_{1-\lambda}(f\|\hat{g})\ge \frac{1}{m}\sum_{i=1}^{m}\ind{\pi_i=\pi}(\kappa/\gamma-2\epsilon)\ge \kappa-2\epsilon.
\end{align}

\paragraph{Case 2:} For all $\pi\in w$, $\KL(f[\pi]\|g[\pi])\le \kappa/\gamma$. In this case we also have $D_{1-\lambda}(f[\pi]\|g[\pi])\le \kappa/\gamma,\;\forall \pi\in w$. Therefore Eq.~\eqref{equ:kcg-1} and Eq.~\eqref{equ:kcg-2} implies that
\begin{align}
	D_{1-\lambda}(f[\pi]\|\hat{g}[\pi])&\ge D_{1-\lambda}(f[\pi]\|g[\pi])-\epsilon,\\
	D_{1-\lambda}(f[\pi]\|g[\pi])&\ge \KL(f[\pi]\|g[\pi])-\epsilon.
\end{align} 
As a result,
\begin{align}
	&D^w_{1-\lambda}(f\|\hat{g})\ge \frac{1}{m}\sum_{i=1}^{m}\(D_{1-\lambda}(f[\pi_i]\|g[\pi_i])-\epsilon\)\ge\frac{1}{m}\sum_{i=1}^{m}\(\KL(f[\pi_i]\|g[\pi_i])-2\epsilon\)\\
	=\;&\KL^w(f\|g)-2\epsilon \ge \kappa-2\epsilon.
\end{align}

Combining the two cases together, we get $D^w_{1-\lambda}(f\|\hat{g})\ge \kappa-2\epsilon.$
\end{proof}

The following lemma shows that for truncated Gaussian distributions, the difference in their density function can be upper bounded by their KL divergence. We use this lemma to prove Condition~\ref{cond:TV-vs-inf} for tabular RL with truncated Gaussian reward.
\begin{lemma}\label{lem:closeness-truncated-Gaussian}
	Consider two truncated Gaussian distributions $p_1,p_2$ with density
	\begin{align}
		p_i(x)=\ind{x\in [-2,2]}\frac{1}{Z_i}\exp\(-\frac{(x-\mu_i)^2}{2}\),
	\end{align}
	where $Z_i$ is the normalization factor. Assuming $\mu_1,\mu_2\in[-1,1]$, we have
	\begin{align}
		\sup_{x\in [-2,2]}\abs{p_1(x)-p_2(x)}\le 36\KL(p_1\|p_2)^{1/6}.
	\end{align}
\end{lemma}
\begin{proof}
	We first prove that $\abs{\mu_1-\mu_2}\lesssim \KL(p_1\|p_2)^{1/6}.$ Then we show that $\sup_{x\in [-2,2]}\abs{p_1(x)-p_2(x)}\lesssim \abs{\mu_1-\mu_2}$.
	
	By Pinsker's inequality, 
	\begin{align}
		\TV(p_1\|p_2)\lesssim \KL(p_1\|p_2)^{1/2}.
	\end{align}
	Now we prove that $\abs{\mu_1-\mu_2}^3\lesssim \TV(p_1\|p_2).$ W.l.o.g., we assume $Z_1\ge Z_2> 1/\sqrt{2\pi}$ and $\mu_1\le \mu_2.$ Then we have, for any $x\in[\mu_1,\mu_1+\frac{1}{4}(\mu_2-\mu_1)]$,
	\begin{align}
		p_1(x)-p_2(x)&\ge \frac{1}{\sqrt{2\pi}}\[\exp(-\frac{(\mu_2-\mu_1)^2}{32})-\exp(-\frac{9(\mu_2-\mu_1)^2}{32})\]\\
		&\ge \frac{1}{\sqrt{2\pi}}\exp\(-\frac{9(\mu_2-\mu_1)^2}{32}\)\(\exp(\frac{1}{4}(\mu_2-\mu_1)^2)-1\)\\
		&\ge\frac{1}{4e^2\sqrt{2\pi}}(\mu_2-\mu_1)^2.
	\end{align}
	As a result,
	\begin{align}\label{equ:ctG-1}
		\TV(p_1\|p_2)\ge \frac{1}{4}(\mu_2-\mu_1)(p_1(x)-p_2(x))\gtrsim \abs{\mu_2-\mu_1}^3.
	\end{align}
	
	Now we prove that $\sup_{x\in [-2,2]}\abs{p_1(x)-p_2(x)}\lesssim \abs{\mu_1-\mu_2}$. By definition, for any $x\in [-2,2]$ we have
	\begin{align}
		\abs{p_1(x)-p_2(x)}&=\abs{\frac{1}{Z_1}\exp\(-\frac{(x-\mu_1)^2}{2}\)-\frac{1}{Z_2}\exp\(-\frac{(x-\mu_2)^2}{2}\)}\\
		&\le \abs{1/Z_1-1/Z_2}+\frac{1}{Z_1}\abs{\exp\(-\frac{(x-\mu_1)^2}{2}\)-\exp\(-\frac{(x-\mu_2)^2}{2}\)},\\
		&\le 4\abs{Z_1-Z_2}+2\abs{\exp\(-\frac{(x-\mu_1)^2}{2}\)-\exp\(-\frac{(x-\mu_2)^2}{2}\)},
	\end{align}
	where the last inequality comes from the fact that $Z_i\ge 1/2$ when $|\mu_i|\le 1.$ For the second term,
	\begin{align}
		&\abs{\exp\(-\frac{(x-\mu_1)^2}{2}\)-\exp\(-\frac{(x-\mu_2)^2}{2}\)}\\
		=\;&\exp\(-\frac{(x-\mu_2)^2}{2}\)\abs{\exp((\mu_2-\mu_1)(\mu_2+\mu_1-x)/2)-1}\\
		\le\;&2|\mu_2-\mu_1|.
	\end{align} 
	For the first term,
	\begin{align}
		\abs{Z_1-Z_2}\le \int_{x=-2}^{2}\abs{p_1(x)-p_2(x)}\dd x\le 8|\mu_2-\mu_1|.
	\end{align}
	As a result, we get 
	\begin{align}\label{equ:ctG-2}
		\abs{p_1(x)-p_2(x)}\le 36|\mu_2-\mu_1|.
	\end{align}
	Combining Eq.~\eqref{equ:ctG-1} and Eq.~\eqref{equ:ctG-2} we prove the this lemma.
\end{proof}

The following lemma can be viewed as an perturbation analysis to the covariance matrix in linear bandits setting. We use the lemma to prove that our complexity measure recovers that in \citep{lattimore2017end}.
\begin{lemma}\label{lem:lb-cap}
	For a fixed positive definite matrix $A\in\R^{d\times d}$ and an unit vector $x\in\R^{d}$, let $G_1=(A+nxx^\top)^{-1}$ and $G_2=\lim_{\rho\to\infty}(A+\rho xx^\top)^{-1}.$ Then
	\begin{align}
		\|G_1-G_2\|_2\le \frac{\sigmamax(A^{-1})}{1+n\sigmamin(A^{-1})}.
	\end{align}
\end{lemma}
\begin{proof}
	By Sherman–Morrison formula we get
	\begin{align}
		G_1&=A^{-1}-\frac{nA^{-1}xx^\top A^{-1}}{1+nx^\top A^{-1}x},\\
		G_2&=\lim_{\rho\to\infty}A^{-1}-\frac{\rho A^{-1}xx^\top A^{-1}}{1+\rho x^\top A^{-1}x}.
	\end{align}
	Then for any $v\in\R^{d}$ such that $\|v\|_2=1$, we get
	\begin{align}
		v^\top(G_1-G_2)v&=\lim_{\rho\to\infty}\frac{\rho v^\top A^{-1}xx^\top A^{-1}v}{1+\rho x^\top A^{-1}x}-\frac{nv^\top A^{-1}xx^\top A^{-1}v}{1+nx^\top A^{-1}x}\\
		&=\lim_{\rho\to\infty} \frac{(\rho-n)v^\top A^{-1}xx^\top A^{-1}v}{(1+nx^\top A^{-1}x)(1+\rho x^\top A^{-1}x)}\\
		&=\frac{v^\top A^{-1}xx^\top A^{-1}v}{x^\top A^{-1}x(1+nx^\top A^{-1}x)}\\
		&\le \frac{v^\top A^{-1}v}{1+nx^\top A^{-1}x}\\
		&\le \frac{\sigmamax(A^{-1})}{1+n\sigmamin(A^{-1})}.
	\end{align}
\end{proof}

The following lemma upper bounds the difference between KL divergence and \renyi divergence, and is used to prove Condition~\ref{cond:uniform-convergence}.
\begin{lemma}\label{lem:KL-renyi-difference}
	For any two distribution $f,g$ and constant $\lambda\in(0,1/2),$ we have
	\begin{align}
		\KL(f\|g)-D_{1-\lambda}(f\|g)\le \frac{\lambda}{2}\E_{\ob\sim f}\[\(\ln \frac{f(\ob)}{g(\ob)}\)^4\]^{1/2}.
	\end{align}
\end{lemma}
\begin{proof}
	Recall that $$D_{1-\lambda}(f\|g)=-\frac{1}{\lambda}\ln\int f(\ob)^{1-\lambda}g(\ob)^\lambda\dd \ob.$$
	Define the function $h(\lambda)\defeq\int f(\ob)^{1-\lambda}g(\ob)^{\lambda}\dd \ob.$ By basic algebra we get
	\begin{align}
		h'(\lambda)&=\int f(\ob)^{1-\lambda}g(\ob)^{\lambda}\ln \frac{g(\ob)}{f(\ob)}\dd \ob\\
		h''(\lambda)&=\int f(\ob)^{1-\lambda}g(\ob)^{\lambda}\ln^2 \frac{g(\ob)}{f(\ob)}\dd \ob.
	\end{align}
	By Taylor expansion, there exists $\xi\in(0,\lambda)$ such that
	\begin{align}
		&h(\lambda)=h(0)+\lambda h'(0)+\frac{\lambda^2}{2}h''(\xi).
	\end{align}
	By definition we have $h(0)=1$ and $h'(0)=-\KL(f\|g).$ As a result, we get
	\begin{align}
		&D_{1-\lambda}(f\|g)=-\frac{1}{\lambda}\ln h(\lambda)\\
		=\;&-\frac{1}{\lambda}\ln \(1-\lambda \KL(f\|g)+\frac{\lambda^2}{2}\int f(\ob)^{1-\zeta}g(\ob)^{\zeta}\ln^2 \frac{g(\ob)}{f(\ob)}\dd \ob\)\\
		\ge\;&\KL(f\|g)-\frac{\lambda}{2}\int f(\ob)^{1-\xi}g(\ob)^{\xi}\ln^2 \frac{g(\ob)}{f(\ob)}\dd \ob.
	\end{align}
	By H\"older's inequality, when $\xi<\lambda<1/2$ we get
	\begin{align}
		&\int f(\ob)^{1-\zeta}g(\ob)^{\xi}\ln^2 \frac{g(\ob)}{f(\ob)}\dd \ob\\
		=\;&\E_{\ob\sim f}\[\(\frac{g(\ob)}{f(\ob)}\)^{\xi}\ln^2 \frac{g(\ob)}{f(\ob)}\]\\
		\le\;&\E_{\ob\sim f}\[\frac{g(\ob)}{f(\ob)}\]^{\xi}\E_{\ob\sim f}\[\(\ln \frac{g(\ob)}{f(\ob)}\)^{\frac{2}{1-\xi}}\]^{1-\xi}\\
		\le\;&\E_{\ob\sim f}\[\(\ln \frac{g(\ob)}{f(\ob)}\)^{4}\]^{1/2}.
	\end{align}
	Combining the inequalities above we get the desired result.
\end{proof}
	
\end{document}